\documentclass[twoside]{article}
\usepackage[accepted]{aistats2023}

\usepackage{amsmath,amsfonts,bm}

\def\eqref#1{equation~\ref{#1}}

\def\1{\bm{1}}

\def\rvc{{\mathbf{c}}}

\def\rve{{\mathbf{e}}}

\def\rvo{{\mathbf{o}}}

\def\rvz{{\mathbf{z}}}

\DeclareMathAlphabet{\mathsfit}{\encodingdefault}{\sfdefault}{m}{sl}
\SetMathAlphabet{\mathsfit}{bold}{\encodingdefault}{\sfdefault}{bx}{n}

\usepackage{hyperref}
\usepackage{url}
\usepackage{graphics}
\usepackage{graphicx}
\usepackage{multirow}
\usepackage{subfigure}
\usepackage{xcolor}
\usepackage{booktabs}
\usepackage{wrapfig}
\usepackage{sidecap}
\usepackage{etoc}

\usepackage{pifont}
\newcommand{\highlight}[1]{\colorbox{blue!10}{#1}}
\definecolor{mygray}{gray}{0.4}

\newcommand{\g}[2]{#1\textsubscript{\textcolor{mygray}{$\pm$#2}}}
\newcommand{\modelname}{{\textsc{RepDIB}}}

\newcommand{\tablestyle}[2]{\setlength{\tabcolsep}{#1}\renewcommand{\arraystretch}{#2}\centering\footnotesize}

\newlength\savewidth\newcommand\shline{\noalign{\global\savewidth\arrayrulewidth
  \global\arrayrulewidth 1pt}\hline\noalign{\global\arrayrulewidth\savewidth}}

\usepackage{comment}
\usepackage{todonotes}
\usepackage{floatrow}
\newfloatcommand{capbtabbox}{table}[][\FBwidth]
\usepackage{blindtext}

\definecolor{mygray}{gray}{0.4}

\usepackage{makecell}
\usepackage{algorithm}
\usepackage{algorithmic}
\usepackage{mathabx}
\usepackage{amsmath}
\usepackage{amssymb}
\usepackage{listings}
\usepackage{xcolor}
\definecolor{codegreen}{rgb}{0,0.6,0}
\definecolor{codegray}{rgb}{0.5,0.5,0.5}
\definecolor{codepurple}{rgb}{0.58,0,0.82}
\definecolor{backcolour}{rgb}{0.95,0.95,0.92}

\lstdefinestyle{mystyle}{
    backgroundcolor=\color{backcolour},   
    commentstyle=\color{codegreen},
    keywordstyle=\color{magenta},
    numberstyle=\tiny\color{codegray},
    stringstyle=\color{codepurple},
    basicstyle=\ttfamily\footnotesize,
    breakatwhitespace=false,         
    breaklines=true,                 
    captionpos=b,                    
    keepspaces=true,                 
    numbers=left,                    
    numbersep=5pt,                  
    showspaces=false,                
    showstringspaces=false,
    showtabs=false,                  
    tabsize=2
}

\begin{document}

\twocolumn[

\vspace{-4mm}
\aistatstitle{Representation Learning in Deep RL via Discrete Information Bottleneck}

\aistatsauthor{Riashat Islam* \And Hongyu Zang* \And Manan Tomar \And Aniket Didolkar
}
\aistatsaddress{Mila, McGill University\\
Microsoft Research Montreal \And Beijing Institute of Technology \And AMII, University of Alberta\\
Microsoft Research Montreal \And 
Mila, University of Montreal} 

\aistatsauthor{Md Mofijul Islam \And Samin Yeasar Arnob \And Tariq Iqbal \And Xin Li
}
\aistatsaddress{University of Virginia \And Mila, McGill University \And University of Virginia \And 
Beijing Institute of Technology} 

\aistatsauthor{Anirudh Goyal \And Nicolas Heess \And Alex Lamb
}
\aistatsaddress{Google DeepMind \And Google DeepMind \And 
Microsoft Research NYC} 
]
\runningauthor{Riashat Islam, Hongyu Zang, Manan Tomar, et al.}

\vspace{-4mm}
\begin{abstract}
\vspace{-2mm}
Several self-supervised representation learning methods have been proposed for reinforcement learning (RL) with rich observations. For real world applications of RL, recovering underlying latent states is crucial, particularly when sensory inputs contain irrelevant and exogenous information. In this work, we study how information bottlenecks can be used to construct latent states efficiently in the presence of task irrelevant information. We propose architectures that utilize variational and discrete information bottlenecks, coined as $\modelname$, to learn structured factorized representations. Exploiting the expressiveness bought by factorized representations, we introduce a simple, yet effective, bottleneck that can be integrated with any existing self supervised objective for RL. We demonstrate this across several online and offline RL benchmarks, along with a real robot arm task, where we find that compressed representations with $\modelname$ can lead to strong performance improvements, as the learnt bottlenecks help predict only the relevant state, while ignoring irrelevant information.  

\end{abstract}
\vspace{-6mm}
\section{Introduction}
\vspace{-2mm}

In the most general reinforcement learning (RL) setting, an agent is tasked with discovering a policy that achieves high long-term reward \cite{sutton2018reinforcement,mnih2013playing}. One of the key challenges of the RL setting is that credit assignment, exploration, and generalization \cite{sutton2018reinforcement} must be addressed even when the agent has seen very little data and thus has low quality representations \cite{kakade2003sample,foster2021sample}.  When the representations are low quality, determining a desirable state to reach and finding a policy to reach that state are both difficult \cite{huang2021sample}.  Intuitively, learning a compressed representation should help to address these challenges.  If extraneous information can be removed, it should be easier to generalize to new samples from the environment.  
 
\begin{figure*}
    \includegraphics[width=0.5\textwidth]{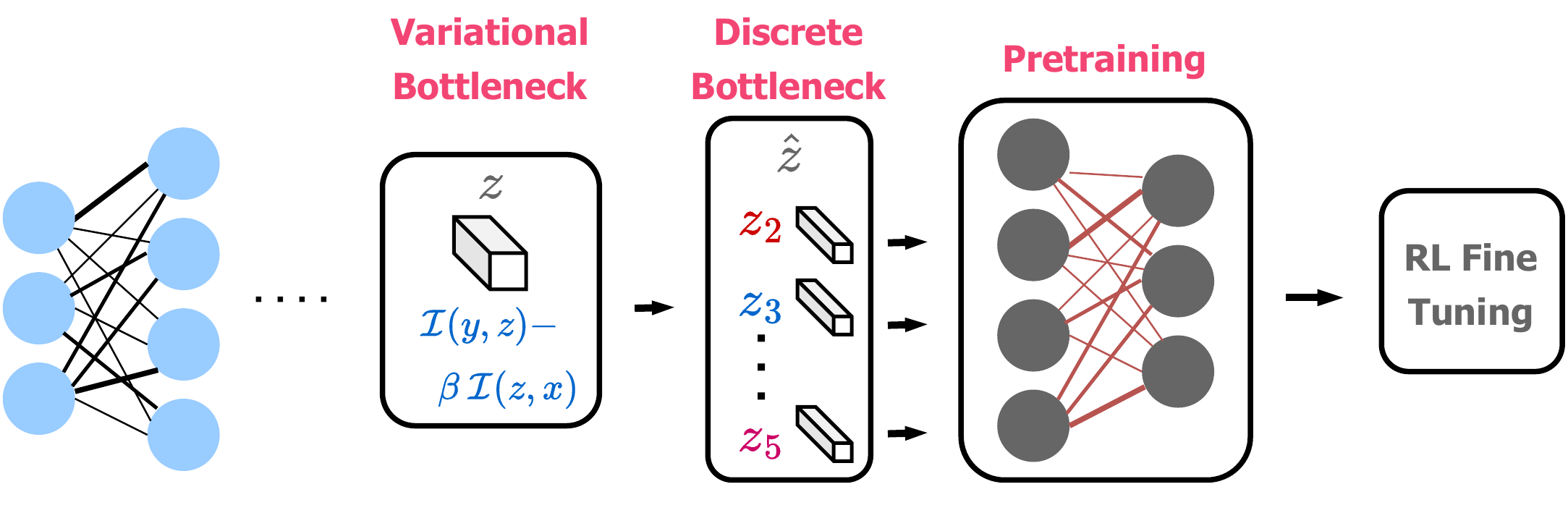}
    \label{fig:archit}%
    \caption{\textbf{Illustration} of the generic approach of $\modelname$, where we learn representations with variational and discrete factorial bottlenecks. We show that pre-training Representations with Discrete Information Bottleneck ($\modelname$) leads to learning of robust representations, especially when observations consist of irrelevant and exogenous information }
\end{figure*}

Approaches from the RL theory literature have shown benefits from compressed representations in the discrete latent state setting \cite{misra2020kinematic}, \cite{Efroni2021ppe}, \cite{du2019provable}, \cite{xiong2021randomized}. The HOMER algorithm \cite{misra2020kinematic} explores by trying to reach the frontier of pairs of the discrete latent states and actions with the lowest counts.  While these algorithms give strong theoretical guarantees \cite{efroni2022colt}, planning and exploring with them does not scale beyond a small number of discrete states.  

We explore the intersection between theoretically-grounded representation learning in small tabular-MDPs and representations for the deep reinforcement learning setting.  We seek to retain the expressiveness of factorial representations while making the representation compressed \cite{liu2021dvnc, liu2022adaptive}.  In our proposed method (Figure \ref{fig:archit}), Representations for RL with Discrete Information Bottleneck ($\modelname$), we make the representations discrete and factorial, while also encouraging them to be parsimonious through a gaussian variational information bottleneck \cite{alemi2016deep, InfoBotGoyal, goyal2019reinforcement, goyal2020variational}. These are expressive enough to model complicated environments, yet avoid the unbounded complexity of unstructured continuous representations. 

This work studies the effectiveness of learning compressed representations for reinforcement learning. We find that by using an information bottleneck that induces a factorial structure in the embedding space, $\modelname$ can learn more robust representations.  This improvement is especially pronounced in settings where the observation contains exogenous noise \cite{Efroni2021ppe, efroni2022colt}, which is any information that is unrelated to the agent's actions.  We propose an easy to use approach effective for improving downstream performance in settings with irrelevant background information. Our work offers the the following contributions: (a) Learning representations 
which more closely match the salient attributes of the environment, and improved robustness by learning factorial representations that can ignore irrelevant information in a practical robot arm task (b) Improved sample efficiency due to structured representations, for better generalization in continuous control (c) bottleneck representations that can improve robustness in offline RL in presence of exogenous distractors. Through range of experiments, we show that $\modelname$ learns compressed representations, which helps in exploration and reward-free pre-training of representations to improve efficiency and robustness on downstream tasks.

\vspace{-3mm}
\section{Related Work}
\label{sec:related_work}
\vspace{-2mm}

\textbf{Self Supervised Representation Learning in RL}. Several prior works have studied representation learning in context of RL, ranging from online to offline settings \cite{NachumRepMatters, KostrikovOfflineCritic, NachumImitation}, while also studying the ability to recover underlying latent states to capture environment dynamics \cite{lamb2022guaranteed, BallLPR21}. Most of these works involve learning representations from high dimensional observations, which may contain irrelevant information. This is formalized as learning under irrelevant exogenous information \cite{Efroni2021ppe, efroni2022colt},  by the theoretical RL community studying representation learning. In this work, we show effectiveness of information bottlenecks with \modelname, when learning under exogenous information, and show that bottlenecks can filter out irrelevant information from observations. Empirically, prior works studied regularized objectives, for learning robust representations \cite{MazoureCDBH20, JaderbergMCSLSK17} while others have exploited empowerment based objectives \cite{mohamed2015variational}. Self supervised objectives, when used for pre-training representations have shown to achieve tremendous performance improvements \cite{CURL, ATC,SchwarzerAGHCB21, SchwarzerRNACHB21}, while when learning with fine-tuning representations, it leads to better exploratory objectives~\cite{yarats2021protorl}. 

\textbf{Learning Minimal Representations with Information Bottleneck}. In this work, we argue that information bottleneck based representations with $\modelname$ can be an effective approach for learning robust representatons in RL, in presence of exogenous information. The information bottleneck principle~\cite{wang2022rethinking,tishby2015deep,shwartz2017opening,tishby2000information} advocates for learning minimal sufficient representations, i.e. those which contain \textit{only} sufficient information for the downstream task. Optimal representations contain relevant information between $X$ and $Y$ that is parsimonious to learn a task. Several approaches have been proposed to design information bottlenecks in deep learning models, such as variational bottlenecks \cite{sun2022graph,alemi2016deep} and discrete representation bottlenecks \cite{discrete_bottleneck}. Most prominently, Alemi et al. \cite{alemi2016deep} introduced a variational approximation to a mutual information objective of the information bottleneck and applied this to deep neural networks.

\textbf{Information Bottleneck for Exploration in Deep Reinforcement Learning}. The exploration problem is inherently coupled with the representation learning problem, since discovering underlying latent structure of the world ensures that the agent learns about the unseen frontiers in observation space to reach. While several recent works have studied representation learning in RL for improving downstream task performance~\cite{SchwarzerAGHCB21, SchwarzerRNACHB21}, the closest to our work is learning with prototypical representations ~\cite{yarats2021protorl}, which studies the coupled problem of representation learning and exploration. \cite{InfoBotGoyal, goyal2020variational} previously studied exploration based on identifying latent bottleneck states, but do not learn an explicit representation with a self-supervised objective. \cite{DropBottleneck} studied bottlenecks for inducing exploration in RL. On the theoretical side, \cite{misra2020kinematic} grounds representation learning and exploration with theoretical guarantees, but cannot scale to rich observation environments. Several prior works in exploration have been proposed, with large observation spaces, such as using pseudo-counts \cite{OstrovskiBOM17, BellemareSOSSM16}, optimism-driven exploration \cite{OsbandRRW19}, intrinsic motivation \cite{OudeyerK09}, random network distillation\cite{burda2018exploration} and curiostiy based exploration with prediction errors \cite{pathak2017curiosity}. While these algorithms proposes exploration in complex high dimensional tasks, they do not necessarily learn and exploit any form of structure in the representation space.

\textbf{Comparisons with Prior Related Works : } An information bottleneck aiming at minimal sufficient representations can be implemented in various ways, including a variational approach (VIB) and architectural choices such as reducing the dimension of deeper layers, or by discretizing layers. Chenjia et al.~\cite{BaiDB}. directly apply the information bottleneck to the dynamics of the system, whereas RepDIB applies it for different downstream targets, such as DQN targets or inverse model targets. RepDIB also combines both kinds of bottlenecks, i.e. architectural (discrete bottlenecks in particular) and variational ones. Previous work in reinforcement learning which enforce bottlenecks have worked with either type independently. Dreamer-v2 and similar variants have included discretization for pixel-level model-based learning. In this paper, we take a zoomed-out perspective on the efficacy of bottlenecks in learning representations for reinforcement learning.

\vspace{-2mm}
\section{Discrete Factorial Information Bottlenecks in Representation Learning}
\label{sec:variational}
\vspace{-2mm}
The goal of this work is to study the effectiveness of variational and discrete information bottlenecks in representation learning. While several prior works have studied representation learning for RL, we show that especially when observations can contain irrelevant information, addition of simple bottlenecks can lead to learning effective robust representations for improving performance on downstream tasks. Through a range of experiments, as in section \ref{sec:experiments}, we show that \modelname learns a structured representation space, via use of discrete information bottlenecks \cite{liu2022adaptive}, that can be quite effective for downstream learning. In this section, we briefly describe our approach for learning robust representations with information bottlenecks. 

The $\modelname$ technique begins with a hidden representation $\rvz \in \mathbb{R}^m$ for a rich-observation $x$.  This $\rvz$ could be the output of a convolutional neural network, a recurrent neural network, transformer, or any other expressive neural model.  We induce a compositional structure in the learnt representation space by using a vector quantization discretization bottleneck \cite{van2017neural}. This is achieved by using a discretization module with $G$ factors each with $L$ codes.  Thus the total number of discrete states that we can express is $L^G$.  We can learn embeddings using multiple G factors and can concatenate them into a single embedding $\hat{\rvz} = \phi(\rvz)$ with $\phi : \mathbb{R}^{m} \xrightarrow{} \mathbb{R}^m$.  Thus the discretization bottleneck $\phi$ preserves the size of the hidden representation.  

\begin{figure}[t]
\label{fig:kmeans}
    \centering
    \includegraphics[width=0.49\textwidth]{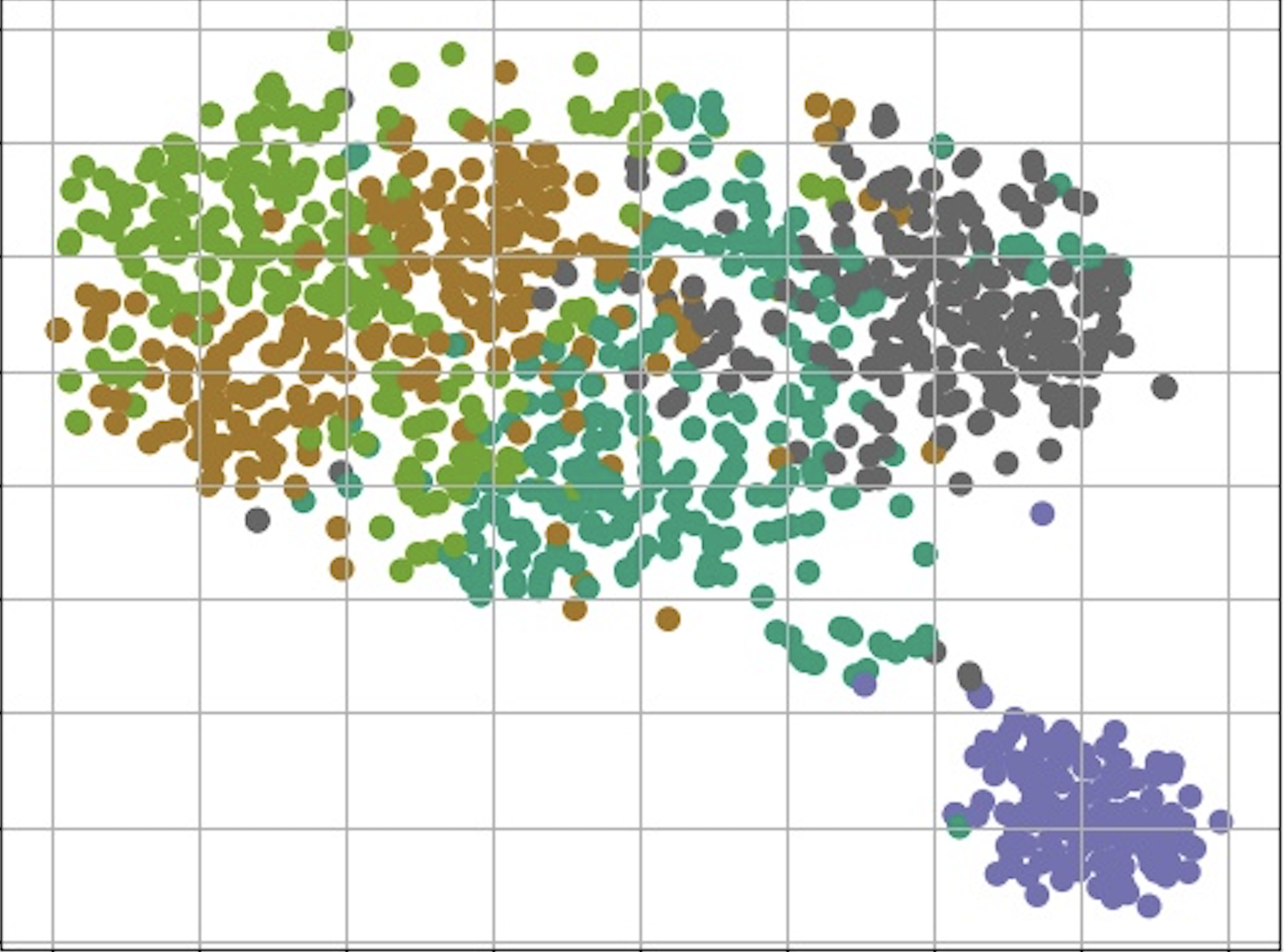}
    \includegraphics[width=0.49\textwidth]{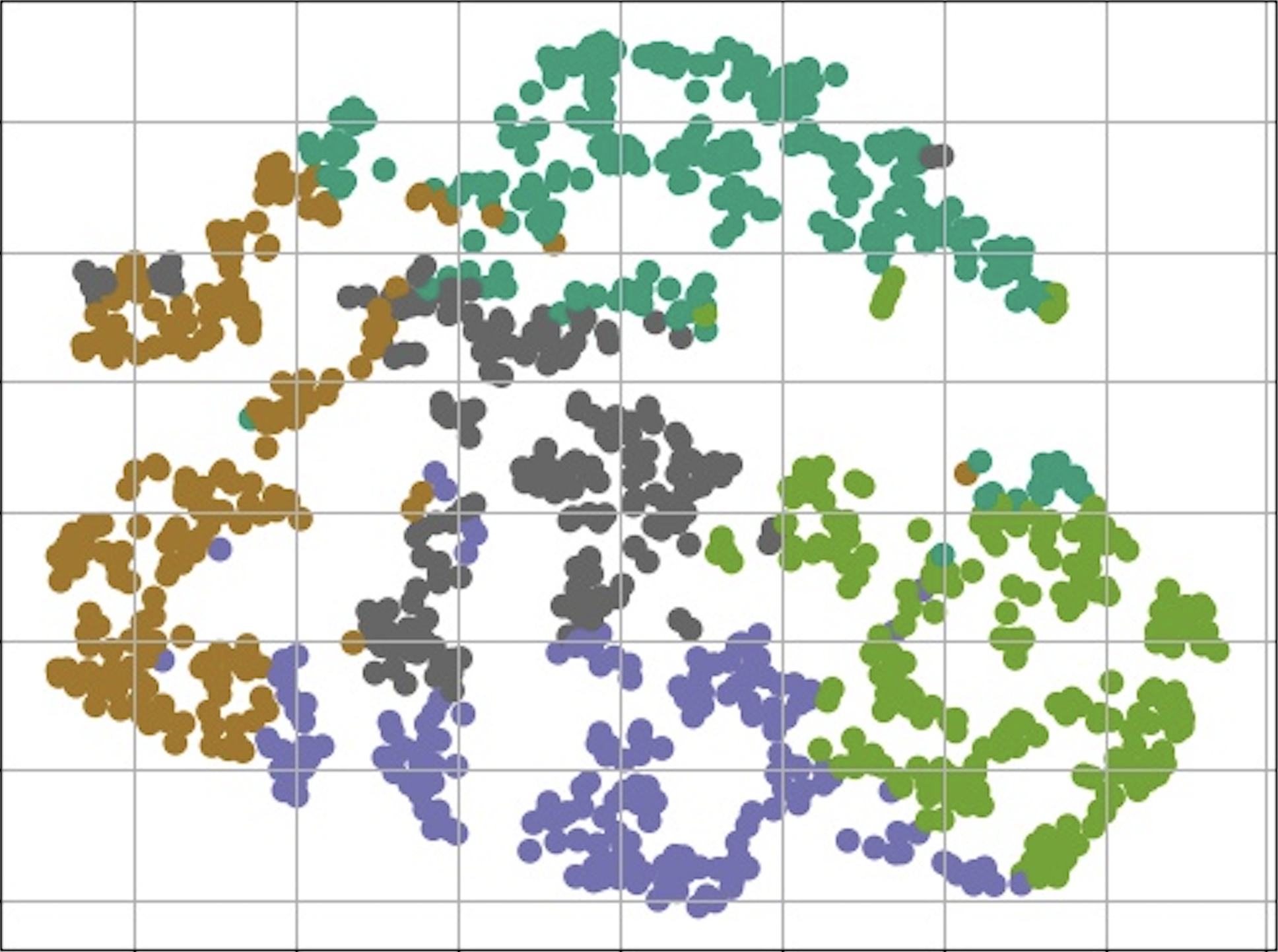}
    \caption{\textbf{T-SNE analysis} comparing representation embeddings. We take the ProtoRL \cite{yarats2021protorl} setup for learning representations in continuous control RL, where \modelname based information bottlenecks are applied on top of the learnt representations from ProtoRL. \textbf{Left}. Latent representations from ProtoRL with discrete prototypes. \textbf{Right}. Factorized latent representations with ProtoRL $+$ \modelname, that learns better structure in the representation space, when we apply a variational (Gaussian) information bottleneck followed by discrete information bottlenecks. \vspace{-5mm}}
\end{figure}

While the compositional structure can solely be achieved through the discretization bottleneck, we additionally add a gaussian information bottleneck \cite{alemi2016deep}.  This is added directly before the discretization function $\phi$.  encourage more parsimonious discrete representations.  Adding an information bottleneck to capture sufficient representations means that the we can achieve better compositionality by using \textit{fewer} discrete codes.  Figure \ref{fig:kmeans} shows the learnt compositional structure in the latent embedding space extracted by $\modelname$, while no apparent structure exists in the latent space for a baseline without any bottleneck. 
Following the learnt embeddings, we then apply the VQ discretization bottleneck, with different grouping factors. To apply discretization bottleneck, we quantize the output of the projector layer into group-based discrete latent embedding. Concretely, instead of assigning each continuous embedding $\rvz_e$ to a single one discrete vector, we first divide each continuous state representation into $G$ different groups as $\rvz_e=\text{concat}(\rvc_1,\rvc_2,\cdots,\rvc_G)$, then we assign each segment $\rvc_i\in\mathbb{R}^{\frac{m}{G}}$ separately to a discrete vector $\rve\in \mathbb{R}^{L\times\frac{m}{G}}$ using a nearest neighbour look-up: $\rve_{\rvo_i}=\textsc{discretize}(\rvc_i), \quad \text{ where }\quad \rvo_i=\operatorname{argmin}_{j \in \{1,\dots,L\}} ||\rvc_{i}-\rve_j||$, where $L$ is the size of the discrete latent space (i.e., an $L$-way categorical variable). After that, we concatenate all segments to obtain the discrete embedding $\rvz_{q}=\textsc{concatenate}(\textsc{discretize}(\rvc_1), \cdots,\textsc{discretize}(\rvc_G))$
This process results in compositionality of latent representation with an information bottleneck.

\begin{figure*} 
\centering
\subfigure{
\includegraphics[
width=0.33\textwidth]{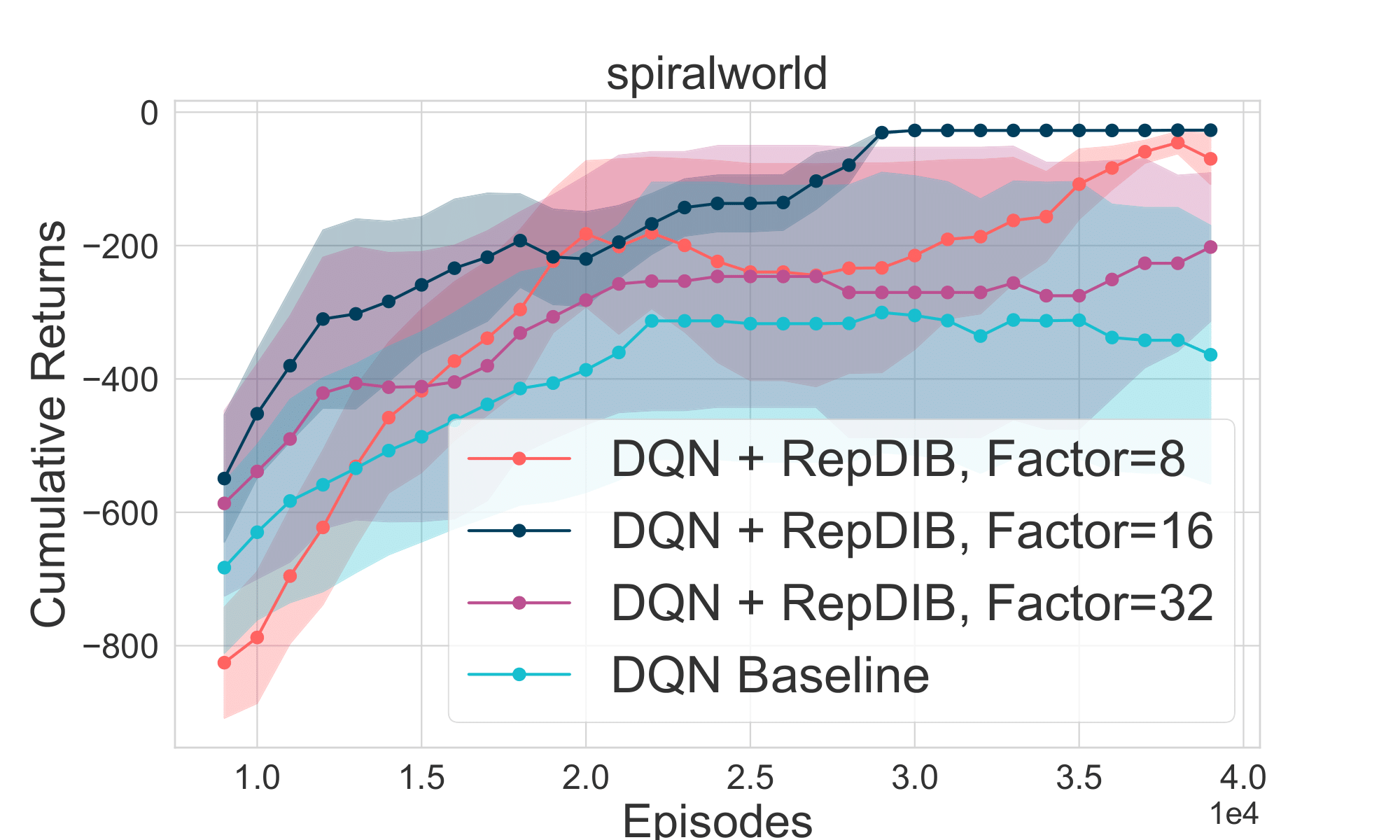}
}
\hspace{-0.8cm}
\subfigure{
\includegraphics[
width=0.33\textwidth]{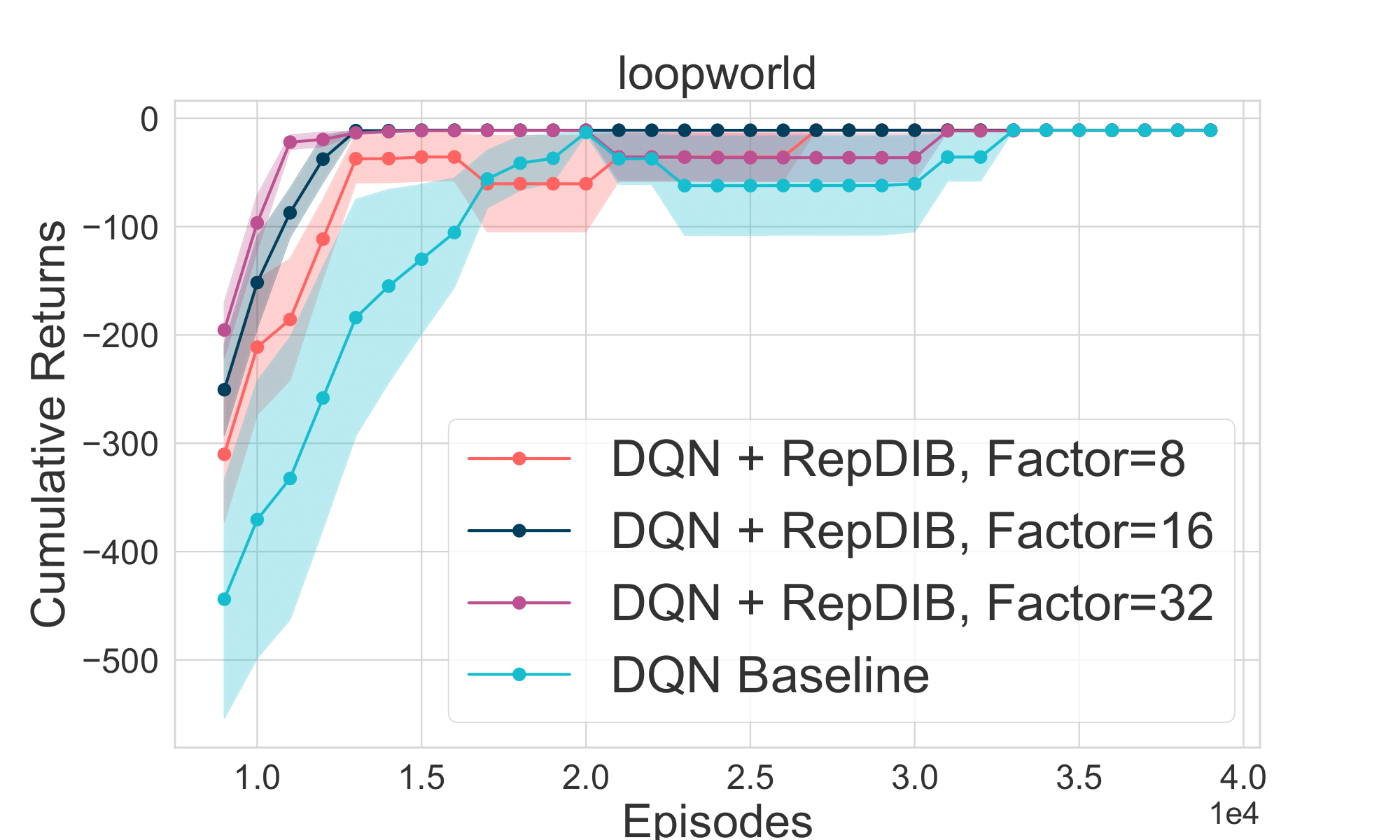}
}
\hspace{-0.8cm}
\subfigure{
\includegraphics[
width=0.33\textwidth]{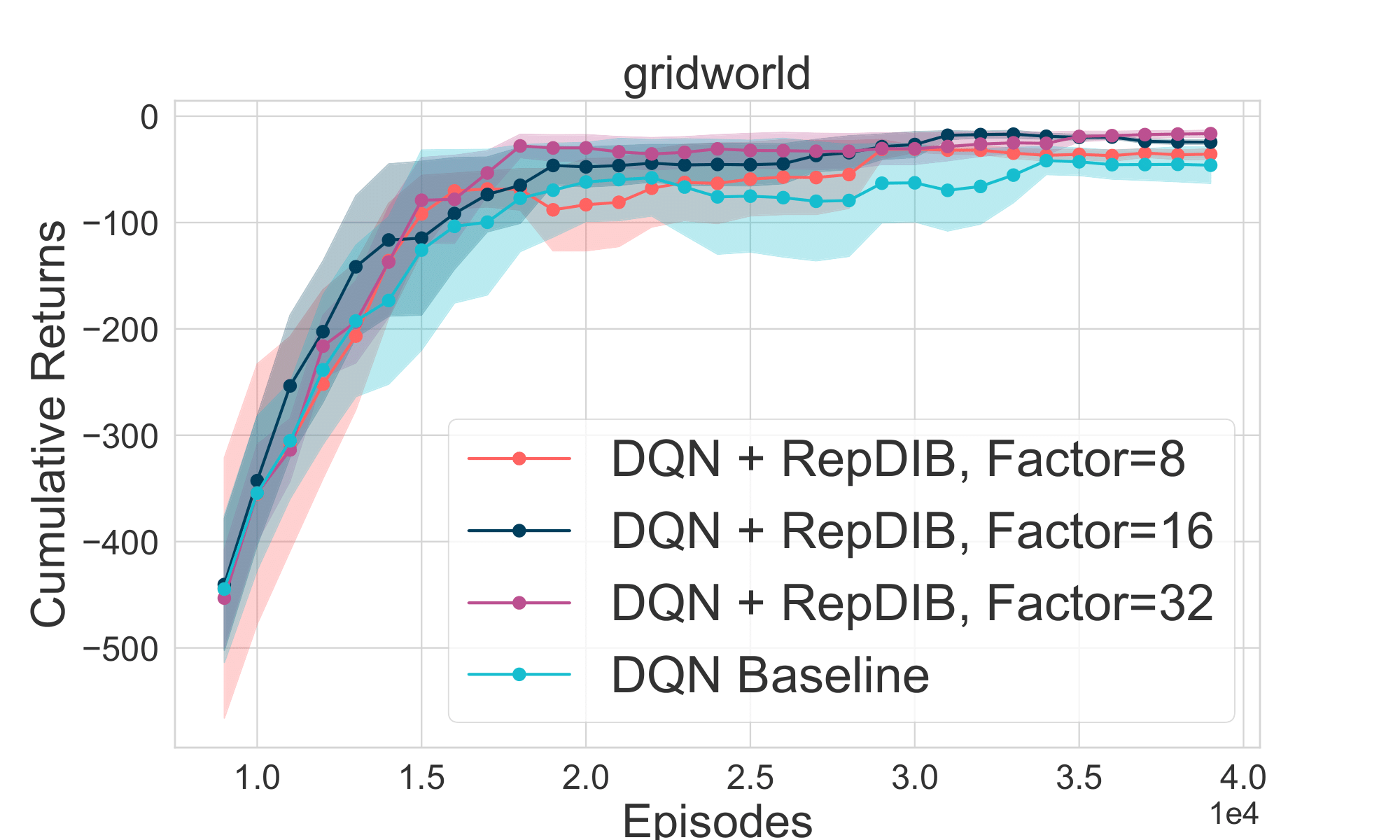}
}
\caption{\textbf{Performance} comparison on 3 different maze navigation tasks, with $\modelname$, using different factors $8, 16, 32$ in the learnt representation, integrated on a baseline DQN agent.}
\label{fig:maze_results}
\end{figure*}

\textbf{$\modelname$  Implementation Details}. We provide technical details of how our approach can be implemented on any existing self-supervised reinforcement learning objective (Figure~\ref{fig:archit}). To enable factorial structure in the representation space, we can integrate a vector quantization discretization bottleneck on top of any encoder that learns a latent state representation. Given an encoder that maps observations $o$ to latent representation $\phi(\cdot)$, we first use a variational information bottleneck (VIB) based on reparameterization, with a uniform Gaussian prior. We then quantize the continuous representation from an information bottleneck into discrete latent variables, generalizing vector quantization in VQ-VAE.

\vspace{-3mm}
\section{Experiments : Representations with Information Bottleneck}
\label{sec:experiments}
\vspace{-3mm}

We seek to understand the effectiveness of information compression in representations. We emphasize that $\modelname$ can be applied as a plug-in approach, on top of any existing framework that learns representations with a self-supervised objective. Through our experiments, we answer the following questions:

\textbf{Does inducing structure in representation space help with exploration?} We first demonstrate on simple toy tasks that learning representations with {\modelname} can induce a factorized structure that can lead to effective exploration. By using discrete information bottlenecks, we can recover the underlying discrete latent states while also learning factorized embedding space, that leads to better exploration in maze tasks when using a simple DQN agent.

\textbf{Does factorized representations with \modelname help learn task agnostic pre-trained representations, for better generalizaton capabilities? } We evaluate $\modelname$ on several complex control tasks using the URLB benchmark \cite{URLB} for testing generalization capabilities. In this setting, we pre-train representations in a reward-free approach on a given task, followed by fine-tuning on different downstream tasks. Most importantly,we show that sample efficiency of $\modelname$ can further be improved as a function of pre-training steps, where $\modelname$ can improve downstream performance with only few pre-training steps.

\textbf{Does information bottleneck help learn parsimoniuous representations to learn relevant representations in a real robot arm task, while ignoring background distractors? } To answer this, we use a real robot arm collected data, with a temporal background structure, where there is background noise from lightnings, TV and video. In this setting, we show that $\modelname$ can capture the relevant factors of variation and ignore irrelevant  distractors through the use of information and discretization bottleneck.



\textbf{Does bottleneck representations help in sequence modelling problem from offline datasets?} We evaluate $\modelname$ in the offline Atari benchmark where environment observations consist of additional exogenous information, using the Decision Transformer \cite{NEURIPS2021DT}, and find that pre-trained representations with $\modelname$ can learn robust representations ignoring distractors.

\textbf{What is the impact of VQ information bottleneck to extract unimodal and fuse multi-modal representations?} We study $\modelname$ using existing human activity recognition based dataset in a multi-modal learning setting. We show that when fusing representations from single modality with information bottleneck, followed by compressing the resulting multi-modal representation, $\modelname$ helps achieve performance improvements compared to existing baselines with information bottleneck on multi-modal representations for activity recognition tasks.

\begin{figure}[!htbp]
\centering
\subfigure{
\includegraphics[
width=0.4\textwidth]{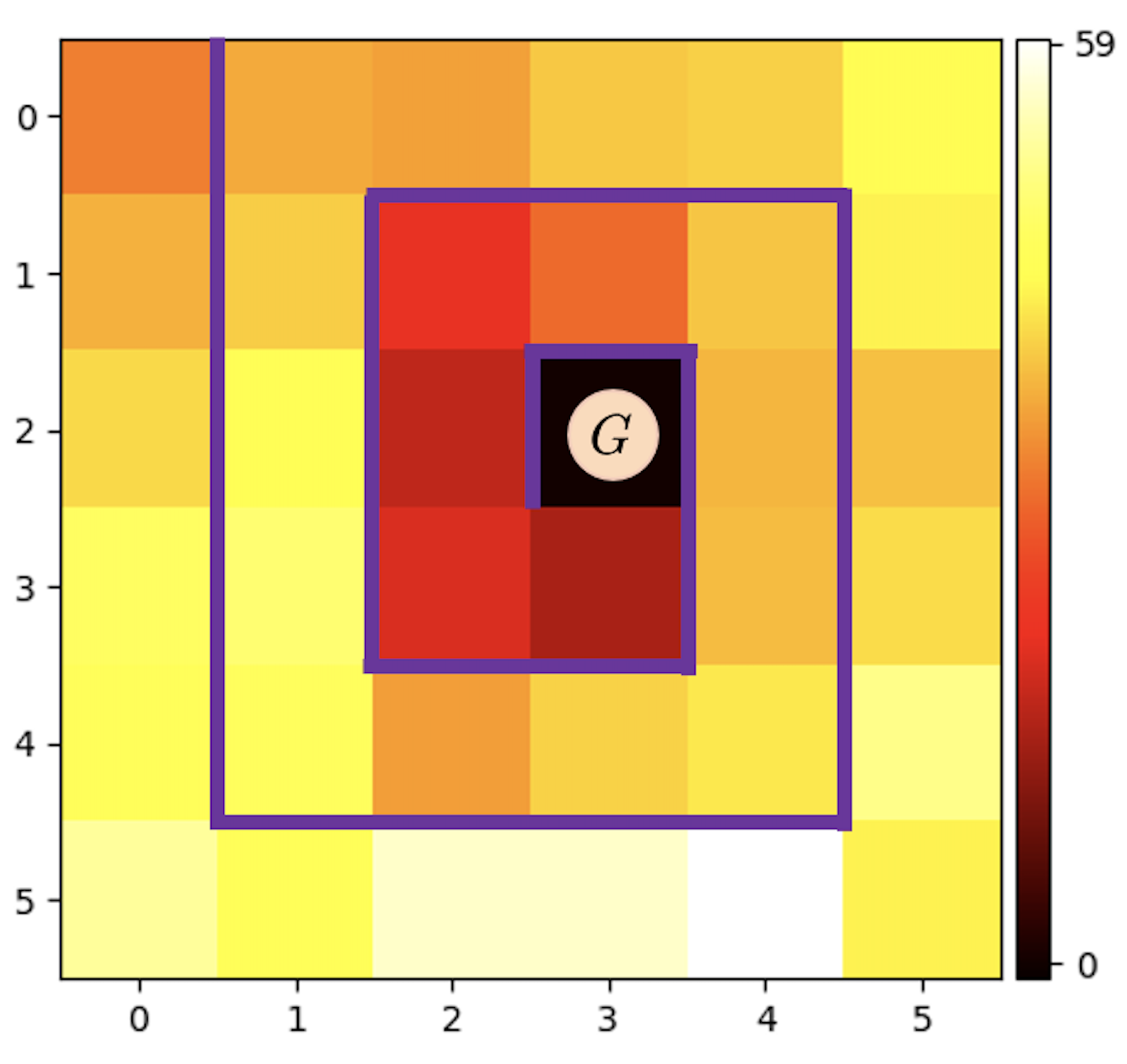}
}
\subfigure{
\includegraphics[
width=0.4\textwidth]{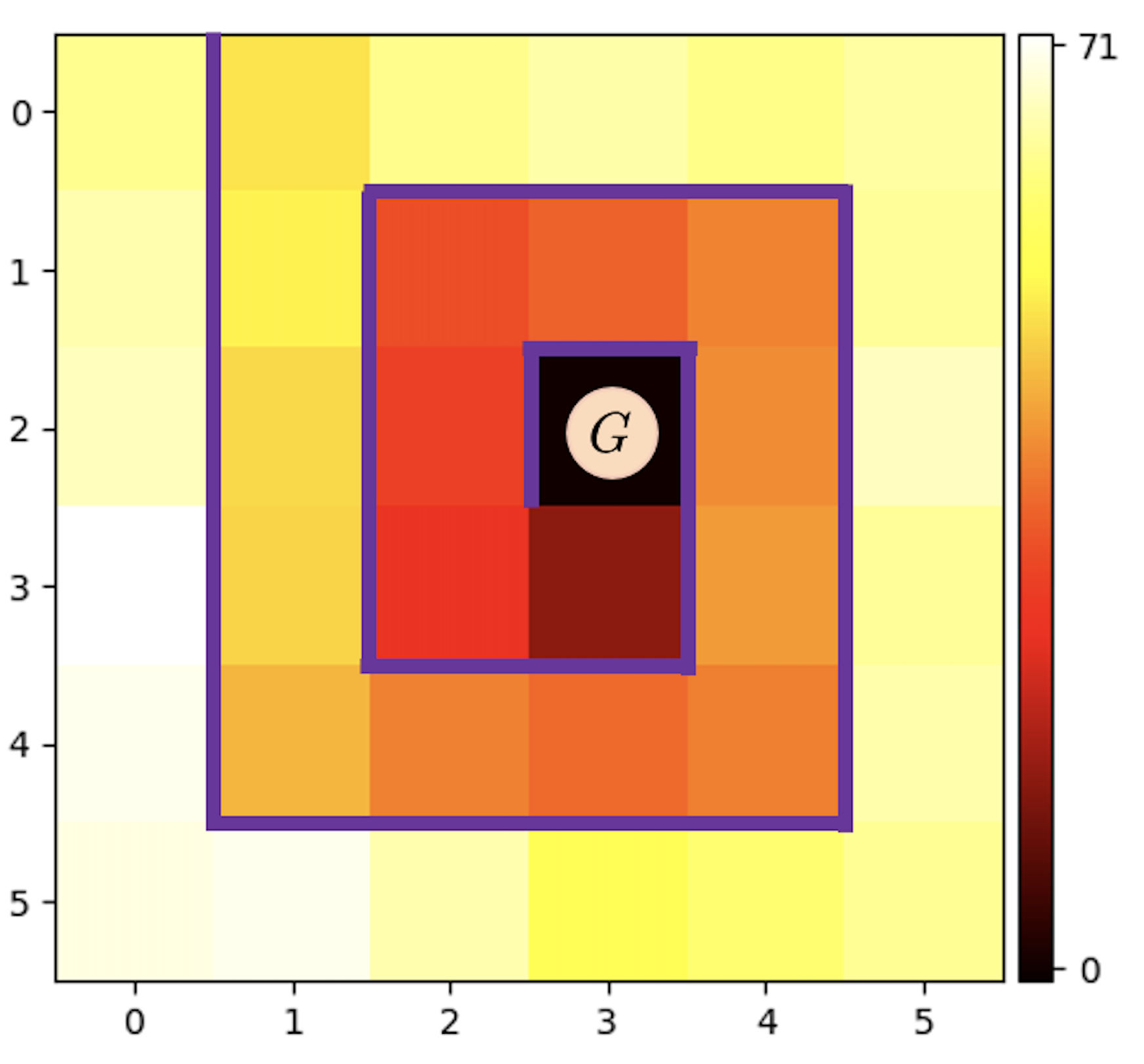}
}
\caption{\textbf{Relative Distance in Representation Space}. (Left): Baseline DQN Agent (Right): DQN Agent with $\modelname$. We show that $\modelname$ learns representations that can better capture underlying topology of how the agent can move in the maze. Darkness of  position shows how similar the representation is to the point in the center.}
\label{fig:topology}
\end{figure}

\vspace{-3.5mm}
\subsection{Maze Navigation Tasks}
\vspace{-2mm}
 \begin{figure*}[htbp]
    \includegraphics[width=0.7\textwidth]{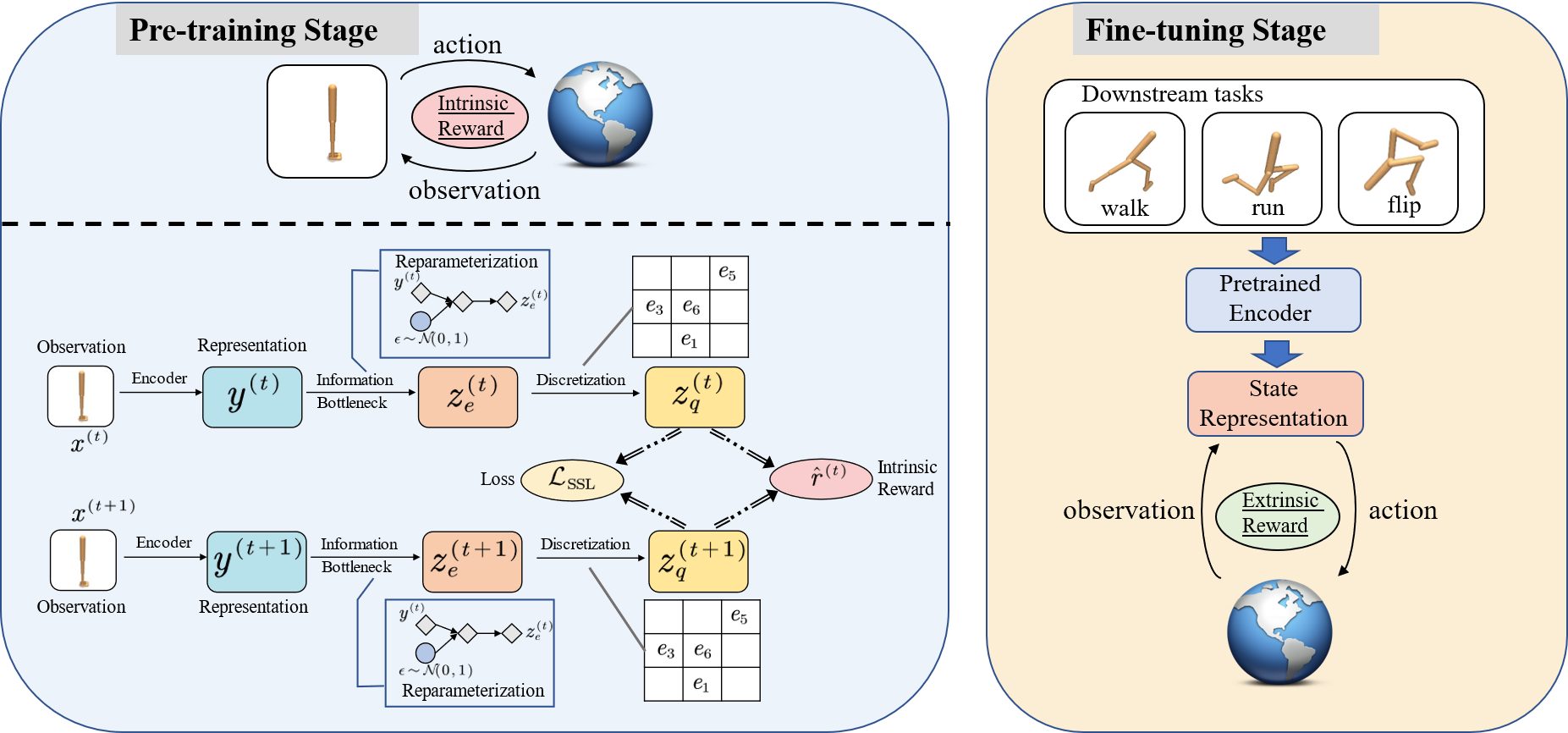}
    \caption{\textbf{Summary of $\modelname$} integrated on top of the ProtoRL baseline \cite{yarats2021protorl} for testing generalization capability in continuous control tasks from URLB benchmark \cite{URLB}. We find that $\modelname$ improves intrinsically-motivated exploration (left).  An information bottleneck is used to encourage the discrete codes to be parsimonious.  The reward-based fine-tuning stage remains unchanged when using $\modelname$ (right). }
    \label{fig:archit2}%
\end{figure*}

\textbf{Experiment Details}. We use three kinds of maze navigation tasks to evaluate the effectiveness of learning parsimonious representations with $\modelname$: \textit{GridWorld}, \textit{SpiralWorld}, \textit{LoopWorld}. We first learn the state representations on \textit{GridWorld} with data collected by a random policy, and then adapt the pre-trained representations to all these three tasks to learn the end-task policy.

\begin{figure}[!htbp]
\centering
\subfigure{
\includegraphics[
width=0.4\textwidth]{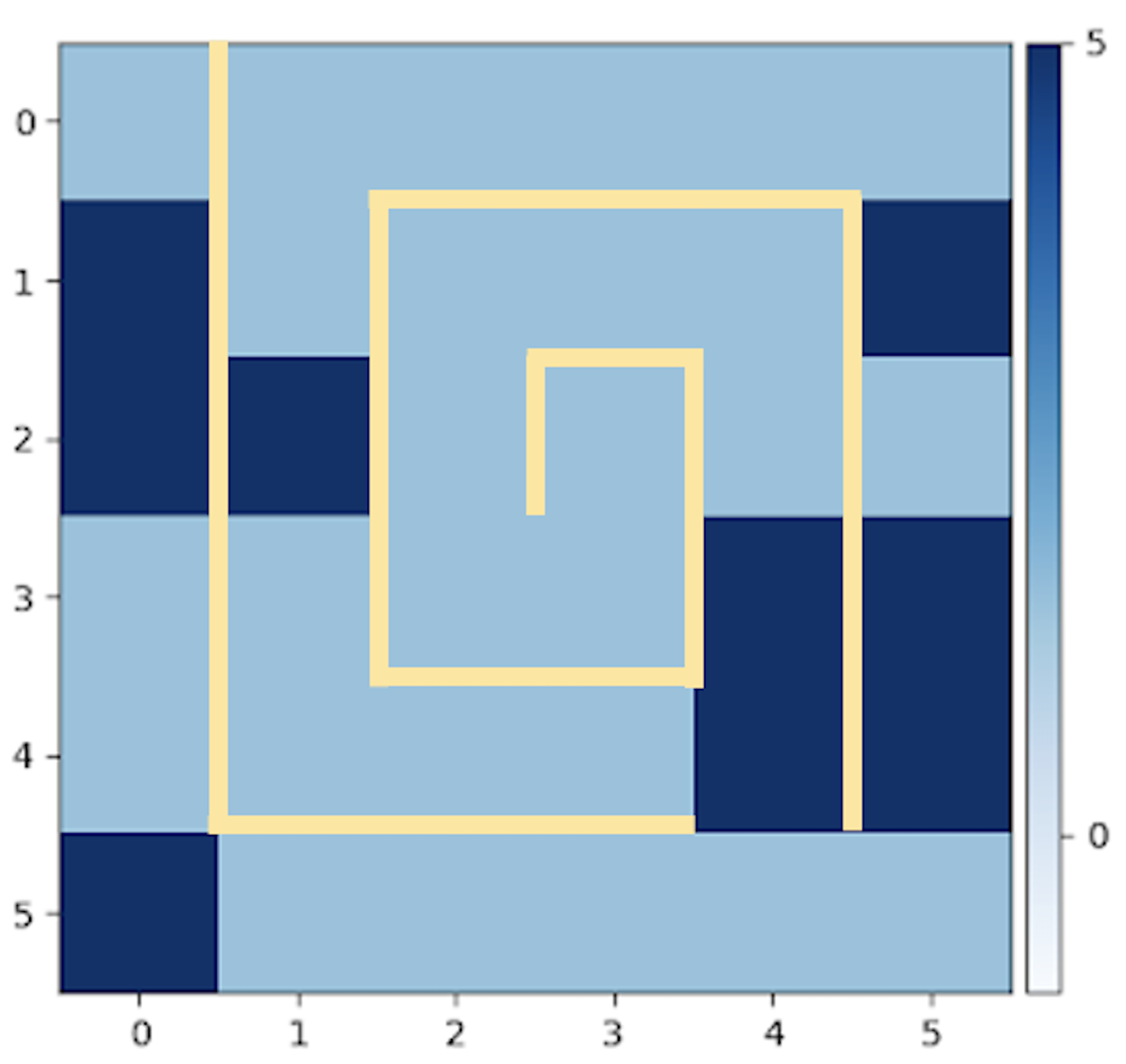}
}
\subfigure{
\includegraphics[
width=0.4\textwidth]{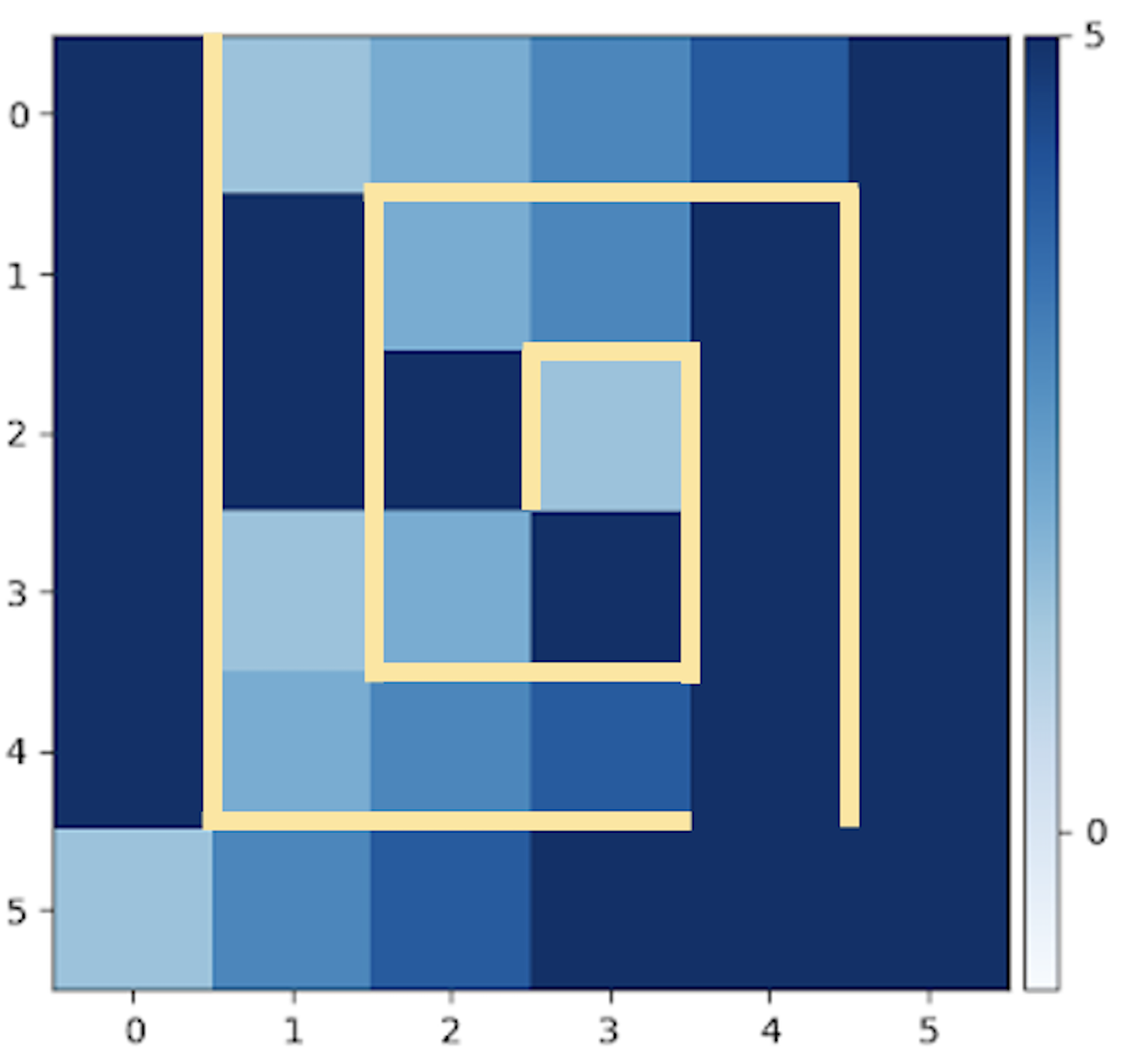}
}
\caption{\textbf{State Space Coverage} comparing DQN agent (left) and DQN agent with $\modelname$ (right). Factorized Representations with $\modelname$ leads to better state space coverage in reward-free exploration. \vspace{-5mm}}
\label{fig:coverage}
\end{figure}

\textbf{Experiment Results}. We study spiral and loop world maze navigation task, with a baseline DQN agent. During fine-tuning based on pre-trained representations from an empty gridworld, we simultaneously update both the representations and the DQN agent, given pixel based observations. We find that the induced factorized representation structure leads to better coverage of the state space, as demonstrated in Figure~\ref{fig:coverage} while also capturing topology of the maze in representation space as shown in figure \ref{fig:topology}. For the self-supervised objective for learning representations, we use DRIML \cite{MazoureCDBH20} for reward-free representation learning, followed by $\modelname$ integrated on top of the encoder. Experiment results, as in figure \ref{fig:maze_results} shows that the induced structure, based on different group factors of the discretization bottleneck, leads to improved performance compared to a baseline DQN agent.

\subsection{Generalization in Continuous Control}
\vspace{-2mm}

\begin{figure*}[htbp]
\centering
\hspace{-0.8cm}
\subfigure{
\includegraphics[
width=0.25\textwidth]{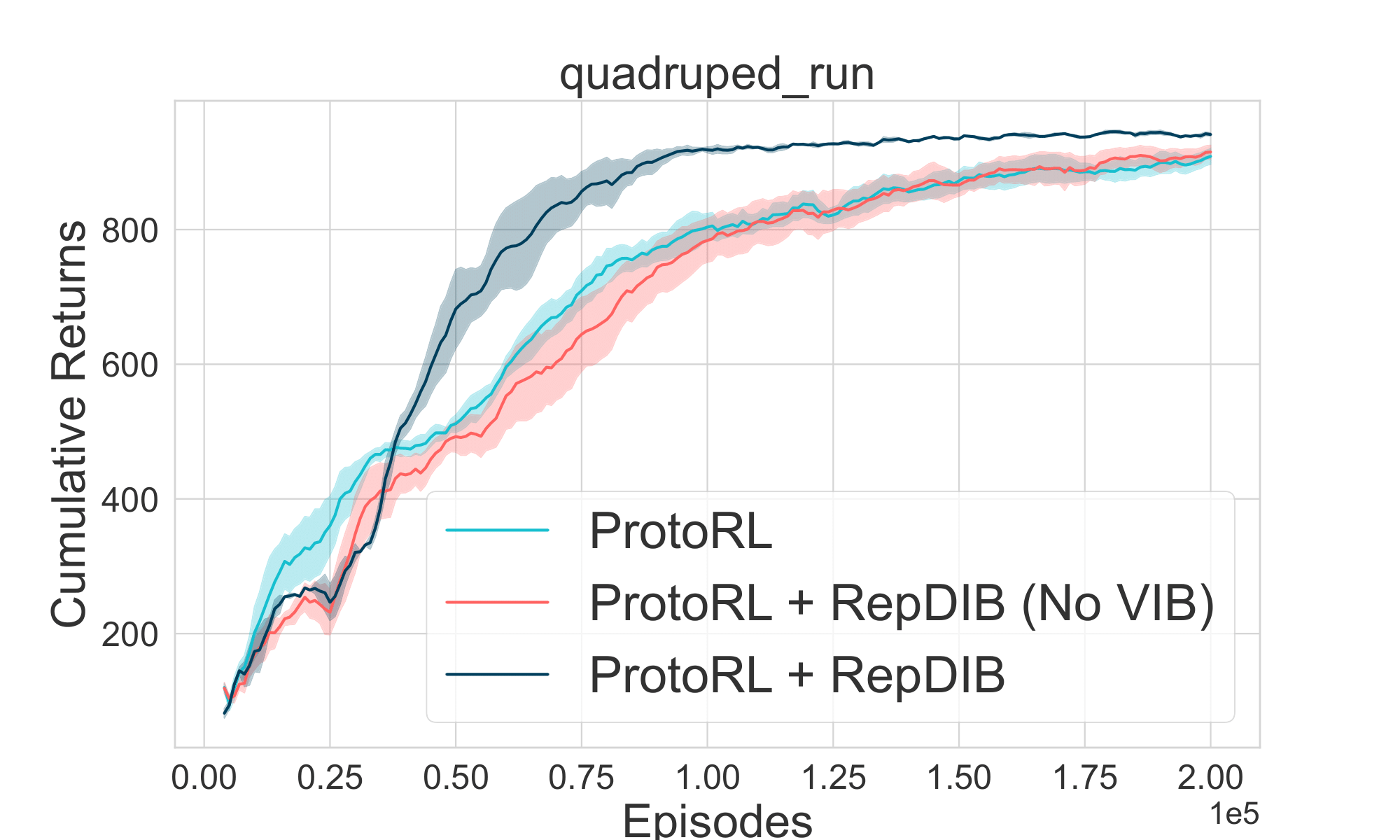}
}
\hspace{-0.8cm}
\subfigure{
\includegraphics[
width=0.25\textwidth]{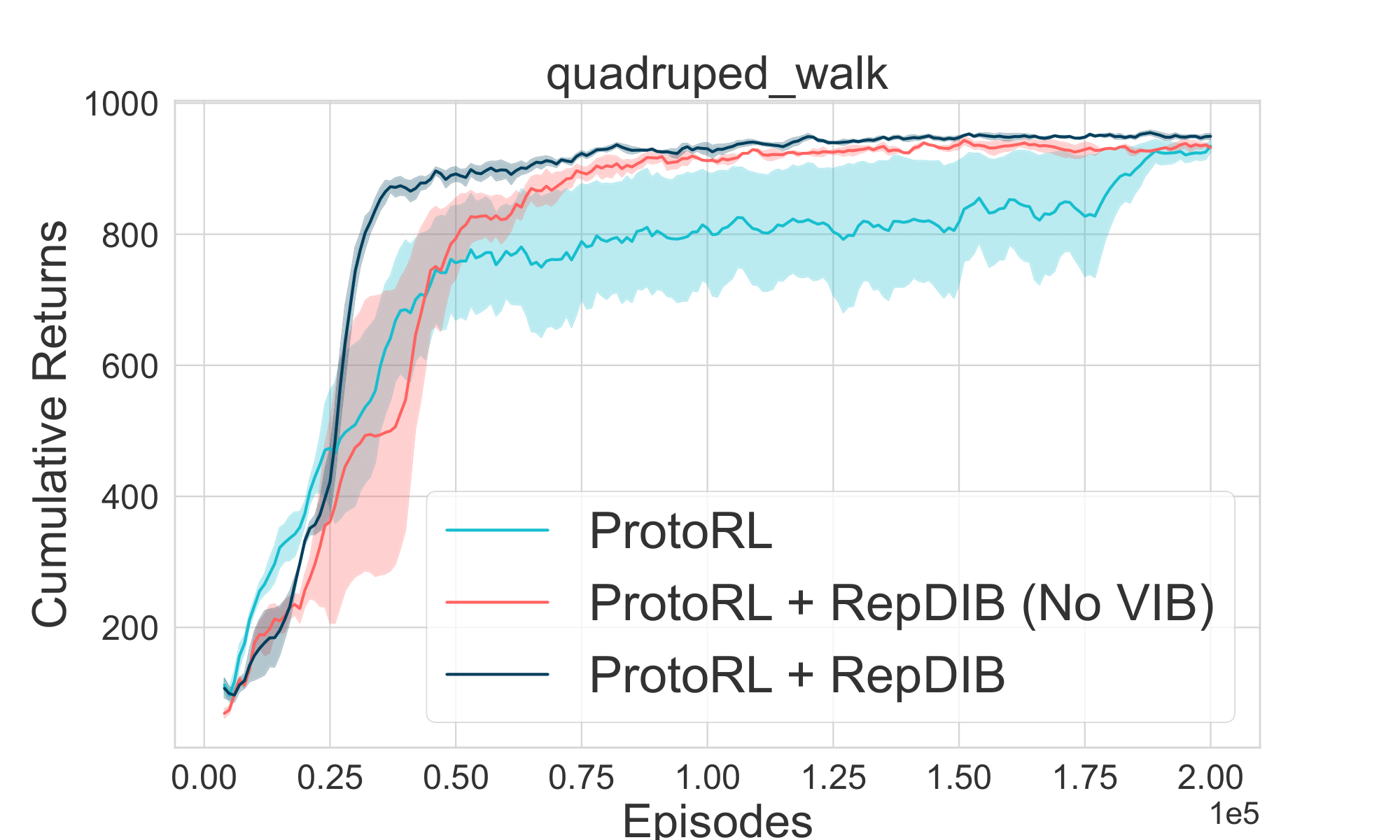}
}
\hspace{-0.8cm}
\subfigure{
\includegraphics[
width=0.25\textwidth]{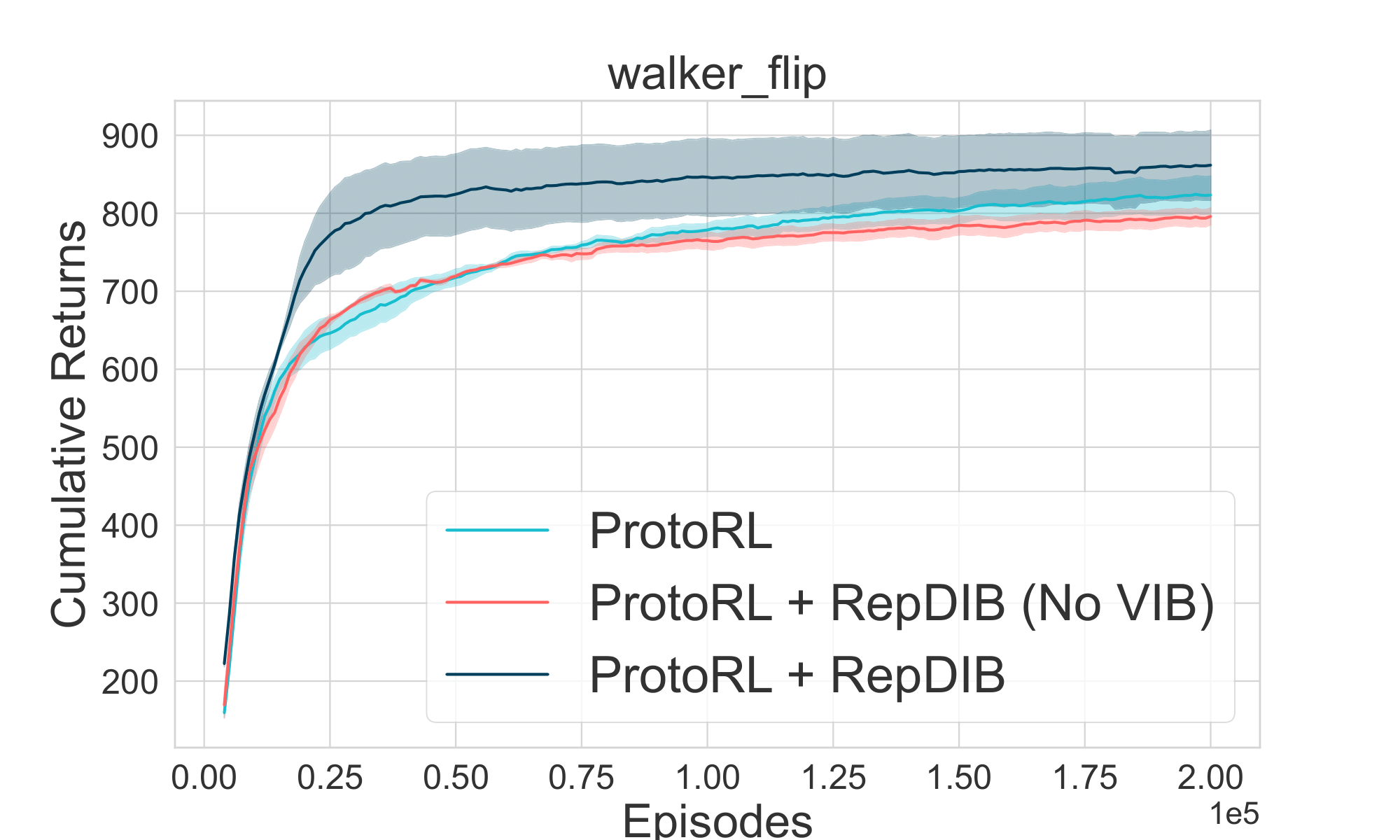}
}
\hspace{-0.8cm}
\subfigure{
\includegraphics[
width=0.25\textwidth]{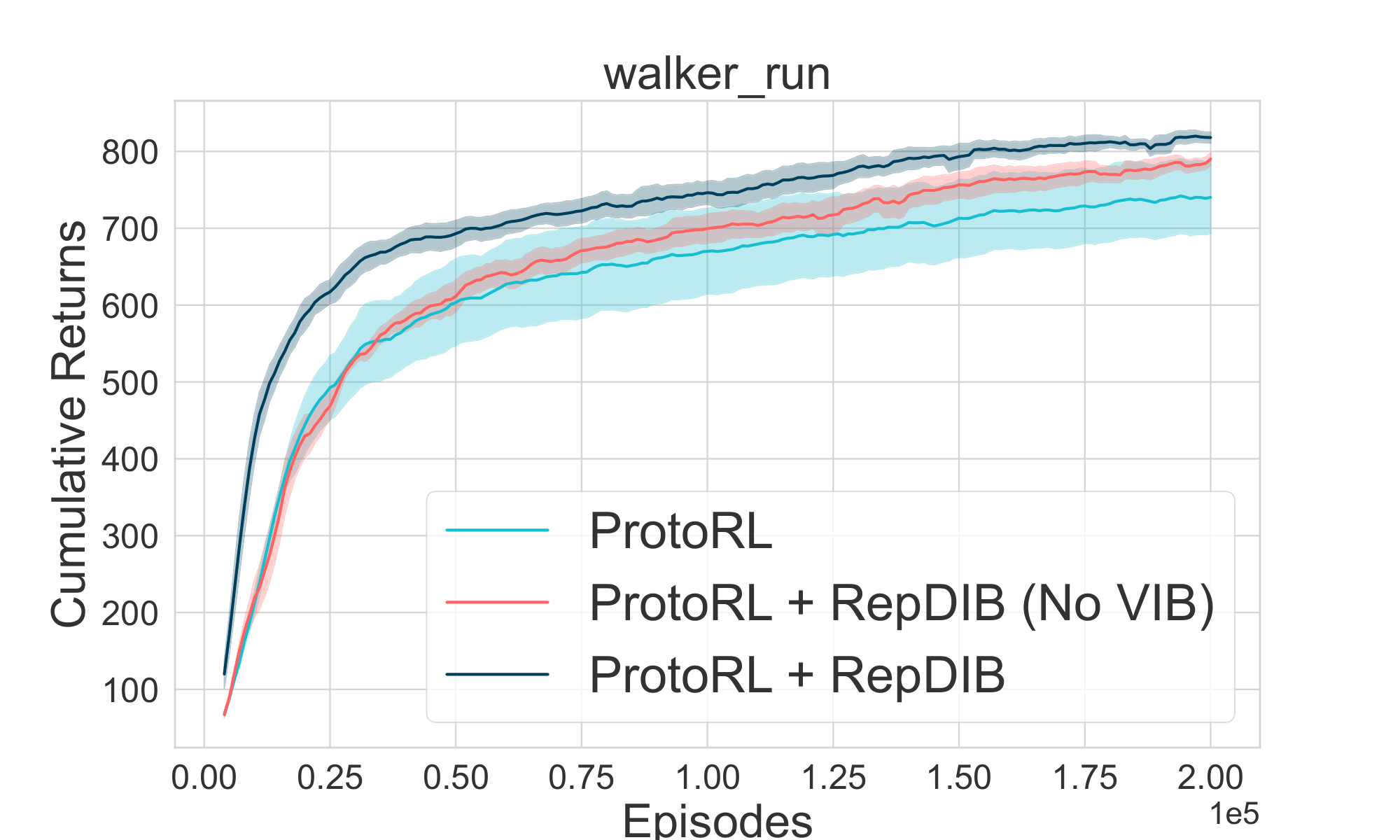}
}
\caption{\textbf{Fine-Tuning performance} on different domains, with pre-trained representations learnt with $\modelname$ and comparison with ProtoRL baseline. \vspace{-4mm} }
\label{fig:dmc_results_1}
\end{figure*}

$\modelname$ is then evaluated on a range of continuous control tasks with visual observations. We integrate $\modelname$ on the state of the art Proto-RL baseline \cite{yarats2021protorl}, as shown in figure \ref{fig:archit2} which has been shown to learn good representations from pre-training, for better fine-tuning performance. We follow the experiment setup from the URLB benchmark \cite{URLB}, and explain our experiment setup below, comparing $\modelname$ with a baseline Proto-RL agent, since it has been shown to outperform other baselines learning self-supervised representations. 

\begin{figure*}[htbp]
\centering
\subfigure{
\hspace{-0.3cm}
\includegraphics[
width=0.49\textwidth]{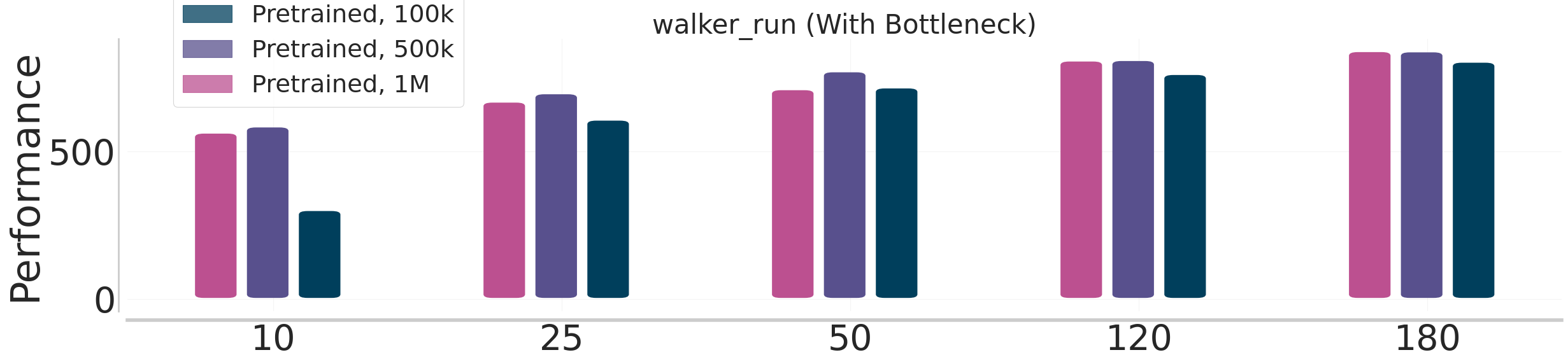}
}
\subfigure{
\includegraphics[
width=0.49\textwidth]{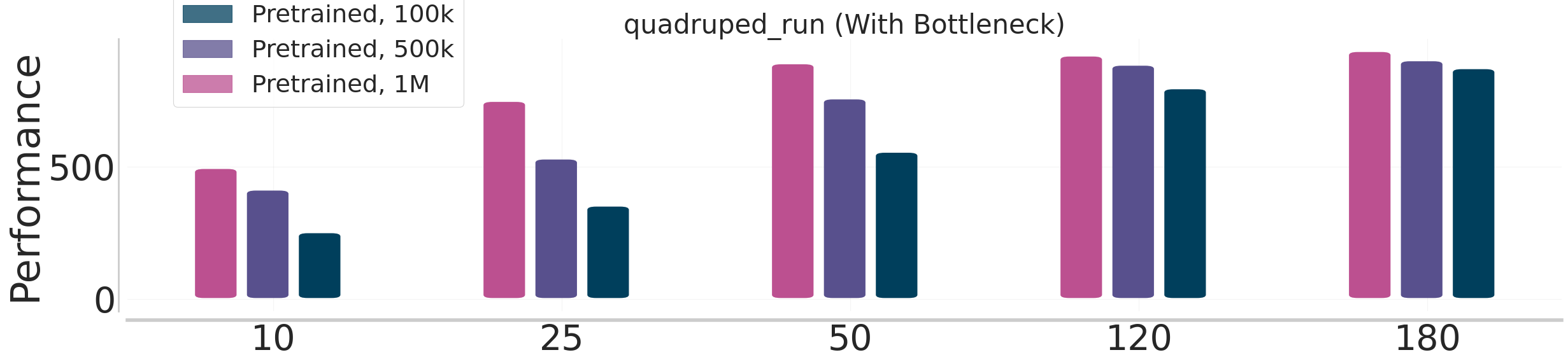}
}\\
\subfigure{
\hspace{-0.3cm}
\includegraphics[
width=0.49\textwidth]{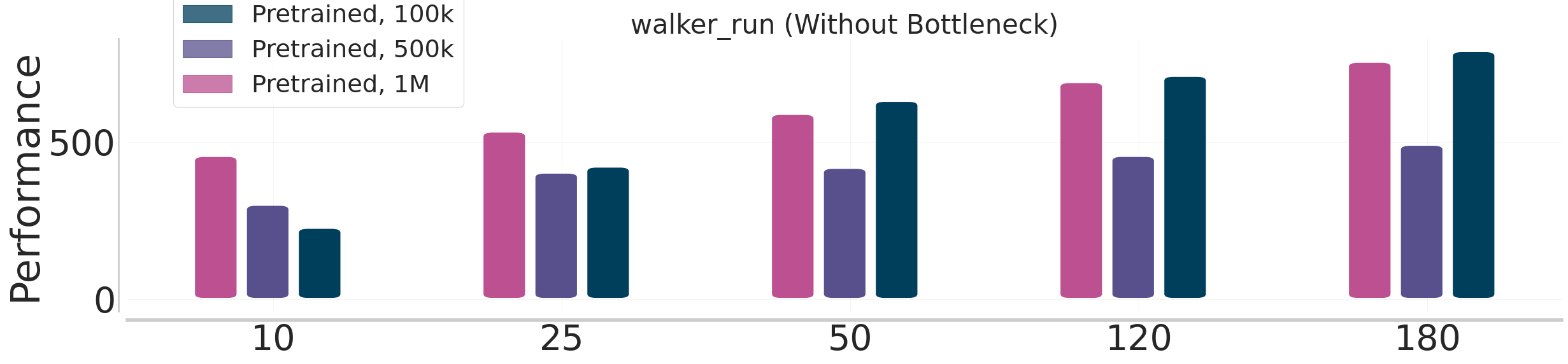}
}
\subfigure{
\includegraphics[
width=0.49\textwidth]{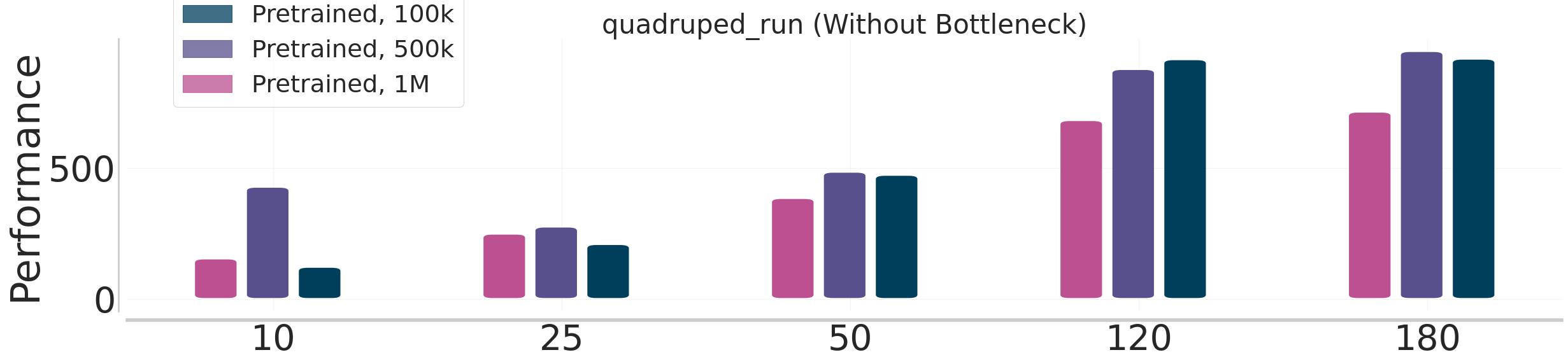}
}
\caption{\textbf{Downstream impact} of varying the number of pre-training steps ($100K$, $500K$ or $1M$ timesteps). X-axis shows different fine-tuning steps. We see that with the use of VIB and a discretiztion bottleneck, there is a gradual improvement in performance (\textbf{top row}); however the performance during fine-tuning can degrade as a function of pre-training steps when a $\modelname$ based bottleneck is not used in the baseline Proto-RL agent (\textbf{bottom row}).} 
\label{fig:pretrained_models}
\end{figure*}

\textbf{Experiment Details}. The key to Proto-RL is to learn a set of prototypical vectors and prototypes by projecting the embeddings onto clusters, referred to as prototypes. To ensure exploration, a latent state entropy distribution is optimized with an approximation, based on the learnt prototypical representations. This can form the basis for an intrinsic reward function, which ensures sufficient coverage in a pre-training phase in a task agnostic reward-free setting. In contrast to Proto-RL, the key to our approach of learning discret prototypes,e $\modelname$ is to ensure that a factorized structure is learnt in the latent representation. We refer to this as \textit{structured exploration}, since $\modelname$ exploits the use of information bottleneck and vector quantization to induce a factorial structure embeddings. 

We use a total of 12 continuous control tasks with varying difficulty (3 different domains with 4 different downstream tasks per domain): \textit{Walker}, \textit{Quadruped} and \textit{Jaco Arm}. The agent is pre-trained on a specific task in a given domain, and then adapts to the other downstream tasks within that domain. We follow the same experiment pipeline as in URLB benchmark~\cite{URLB}. We checkpoint the agent at 100k, 500k, 1M, 2M time-steps during pre-training, and then evaluate the adaptation ability of the method by adapting the pre-trained policy to downstream tasks.

\textbf{Pre-Training:} During the pre-training phase, we train $\modelname$ agent in a task agnostic reward free setting. The goal here is to encourage agent to reach unseen regions to collect more diverse data, such that this can further help in learning better representations. For this, we follow a similar procedure as Proto-RL,  where the agent is trained to maximize coverage by estimating an approximation to the entropy of the latent state distribution \cite{yarats2021protorl}. In case of $\modelname$, instead of estimating entropy based on an unstructured representation, $\modelname$ utilize the factorization structure in the representation space, through the use of the information bottleneck followed by the discretization module. Therefore, $\modelname$ computes the intrinsic reward based on the discrete factorial embeddings for more efficient structured exploration.

\textbf{Fine-Tuning}.  In the second phase, the agent is fine-tuned to solve new tasks to test its generalization capability. We use the learnt representation to then collect a dataset, which is then used by any standard state based off-policy RL algorithm such as soft actor critic (SAC) \cite{SAC}. Since the compositional structure is mostly exploited during pre-training phase, during fine-tuning we freeze the learnt representation to study the effectiveness of reward free representation. 

\textbf{Experiment Results on Control Tasks}. Since Proto-RL is shown to outperform existing baselines, including random exploration DrQ \cite{DrQ}, curiosity based exploration ICM \cite{ICM} and unsupervised active pre-training (APT) \cite{APT}; in this work we mostly compare to the state of the art Proto-RL baseline. We provide comparisons for $\modelname$ including the variational information bottleneck and $\modelname$ which only includes the discretization bottleneck (denoted as $\modelname$ only). Having pre-trained the representation encoder with and without a bottleneck in reward-free setting, we test the fine-tuning performance of the RL algorithm based on the fixed and learnt representation. Figure~\ref{fig:dmc_results_1} demonstrates the significance of $\modelname$ algorithm. We find that the use of information bottleneck prior to the discretization, can significantly help to improve sample efficiency during fine-tuning. We further examine the significance of the information bottleneck with different KL weightings in Appendix  figure~\ref{fig:vib_kl_weightings}

\textbf{Fine-Tuning Performance as a Function of Pre-Trained Unsupervised Representation Steps}. Downstream performance should monotonically improve with more steps of pre-training. However, it has been found that downstream performance sometimes degrades with more pre-training steps and that this counter-intuitive failure mode is common to all of the most widely used unsupervised RL algorithms \cite{URLB}. We reproduced this phenomenon in our prototypical-RL baseline. We found that \modelname alleviates this problem, resulting in monotonic improvements in downstream performance with more pre-training steps (Figure \ref{fig:pretrained_models}).  

\vspace{-2mm}
\subsection{Offline Experiments with Exogenous Distractors}  
\vspace{-2mm}
\label{sec:offline_exo}

\textbf{Experiment Setup with Atari using Decision Transformer}. We first consider the reward-conditioned behavior cloning setup with decision transformers \cite{NEURIPS2021DT}, where the goal is to learn representations that can ignore noisy or background information not relevant to the task using $\modelname$.  We consider the 4 games considered in \cite{NEURIPS2021DT} (Pong, Breakout, Seaquest, Qbert), using offline dataset \cite{agarwal2020optimistic} for training. The model is trained using a sequence modeling objective to predict the next action given the past states, actions, and returns-to-go.

To add exogenous information to the observation space, we append a randomly sampled cifar10 \cite{Krizhevsky_2009_17719} image to each frame. We keep the cifar image fixed in an episode but use a different image across episodes. We first pretrain our convolutional encoder with multi-step inverse objective introduced in \cite{lamb2022guaranteed}. We then train the Decision Transformer for action prediction keeping the convolutional encoder fixed. For the proposed approach, we discretize the output the encoder as described in method section \ref{sec:variational} before applying multi-step inverse objective.

\begin{table}[]
\caption{\textbf{Atari Results}. We compare the proposed \textsc{Multi-Step Inverse + \modelname} to \textsc{Multi-Step Inverse} on 4 atari game using the Decision Transformer setup. We can see that the proposed approach outperforms the baseline in all cases. Results averaged across 5 seeds. \vspace{-4mm}}
\label{tab:atari_results}
\scriptsize
\tablestyle{4pt}{1.2}
\resizebox{\columnwidth}{!}{%
\begin{tabular}{l| c c}
 \textsc{Game} & \textsc{Multi-Step Inverse} & \textsc{Multi-Step Inverse + \modelname} \\
 \shline
 \textsc{Pong} & \g{11.4}{2.653} & \highlight{\g{12.8}{2.561}} \\
 \textsc{Qbert} & \g{878.6}{745.146} & \highlight{\g{1100.0}{898.499}} \\
 \textsc{Breakout} & \g{19.8}{3.059} & \highlight{\g{41.8}{7.305}} \\
 \textsc{Seaquest} & \g{915.2}{126.368} & \highlight{\g{1058.4}{116.629}} \\
\end{tabular}
}
\end{table}

\textbf{Experiment Results}. Table \ref{tab:atari_results} summarizes the Atari results, where  \textsc{Multi-Step Inverse + $\modelname$} outperforms \textsc{Multi-Step Inverse} in all games thus showing the effectiveness of the VQ bottleneck. We use a discretization module with 32 factors for all the games. Additional results analysing the effect of the number of discretization factors is presented in appendix.\\

\begin{figure*}[!htb]
\centering
\subfigure{
\includegraphics[
trim=1cm 0cm 1cm 0cm, clip=true,
width=0.25\textwidth]{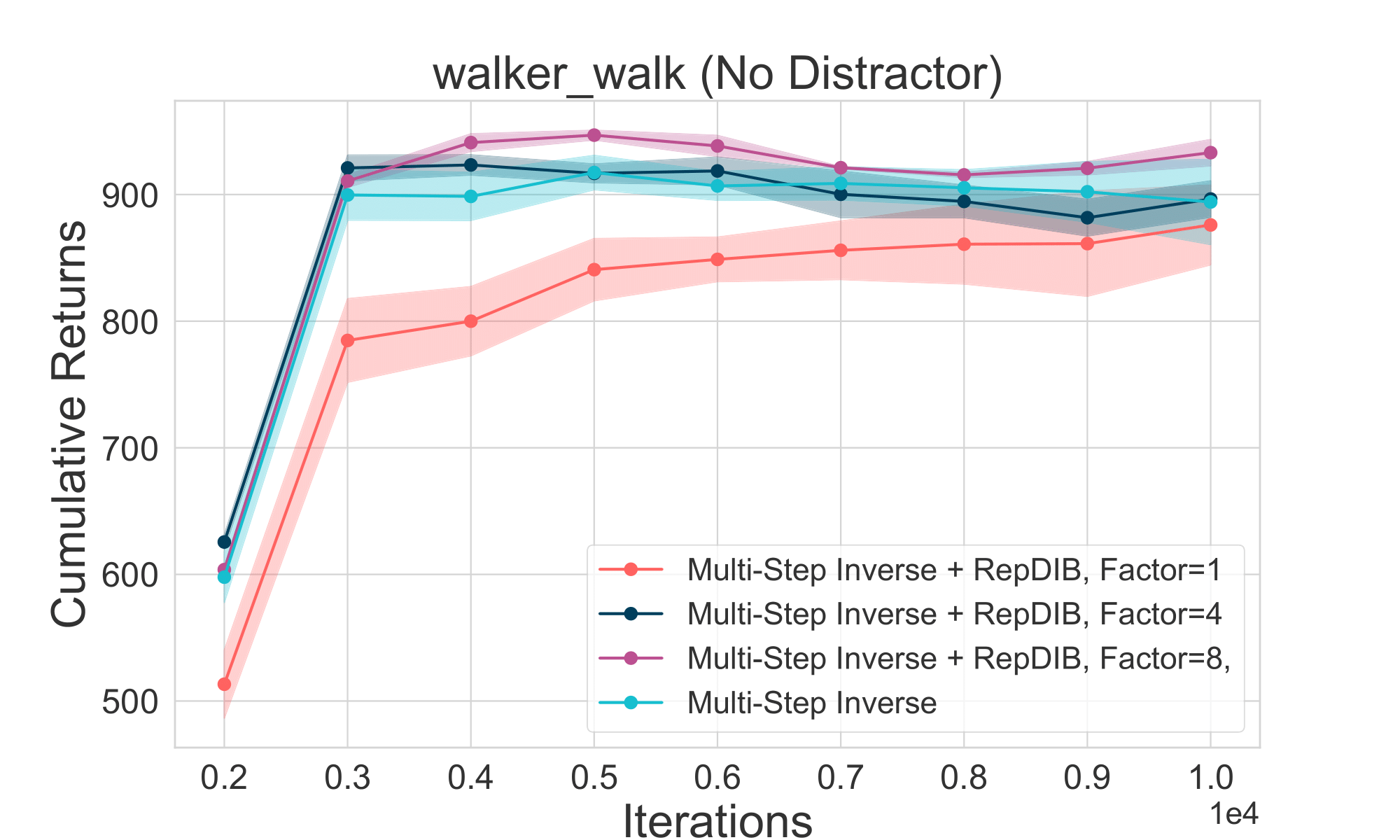}
}
\subfigure{
\includegraphics[
trim=1cm 0cm 1cm 0cm, clip=true,
width=0.25\textwidth]{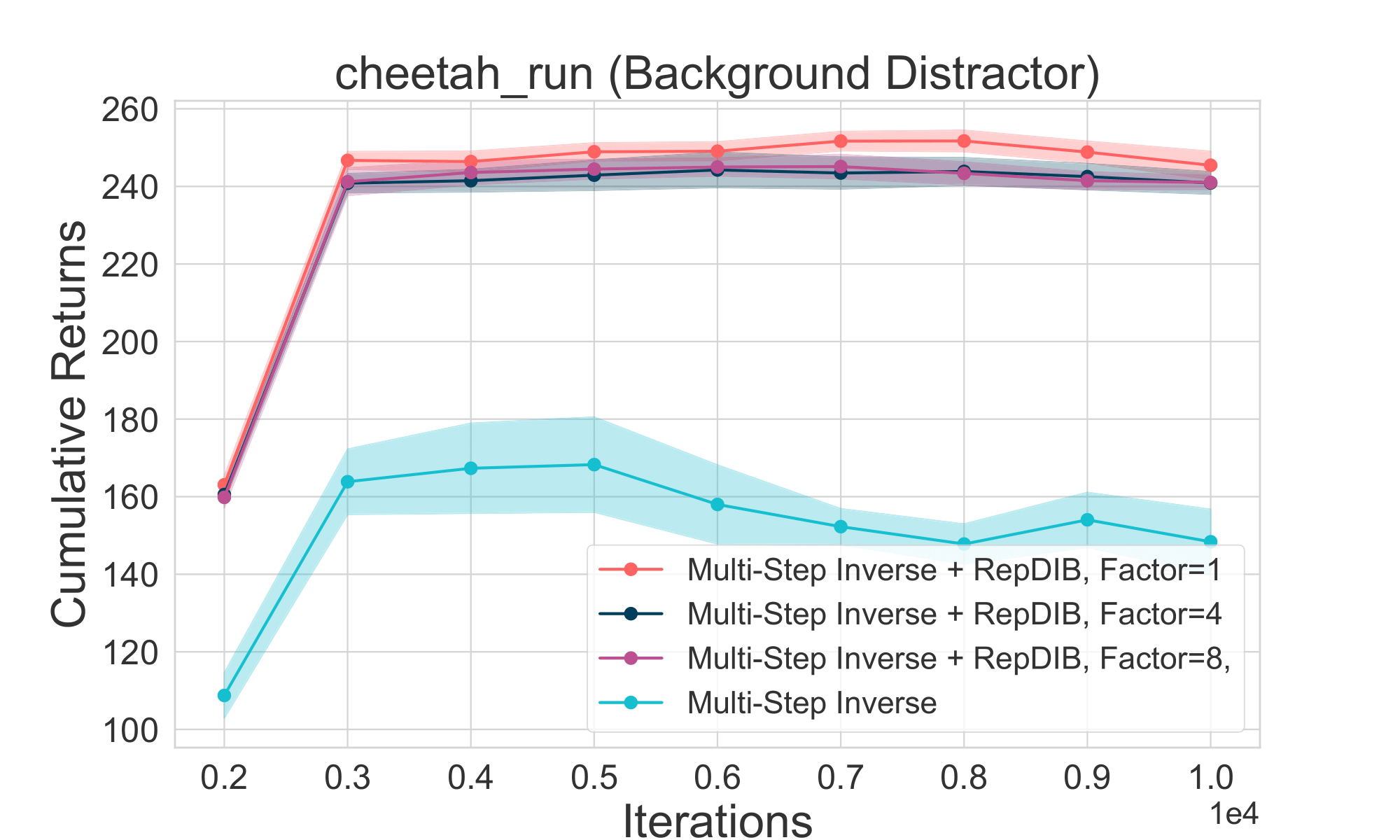}
}
\subfigure{
\includegraphics[
trim=1cm 0cm 1cm 0cm, clip=true,
width=0.25\textwidth]{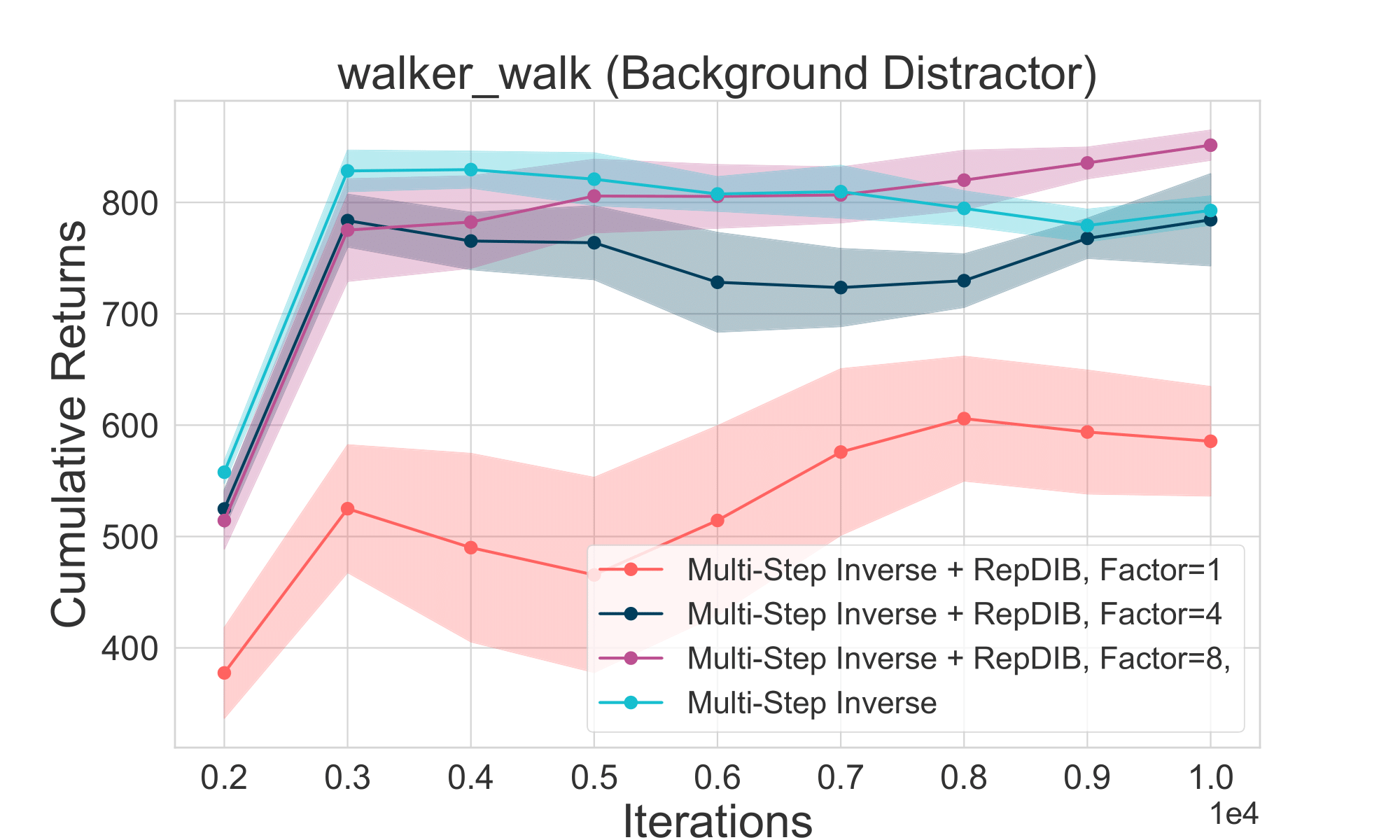}
}
\caption{\textbf{$\modelname$ can learn more robust representations} due to information bottleneck, in presence of background exogenous distractors, when using the offline visual control setup from \cite{vd4rl}. In contrast, performance is almost similar in settings without distractors. }
\label{fig:vd4rl}
\end{figure*}

\textbf{Experiments with Visual Offline RL.} We then consider the visual pixel-based offline dataset \cite{vd4rl} for control, where we learn representations using a \textsc{Multi-Step Inverse} model \cite{lamb2022guaranteed}. We consider two settings : with no visual background distractors and another where we add time correlated exogenous image distractors in the background. Figure \ref{fig:vd4rl} summarizes the results, where we find that in presence of exogenous image distractors, $\modelname$ can learn more robust representations during pre-training; whereas performance is similar in the setting without any additional exogenous distractors.

\paragraph{Comparisons with Other Information Bottleneck Approaches :} We now compare \modelname with several other bottleneck baselines in the pixel based offline RL setup. We follow the same experiment setup as described in section \ref{sec:offline_exo} and integrate information bottleneck approaches on top of three existing representation learning objectives, namely AC-State \cite{lamb2022guaranteed}, One step inverse dynamics \cite{pathak2017curiosity} and DRIML \cite{MazoureCDBH20}. We compare with \textbf{three different baselines} along with comparisons of variations of \modelname bottlneck. 

\begin{figure*}[!htbp]
\centering
\subfigure{
\includegraphics[
trim=1cm 0cm 1cm 0cm, clip=true,
width=0.25\textwidth]{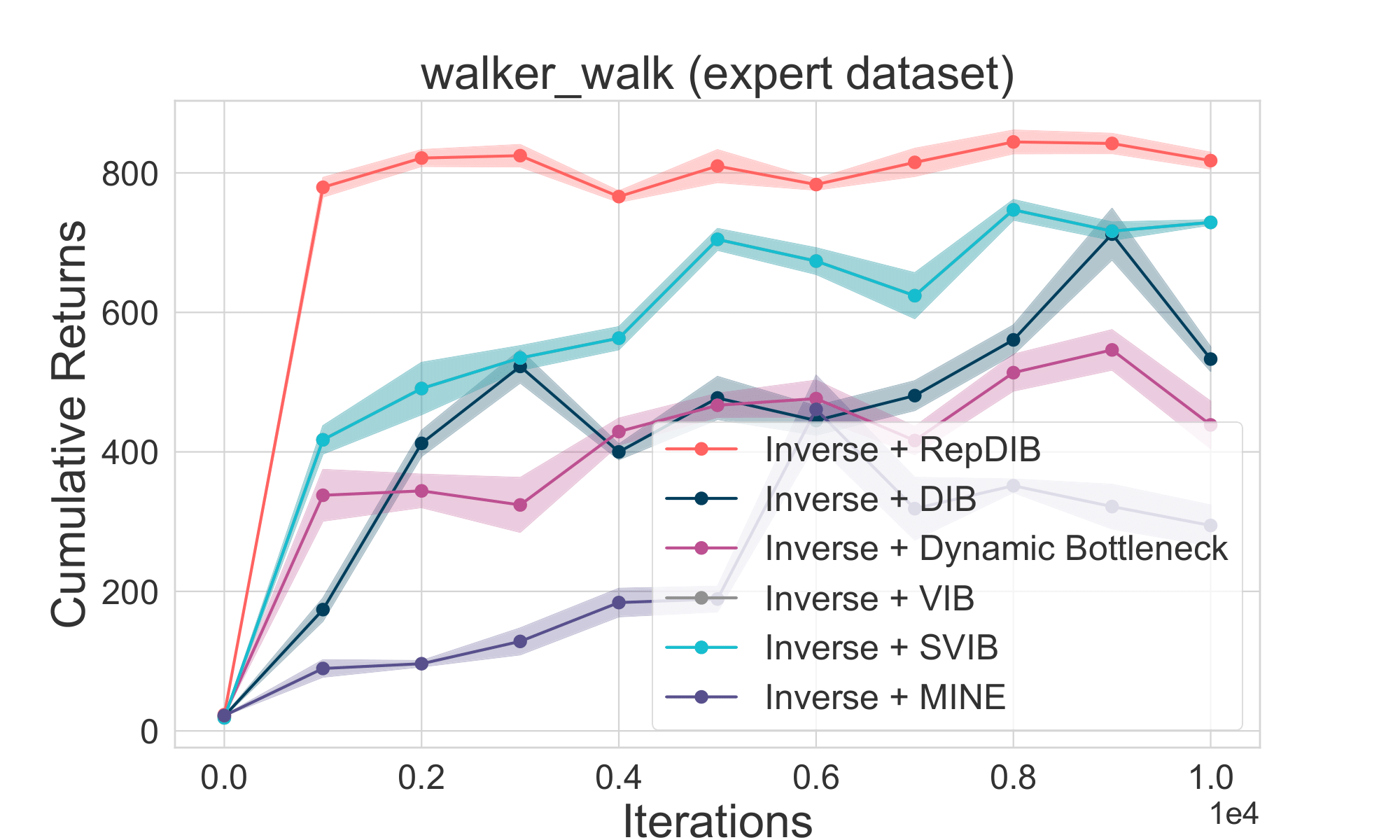}
}
\subfigure{
\includegraphics[
trim=1cm 0cm 1cm 0cm, clip=true,
width=0.25\textwidth]{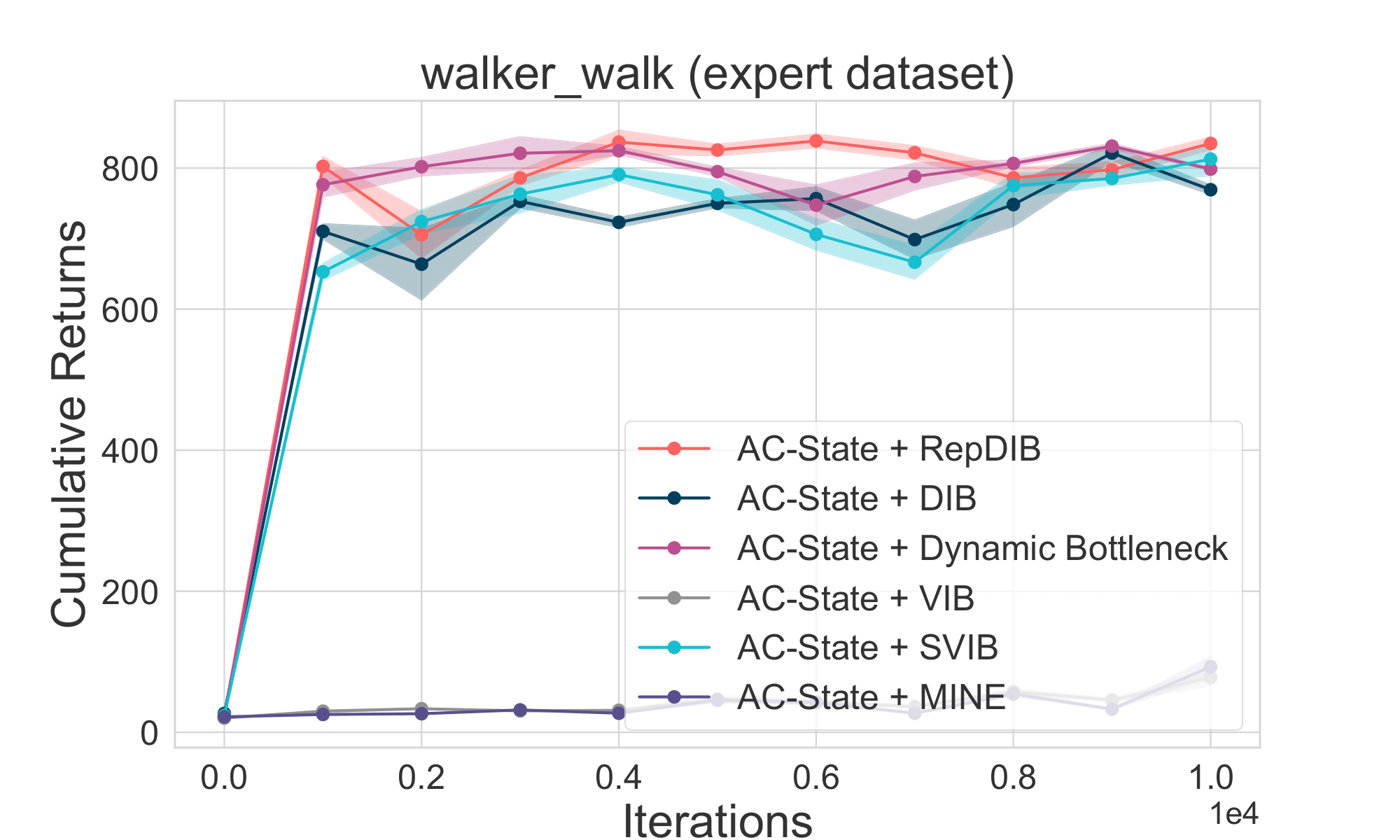}
}
\subfigure{
\includegraphics[
trim=1cm 0cm 1cm 0cm, clip=true,
width=0.25\textwidth]{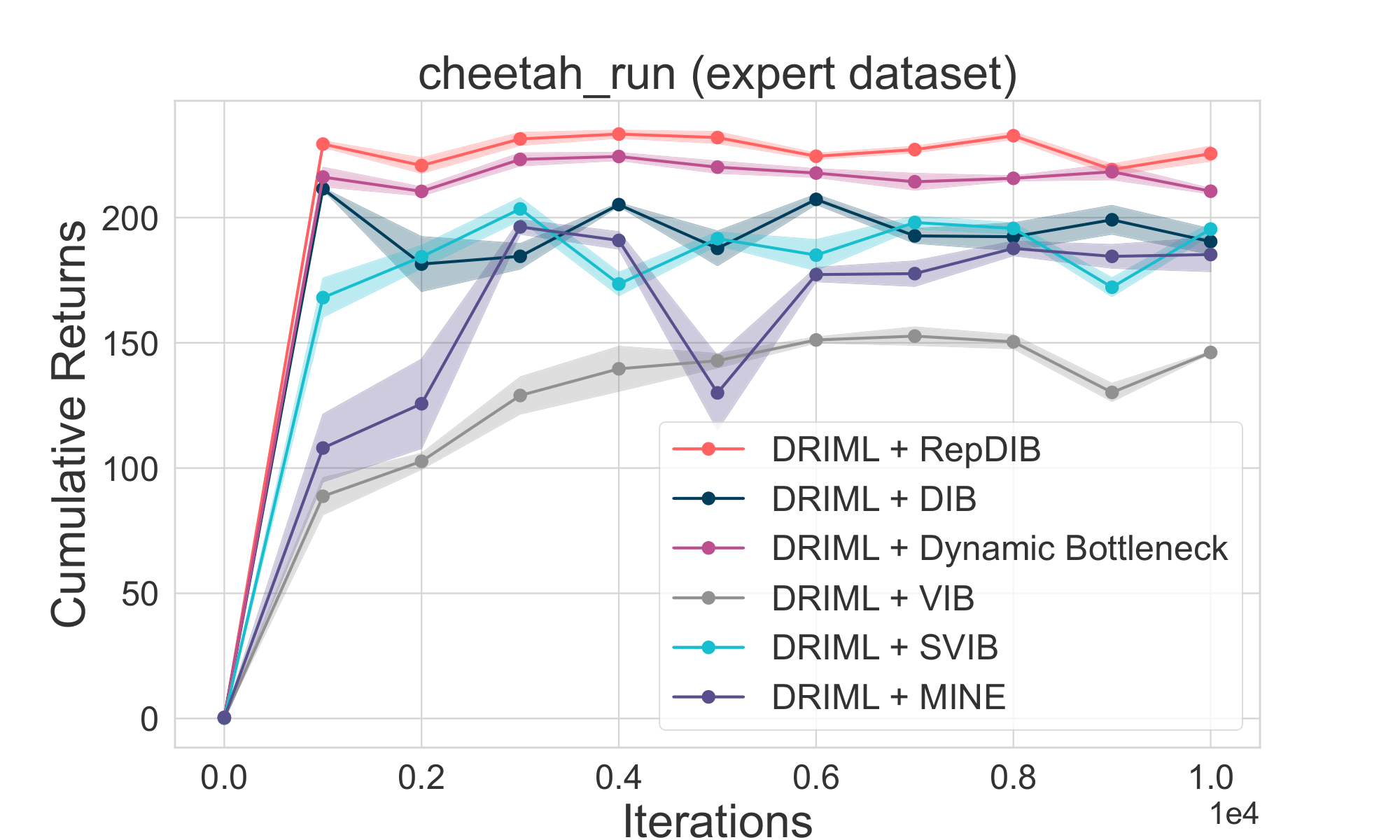}
}
\caption{\textbf{Time correlated exogenous images in the background.} Comparison of \modelname with other approaches based on information bottleneck based approximations in the offline RL setup. Following our previous results in the main, we now compare different bottleneck based approaches on top of existing representation learning objectives. }
\label{fig:rebuttal_comparison_bottlenecks_images}
\end{figure*}

\begin{figure*}[!htbp]
\centering
\subfigure{
\includegraphics[
trim=1cm 0cm 1cm 0cm, clip=true,
width=0.25\textwidth]{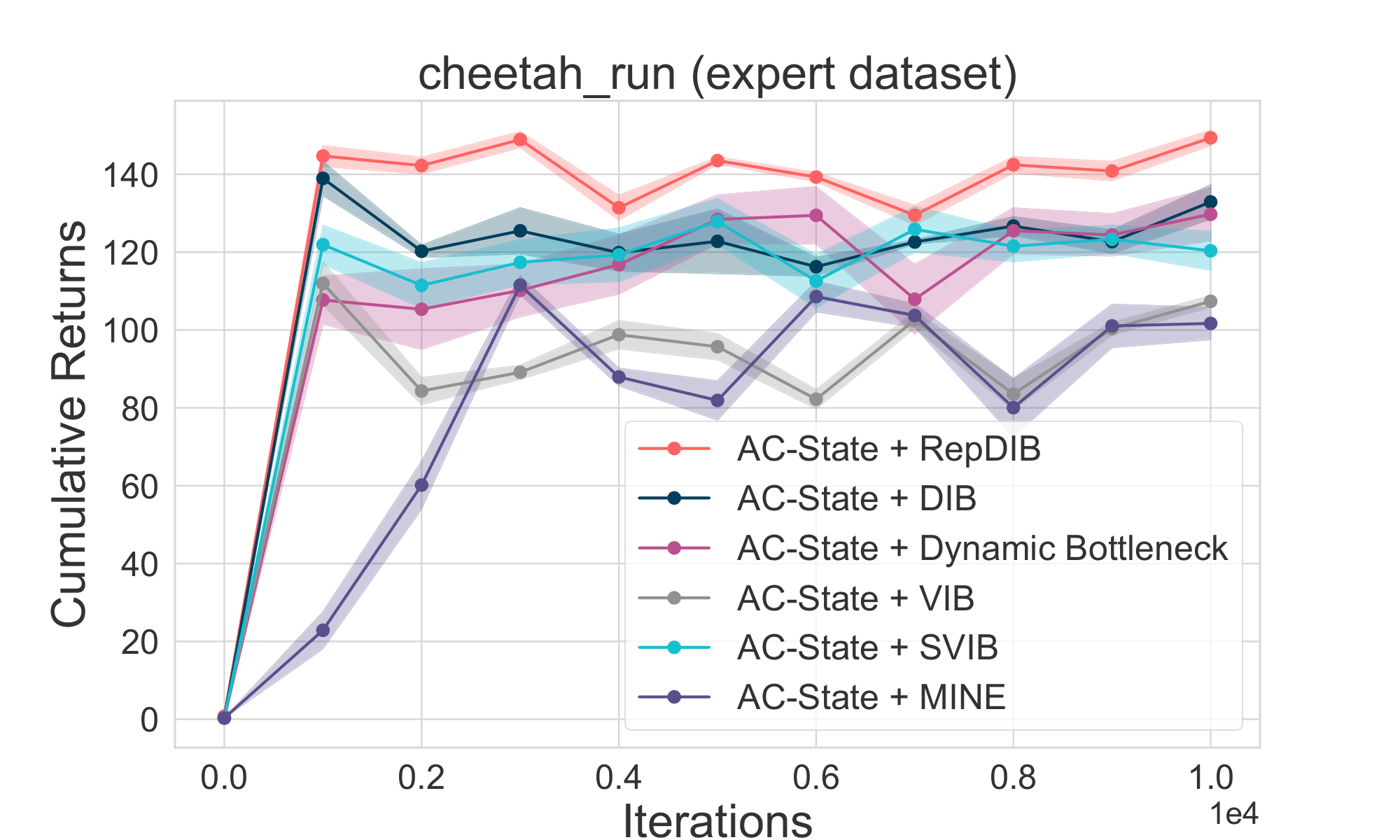}
}
\subfigure{
\includegraphics[
trim=1cm 0cm 1cm 0cm, clip=true,
width=0.25\textwidth]{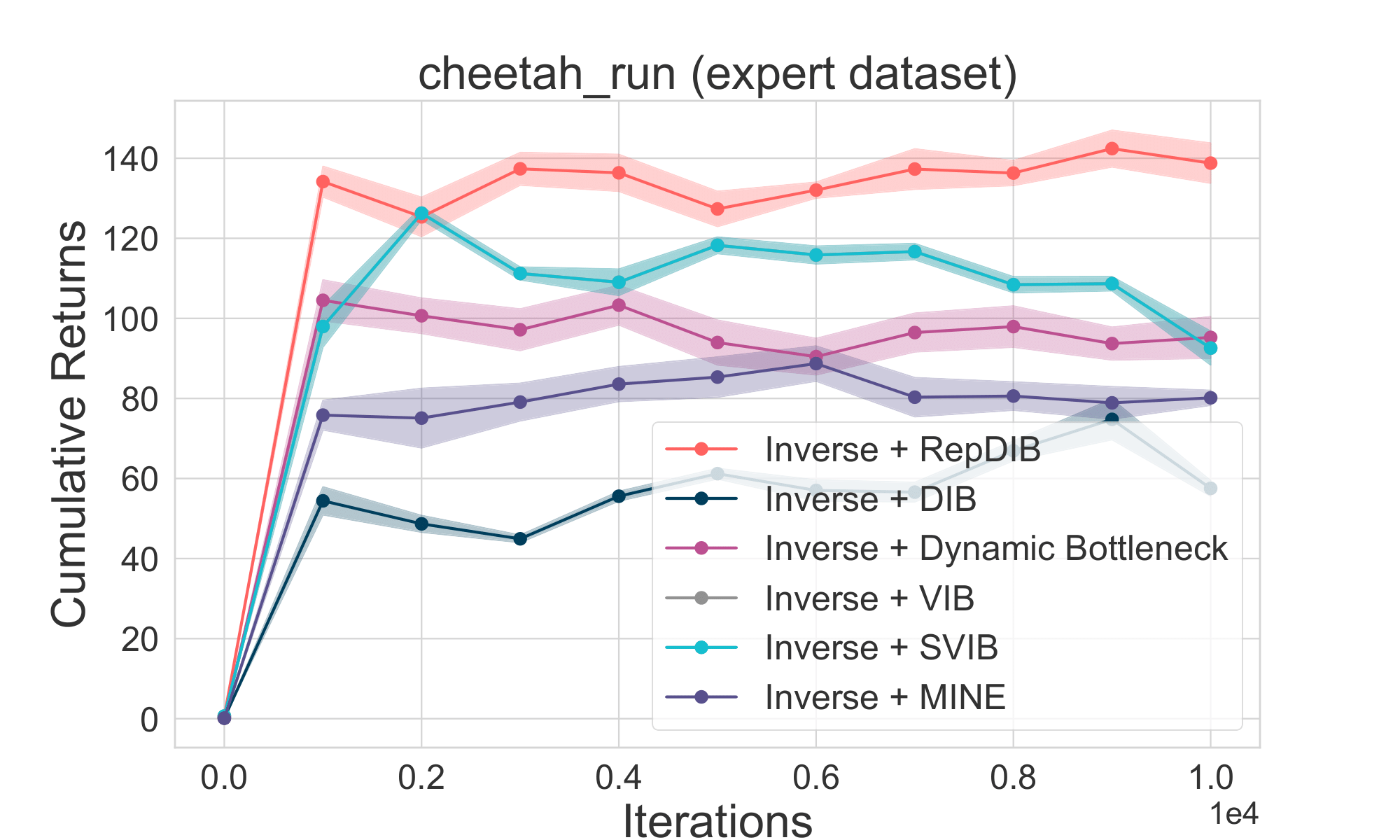}
}
\subfigure{
\includegraphics[
trim=1cm 0cm 1cm 0cm, clip=true,
width=0.25\textwidth]{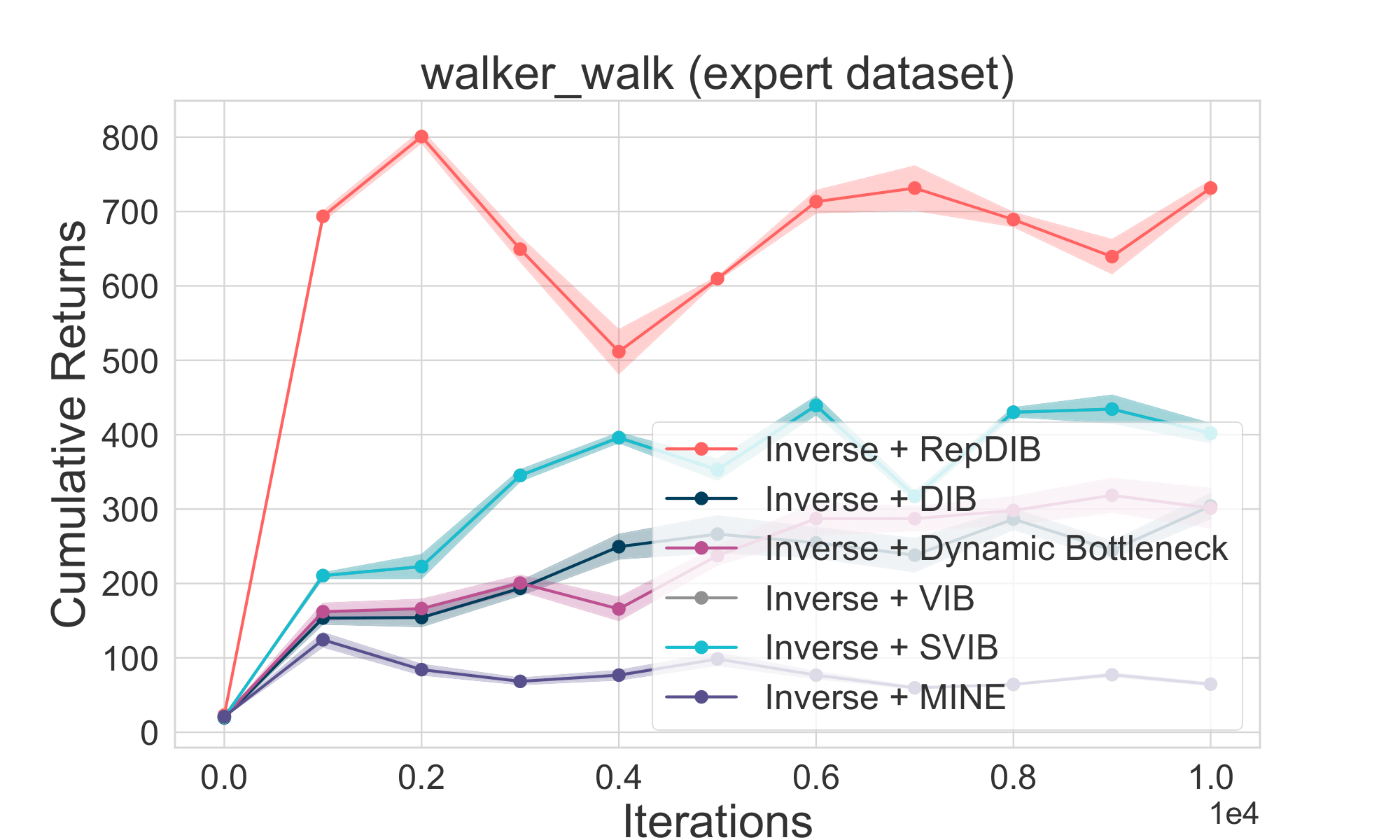}
}
\caption{\textbf{Changing video distractors, as exogenous noise in the background}. We now consider a slightly more difficult setup where we have changing video distractors in the background. We again compare \modelname with other information bottleneck based approaches, when integrated on top of existing representation learning objectives}
\label{fig:rebuttal_comparison_bottlenecks_video}
\end{figure*}

Note that the other baselines we compare with are all based on approximations of a mutual information based objective. In contrast, \modelname does not require any MI based approximations. We mainly compare with EMI(with MINE objectives)\cite{kimICML19}, DB(Dynamic Bottleneck) \cite{BaiDB} and SVIB \cite{SVIBa,SVIBb} as reviewers have pointed out, and show in figures \ref{fig:rebuttal_comparison_bottlenecks_images} and \ref{fig:rebuttal_comparison_bottlenecks_video} how \modelname compares with other baselines. Specifically, EMI proposes to maximize the mutual information of state embedding representations and action embedding representations by maximizing the estimated lower bounds of both mutual information. DB follows the Information Bottleneck principle to learn dynamics-relevant representation by maximizing the mutual information $I(Z_t; S_{t+1})$ while minimizing the mutual information $I([S_t, A_t];Z_t)$, where $Z_t$ is a compressed latent representation of $(S_t,A_t)$, $S_t, A_t$ correspond to the current state and current action respectively. SVIB utilizes the mutual information between the observation and its corresponding representation as an additionally penalized term for standard loss function in RL, optimizing all networks by Stein Variational Gradient Descent (SVGD). Notably, for a fair comparison with the other bottlenecks, we update all networks by just using Adam optimizer instead of SVGD. We emphasize that compared to the baselines, \modelname is easy to integrate since it only requires adding a VQ-VAE based factorization with a variational information bottleneck. 

\subsection{Robot Arm Experiment in Presence of Irrelevant Background Information}


\begin{figure*}[htbp]
    \centering
    \includegraphics[width=0.15\linewidth,trim={0 0 18.1cm 0},clip]{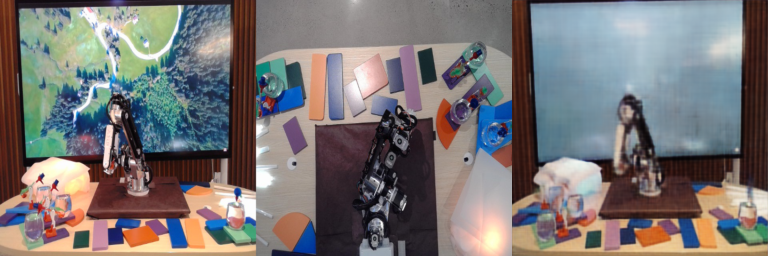}
    \includegraphics[width=0.15\linewidth,trim={0 0 18.1cm 0},clip]{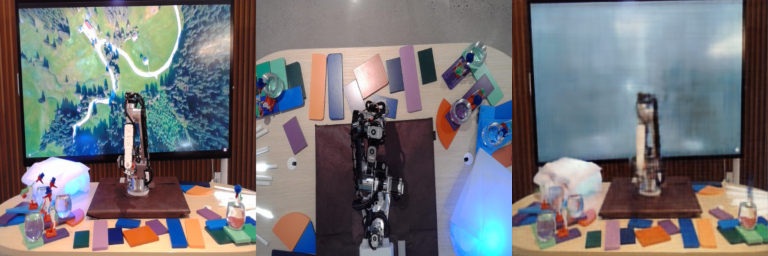}
    \includegraphics[width=0.15\linewidth,trim={0 0 18.1cm 0},clip]{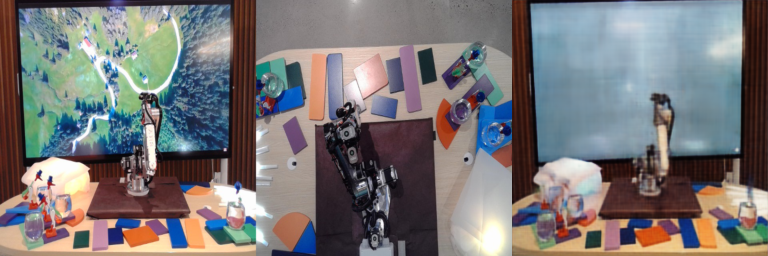}
    \tablestyle{5pt}{1.2}
    \begin{tabular}{c|c c c}
         \textbf{Bottleneck} & \textbf{None} & \textbf{\modelname (No VIB)} & \textbf{\modelname}  \\
         \shline
         Temporal Noise Relative Error & 1.0 & 0.7043 & \highlight{0.6650} \\
         Visual Noise Relative Error & 1.0 & 0.9725 & \highlight{0.9713} \\
         State Estimation Relative Error & 1.0 & 1.001 & \highlight{0.9988} \\
    \end{tabular}
    \caption{\textbf{Representations learned from videos of a real robotic arm} (with various distractors such as  a TV and color-changing lights). We evaluate the representation quality with various types of bottlenecks.  $\modelname$ is best able to remove noise from the representation without removing information about the true state of the robot}
    \label{fig:robot}
\end{figure*}

\textbf{Experiment Details}. We then evaluate $\modelname$ on a challenging real robot dataset, containing high resolution video of a robot arm in presence of a rich temporal background noise \cite{lamb2022guaranteed}. For learning a latent state representation of the images, we use a multi-step inverse model \cite{lamb2022guaranteed,Efroni2021ppe}, and integrate $\modelname$ with the information and discretization bottleneck on the learnt representation. In this task, the robot arm moves on top of a grid layout, containing $9$ different positions. We denote these as the \textit{true states}. We collect a dataset containing pixel based observations only, where the images consist of the robot arm along with the background distractors. Inspired by the exogenous noise information setup \cite{Efroni2021ppe}, we setup the robot task while there is a TV playing a video in the background, with other flashing lights nearby. The offline dataset consists of $6$ hours of robot data, with $14000$ samples from the arm, taking high level actions of move left, right, up and down. A sample point image is collected after each action, and the background distractors changes significantly, due to video and lighting in the background. The goal of the experiment is to predict accurately the ground truth state position by learning latent representations with $\modelname$.

\textbf{Experiment Results}. We evaluate the ability of $\modelname$ to accurately reconstruct the image, by learning the latent state representation while also ignoring the background distractors. This is denoted as the \textit{Image Noise}, where we compare $\modelname$ with and without VIB, alongside a baseline agent which only learns a representation. For learning latent representations, we use a multi-step inverse dynamics model \cite{Efroni2021ppe}. In addition, we compare the ability of $\modelname$ to accurately predict the ground truth states, denoted by \textit{State Accuracy} solely from the observations, as a classification task. This is challenging since the learnt representation needs to predict ground states while ignoring the irrelevant background information. Furthermore with the learnt model, we predict the time-step for each observation as an additional metric to determine effectiveness of $\modelname$. The time-step is an indicator of the background noise that appeared in each sample; and with \textit{Temporal Noise}, we evaluate $\modelname$ to predict the time step while ignoring irrelevant information from observations. Experiment results in Figure~\ref{fig:robot} shows that the use of VIB helps improve the ability of $\modelname$ to remove noise from the representation, while being able to almost perfectly predict the ground truth state of the robot.

\vspace{-2mm}
\subsection{Multi-Modal Representation Learning with Information Bottleneck}
\vspace{-1mm}

\textbf{Experiment Details}. We evaluated the impact of $\modelname$ to learning multi-modal representations for human activity recognition task. We extended the baseline multi-modal models in two ways to incorporate VQ bottleneck: \textbf{\modelname+MM: } We extract multi-modal representations using existing models (e.g. Keyless \cite{keyless} and HAMLET \cite{islam2020hamlet}) and then apply VQ bottleneck on the fused multi-modal representations. \textbf{\modelname+MM(\modelname+Uni):} We applied VQ bottleneck in two steps. First, we extract unimodal representations and apply VQ bottleneck to produce discretized unimodal representations. These discretized representations are fused and passed thorough a VQ bottleneck to produce task representations for the activity recognition.
In the baselines, we used five modalities: two viewpoints of RGB videos and three wearable sensors (acceleration, gyroscope, and orientation). We evaluated all the baselines on the MMAct dataset in a cross-subject evaluation setting and reported F1-Score of activity recognition task \cite{kong2019mmact}.

\begin{table}[!t]
\caption{\textbf{Cross-subject performance} comparison (F1-Score) of multi-modal learning model on MMAct dataset \vspace{-6mm}}
\label{tab:mmact_subject}
\tablestyle{4pt}{1.2}
\begin{tabular}{ccc}
\shline
    Method & F1-Score (\%) \\ \shline
    SMD \cite{hinton2015distilling}  & 63.89 \\
    Multi-Teachers \cite{kong2019mmact} & 62.67 \\
    MMAD \cite{kong2019mmact}  & 66.45\\
    HAMLET \cite{islam2020hamlet}  & 69.35 \\
    Keyless \cite{keyless}  & 71.83 \\
    \modelname+MM(HAMLET)  & 57.47 \\
    \modelname+MM(Keyless)  & 63.22 \\
    \modelname+MM(\modelname+Uni) & 69.39 \\
\shline
\end{tabular}
\end{table}

\textbf{Experiment Results}. The results in Table~\ref{tab:mmact_subject} suggest that applying VQ bottleneck on the multi-modal representations degrades the performance of multi-modal models for the activity recognition task. For example, applying VQ bottleneck on multi-modal representations from Keyless model $(\modelname+MM(Keyless))$ significantly degrades the F1-Score of the activity recognition task to $57.47\%$ from $69.35\%$. In these models, non-discretized unimodal representations are fused to produce a compressed and non-discretized multi-modal representation. The results suggest that applying VQ bottleneck on non-discretized multi-modal representation can not ensure retaining salient representations for task learning. 

On the other hand, applying VQ bottleneck both on the unimodal and multi-modal representations improves the performance of the models compared to the models that do not use $\modelname$ or use $\modelname$ only on the multi-modal representations. For example, $\modelname+MM(\modelname+Uni)$ model uses the same HAMLET model and applies $\modelname$ on the unimodal and multi-modal representations. $\modelname+MM(\modelname+Uni)$ slightly improves the performance of HAMLET. $\modelname+MM(\modelname+Uni)$ fuses the discretized unimodal representations using a modality weighting approach, which is modeled as 1D-CNN. As several works on multi-modal representation learning showed that the way to fuse unimodal representations could impact the performance of the downstream task \cite{mumu,maven,liang2022foundations}, there is room for improvement by effectively fusing the discretized unimodal representations. Moreover, as a couple of hyper-parameters in VQ bottleneck impact the model performance, such as the number of groups and number of embeddings, finding the appropriate value of hyper-parameters can improve the model performance. Thus, our experimental results show a crucial future avenue of research to utilize $\modelname$ information bottleneck for extracting salient multi-modal representations.

\vspace{-4mm}
\section{Discussion}
\vspace{-3mm}
\label{sec:discussion}
\textbf{Conclusion}. Representation learning methods in RL have been extensively studied in the recent past. However, when learning directly from observations consisting of exogenous information, the need for learning robust representations becomes vital. To this end, we propose $\modelname$ that learns robust representations by inducing a factorized structure in embedding space. Our work shows that discrete bottleneck representations that compress the relevant information from observations, can lead to substantial improvements in downstream tasks, as shown in our experimental results.

\textbf{Limitations and Future Work}. Whether the bottlenecks with different factors \textit{truly} lead to a compositional representation space that can disentangle different factors of observations is an interesting avenue for future work.  While we enforce discrete factorization, we provide no theoretical proof that this corresponds to an actual factorization structure in the data.  We believe that inducing such compositional structure can shape the path towards truly achieving better generalization capabilities of RL agents. How to achieve and leverage a compositional representation space for better generalization remains an interesting question, both theoretically and empirically.

\vspace{-1em}
\subsubsection*{Acknowledgements}
\vspace{-1em}
The authors would like to thank Remi Tachet Des Combes, Romain Laroche, Harm Van Seijen, and Doina Precup for valuable feedback on the draft. Hongyu Zang and Xin Li were partially supported by NSFC under Grant 62276024 and 92270125.

\bibliography{references}  
\bibliographystyle{plain}

\onecolumn
\aistatstitle{Supplementary Materials}

\section*{Appendix}
\section{Additional Experiment Results and Details}

\subsection{Visual Offline RL with Exogenous Observations in Datasets}
\label{sec:visual_offline_exps}
\begin{figure}[!htbp]
\centering
\subfigure{
\includegraphics[
trim=1cm 0cm 1cm 0cm, clip=true,
width=0.3\textwidth]{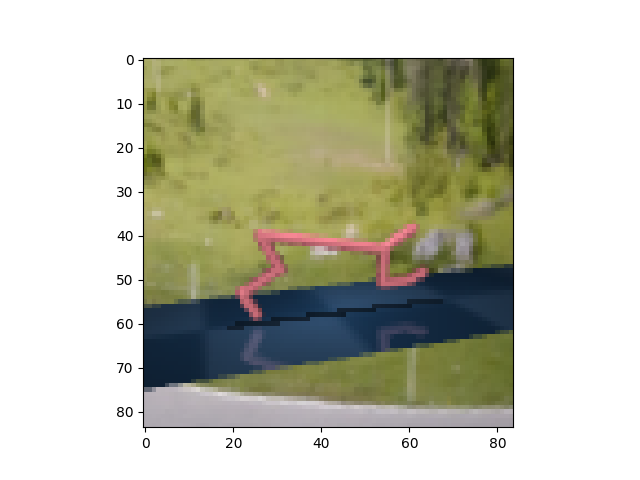}
}
\subfigure{
\includegraphics[
trim=1cm 0cm 1cm 0cm, clip=true,
width=0.3\textwidth]{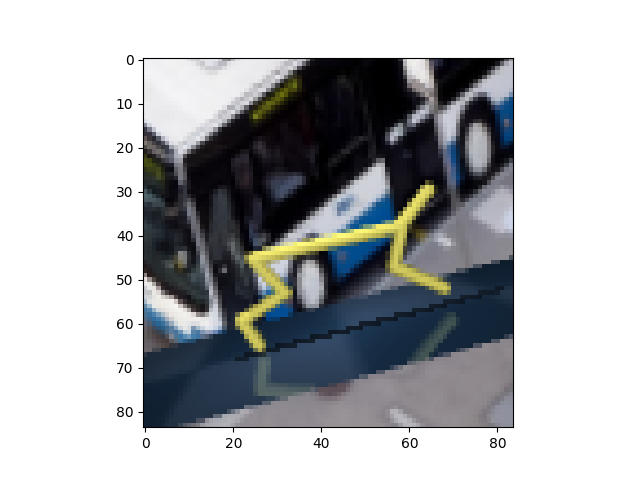}
}
\subfigure{
\includegraphics[
trim=1cm 0cm 1cm 0cm, clip=true,
width=0.3\textwidth]{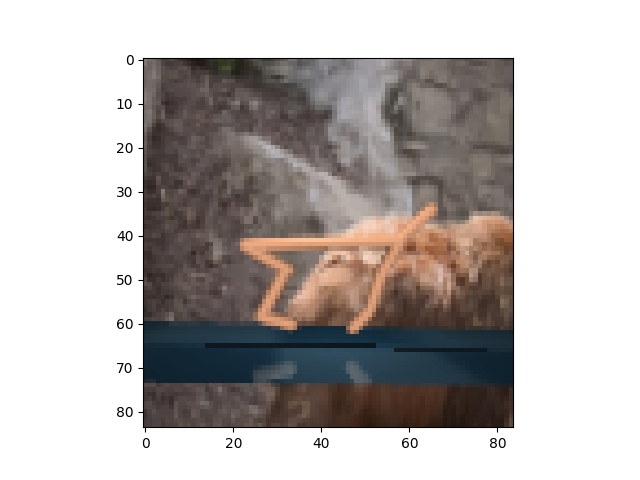}
}

\caption{Sample observations from the visual offline datasets with exogenous time correlated images in the background. The exogenous background image changes per episode during offline data collection}
\label{fig:offline_correlated_viz}
\end{figure}

\begin{figure}[!htbp]
\centering
\subfigure{
\includegraphics[
trim=1cm 0cm 1cm 0cm, clip=true,
width=0.3\textwidth]{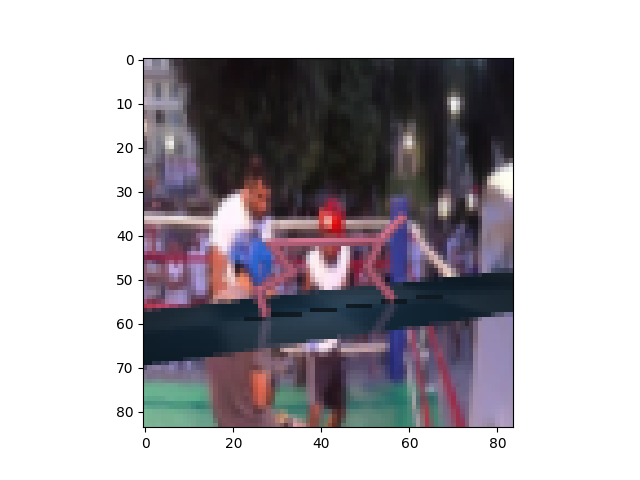}
}
\subfigure{
\includegraphics[
trim=1cm 0cm 1cm 0cm, clip=true,
width=0.3\textwidth]{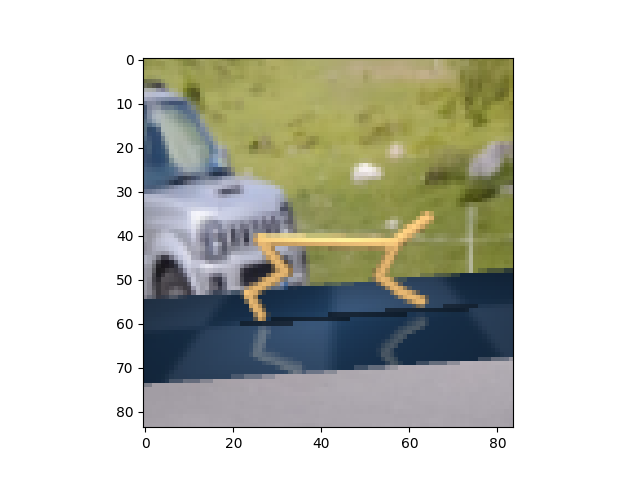}
}
\subfigure{
\includegraphics[
trim=1cm 0cm 1cm 0cm, clip=true,
width=0.3\textwidth]{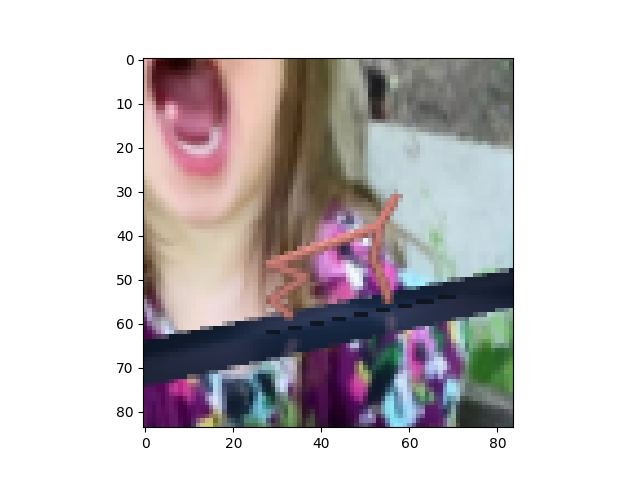}
}

\caption{Sample observations from the visual offline datasets with exogenous changing video in the background. The exogenous background video distractor changes per episode during offline data collection}
\label{fig:offline_video_viz}
\end{figure}

\textbf{Experiment Setup and Details : } We evaluate $\modelname$ on visual offline datasets from the v-d4rl benchmark \cite{vd4rl}. Data collection details on different domains are provided in \cite{vd4rl}. In addition, we also consider an extension of the v-d4rl benchmark, where we re-collect the data with additional \textit{exogenous noise} present in the observations. We follow the same data collection procedure as in v-d4rl, except during data collection, there are two variations of exogenous noise that is considered. $1.$ We first consider a time correlated exogenous noise setting where during data collection, at each episode the agent sees the environment observation and an additional background image from the CIFAR dataset. This image changes per episode of data collection, and we introduce this such that learning robust representations by avoiding the distractors plays an important role for policy learning. $2.$ We then consider a setting where instead of images that changes per episode, we now have a video distractor that changes at every episode of data collection. This is considered an even harder setting since the agent sees the observations while in addition there are unrelated video data playing in background. 

For the baseline policy optimization RL algorithms, we follow the same experiment pipeline as in \cite{vd4rl}. The major difference being, we additionally train the encoders with a representation learning objective where we pre-train the encoders with a fixed $100k$ timesteps. Following that, the learnt representations are kept fixed and we fine tune the downstream policy learning algorithms on top of the fixed pre-trained representations. For the RL algorithm, as in \cite{vd4rl} we use the TD3 + BC algorithm, since it has recently been shown to achieve state of the art performance on offline control tasks.

We provide additional results evaluating $\modelname$ on top of learnt representations in the visual pixel based offline RL setting. We implement $\modelname$ on top of the multi-step inverse dynamics objective \cite{lamb2022guaranteed}, 1-step inverse dynamics \cite{pathak2017curiosity} and the temporal contrastive learning based DRIML \cite{MazoureCDBH20} objective. We show that $\modelname$ additionally compresses the learnt latent representations using the factorial bottlenecks, which makes the method quite effective and robust especially when there is additional exogenous information present in the observations \cite{Efroni2021ppe}. We use different types of distractors in the offline datasets, where exogenous information can be either in the form of correlated background images or changing video distractors playing in the background during data collection. Figures \ref{fig:offline_correlated_viz} and \ref{fig:offline_video_viz} shows sample observations from the offline dataset.

\begin{figure}[!htbp]
\centering
\subfigure{
\includegraphics[
trim=1cm 0cm 1cm 0cm, clip=true,
width=0.3\textwidth]{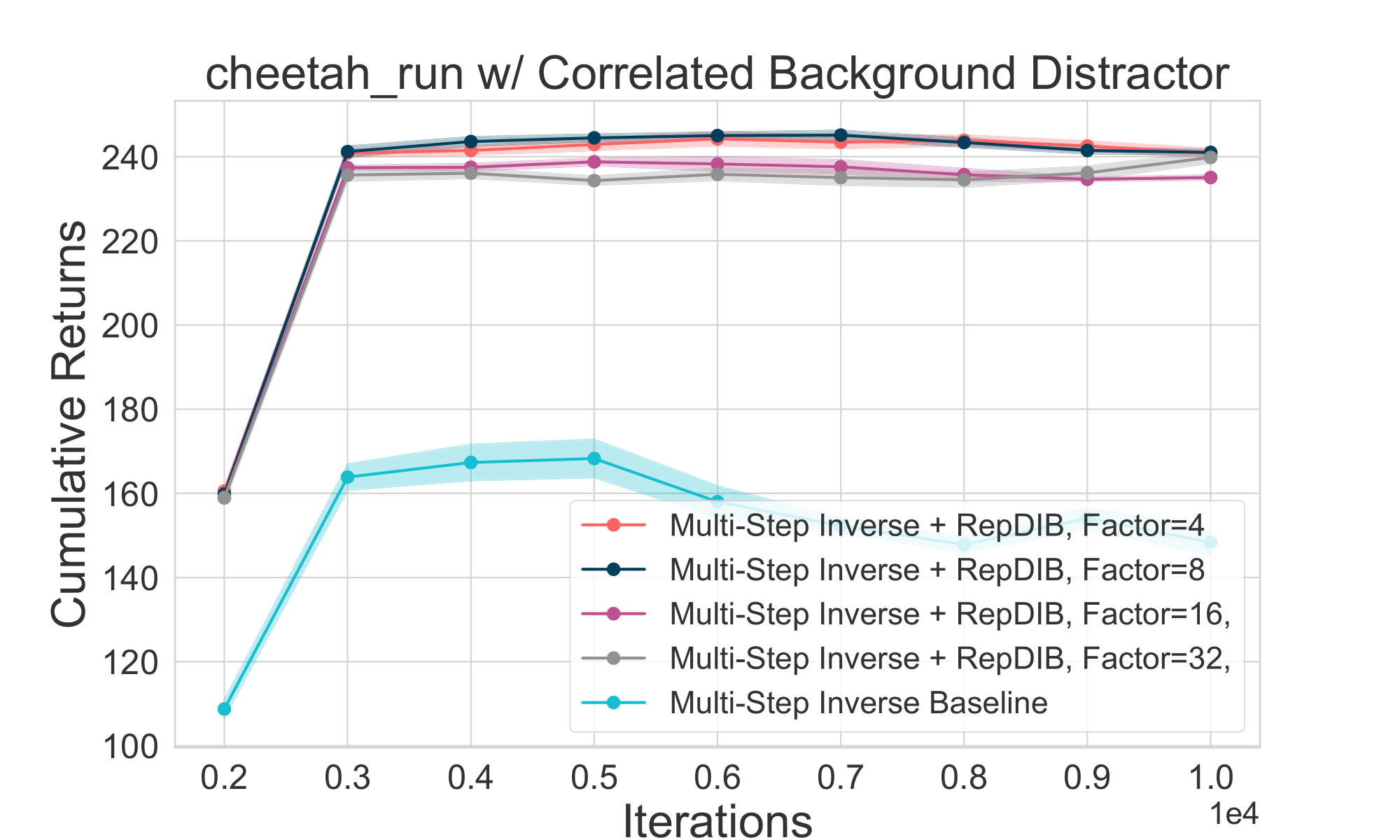}
}
\subfigure{
\includegraphics[
trim=1cm 0cm 1cm 0cm, clip=true,
width=0.3\textwidth]{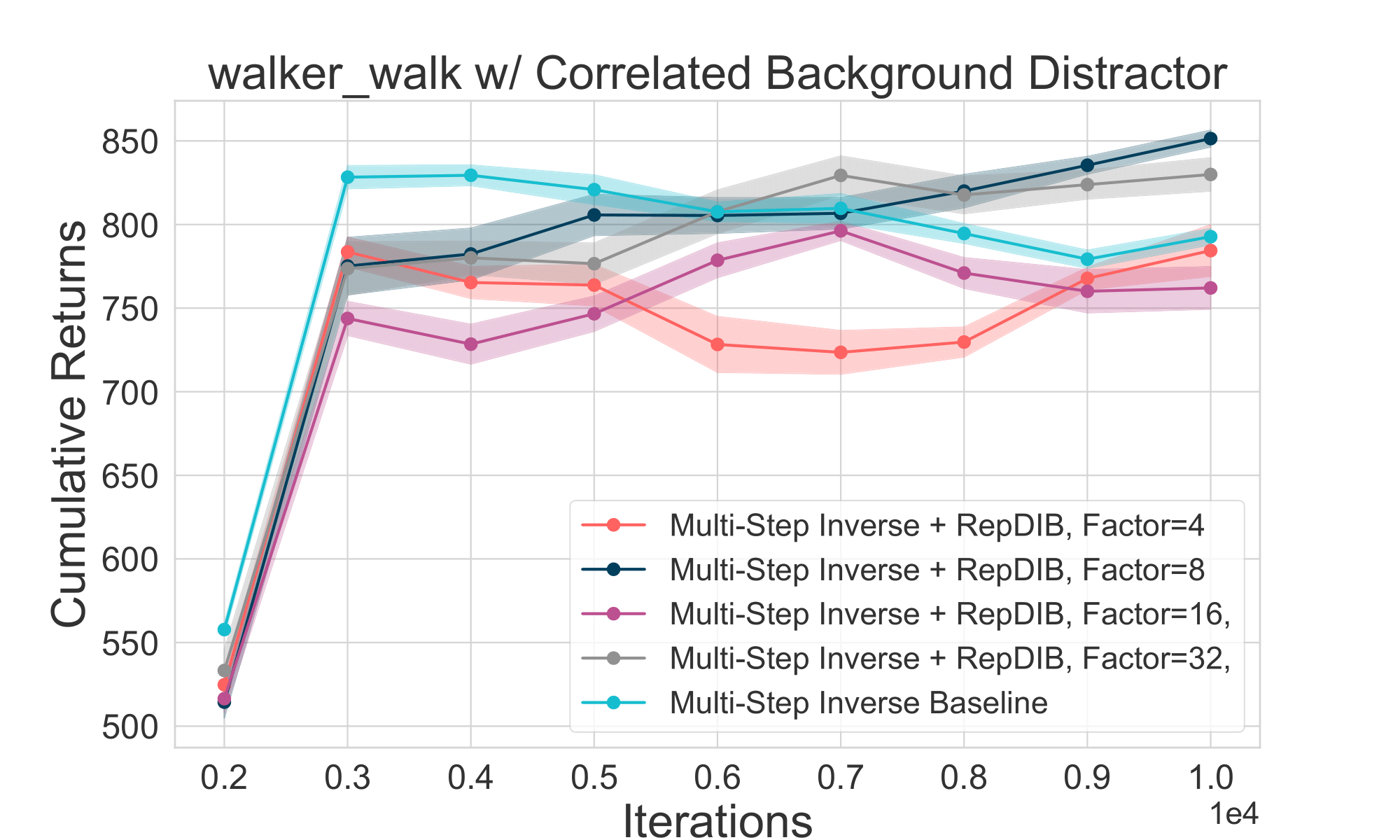}
}
\subfigure{
\includegraphics[
trim=1cm 0cm 1cm 0cm, clip=true,
width=0.3\textwidth]{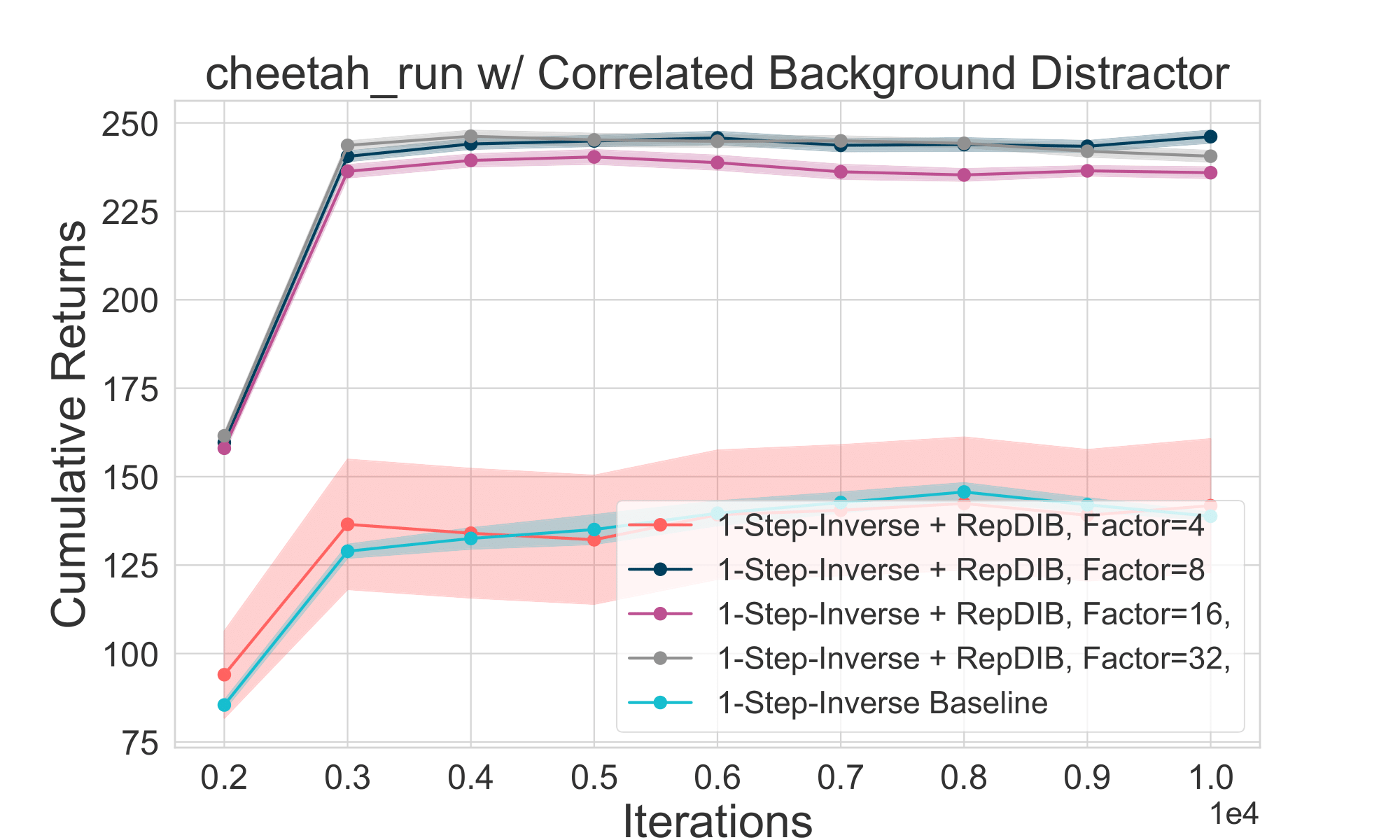}
}
\\
\centering
\subfigure{
\includegraphics[
trim=1cm 0cm 1cm 0cm, clip=true,
width=0.3\textwidth]{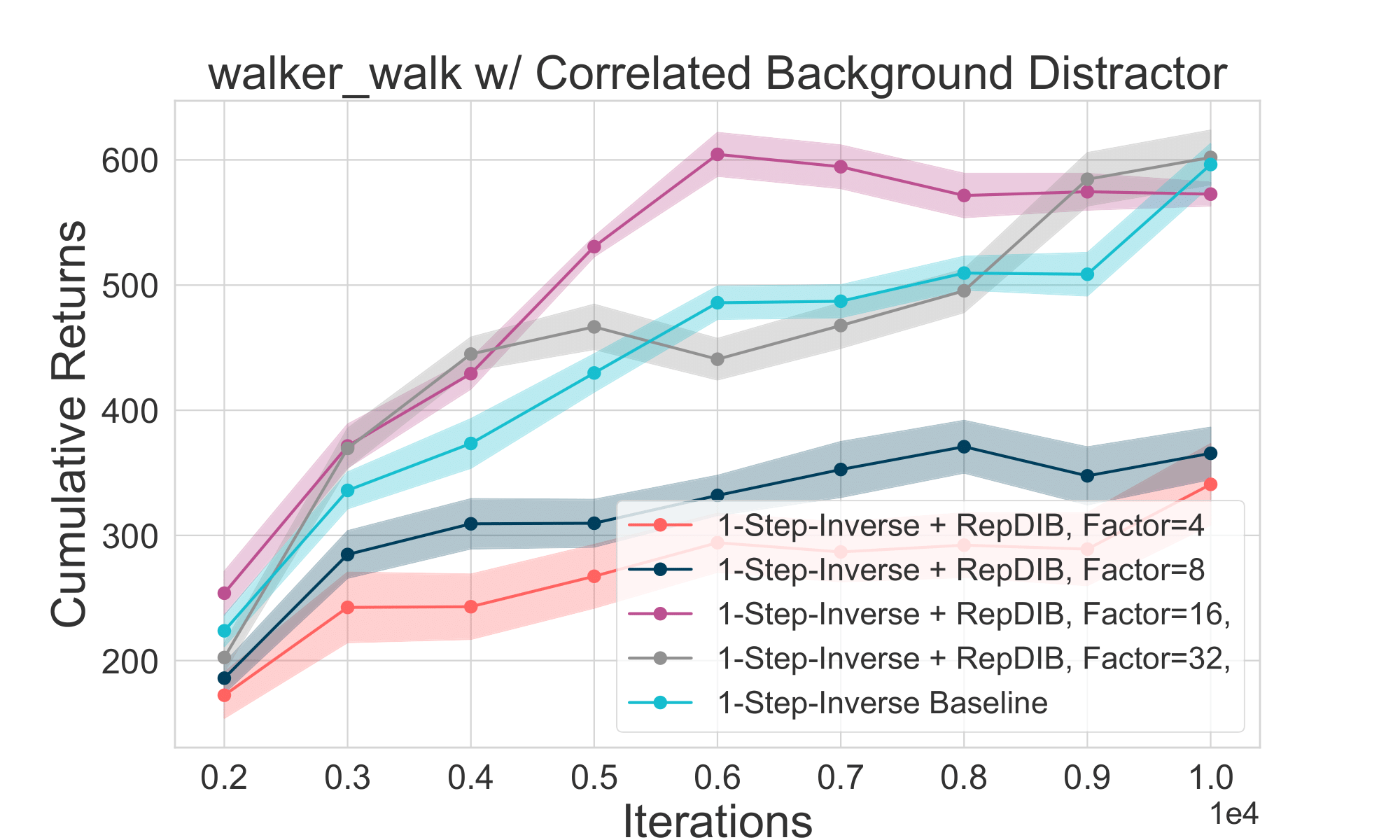}
}
\subfigure{
\includegraphics[
trim=1cm 0cm 1cm 0cm, clip=true,
width=0.3\textwidth]{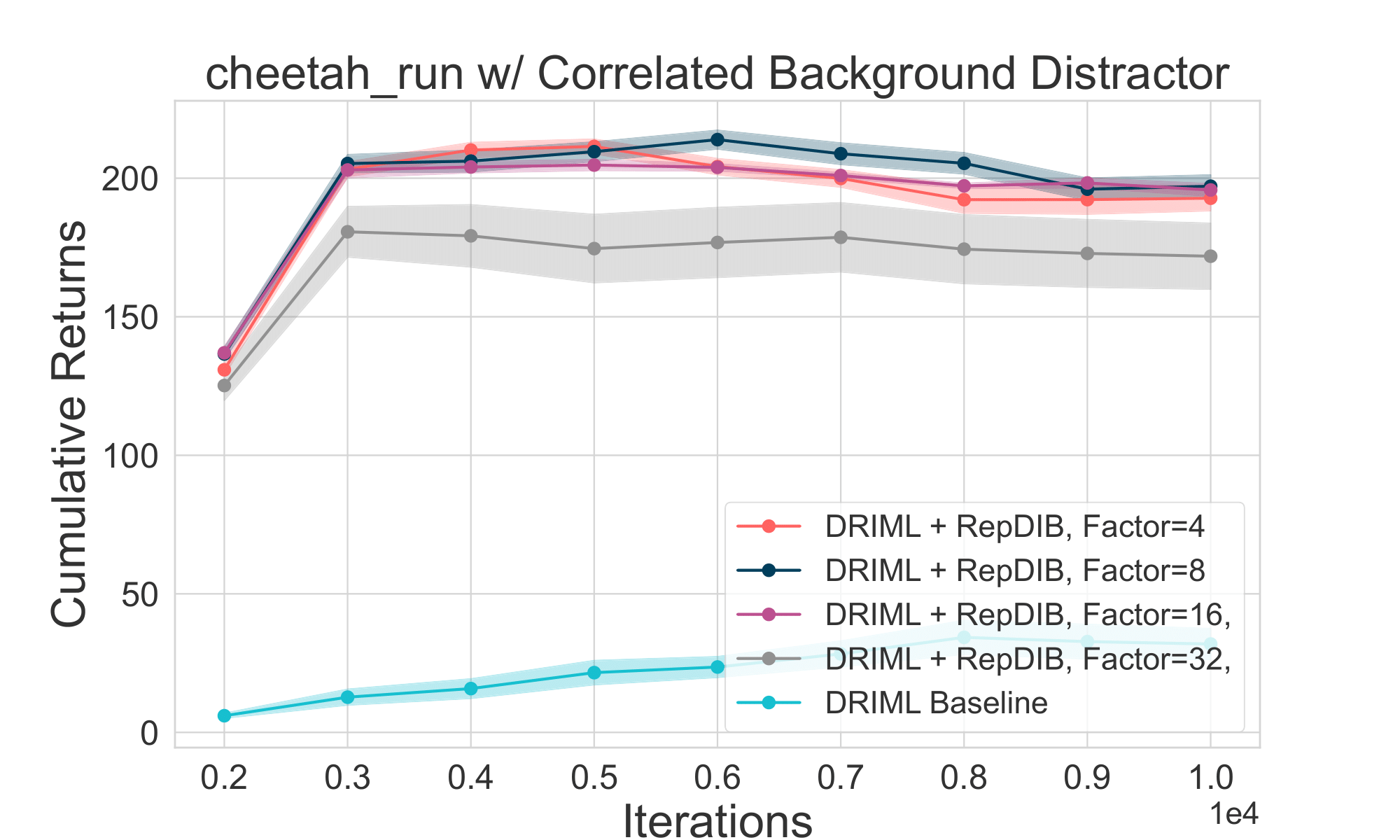}
}
\subfigure{
\includegraphics[
trim=1cm 0cm 1cm 0cm, clip=true,
width=0.3\textwidth]{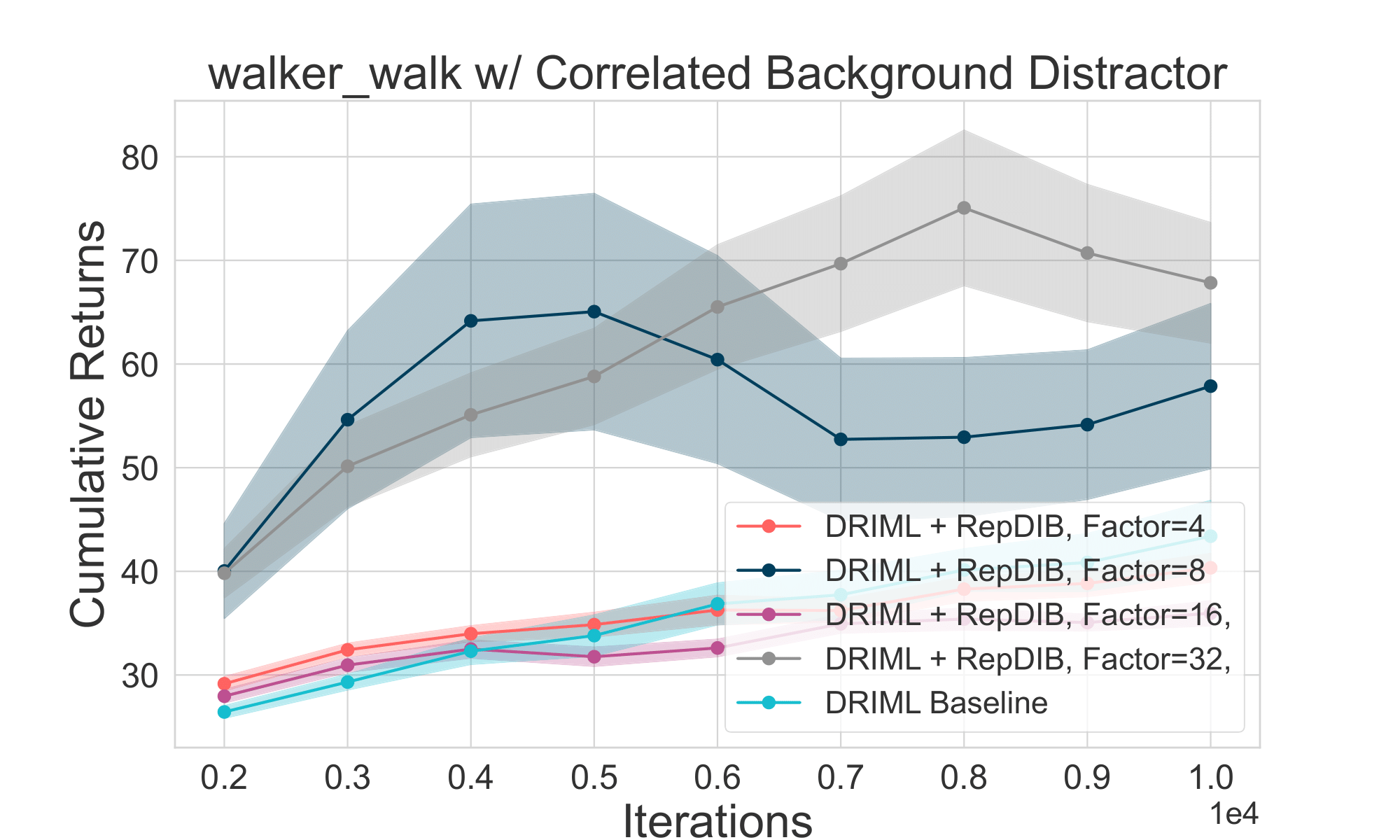}
}
\caption{\textbf{Time correlated changing background image exogenous noise in offline datasets}. We consider a setting where the observations in the pixel based offline data consists of changing background image distractors in the background. The background exogenous noise is introduced during the data collection procedure. Such a setting requires learning of robust representations that can be invariant to the exogenous images. We consider 3 different representation learning objectives (a) Multi-Step Inverse (b) One-Step Inverse and (c) DRIML, where the encoders are pre-trained with these self-supervised objectives, followed by $\modelname$. We show that for different factorial representations based on groups of factors $4, 8, 16, 32$, the ability of these methods to learn robust representations due to $\modelname$ significantly increases, making them more robust to the exogenous offline datasets.}
\label{fig:offline_correlated_appendix}
\end{figure}

\textbf{Experiment Results :} Our experiments show that existing representation learning methods can suffer in presence of this exogenous noise being present, since the representations cannot fully avoid the distractors. In contrast, when adding $\modelname$ on top of the learnt representations, we find that compressed bottleneck representations can help in avoiding the distractors, improving the overall performance in the downstream offline RL tasks consisting of visual observations. We evaluate $\modelname$ on top of learnt representations using a 1-step inverse dynamics objective \cite{pathak2017curiosity}, a multi-step inverse dynamics objective \cite{lamb2022guaranteed, Efroni2021ppe} and the temporal contrastive learning based DRIML objective \cite{MazoureCDBH20}. Our results show that especially when exogenous noise is present in the observations, existing state of the art representation learning methods can suffer dramatically, leading to an overall degradation of performance. In contrast, addition of $\modelname$ can lead to improved performance due to bottlenecks that can capture factorial representations, while avoiding the exogenous distractors. 

\begin{figure}[!htbp]
\centering
\subfigure{
\includegraphics[
trim=1cm 0cm 1cm 0cm, clip=true,
width=0.3\textwidth]{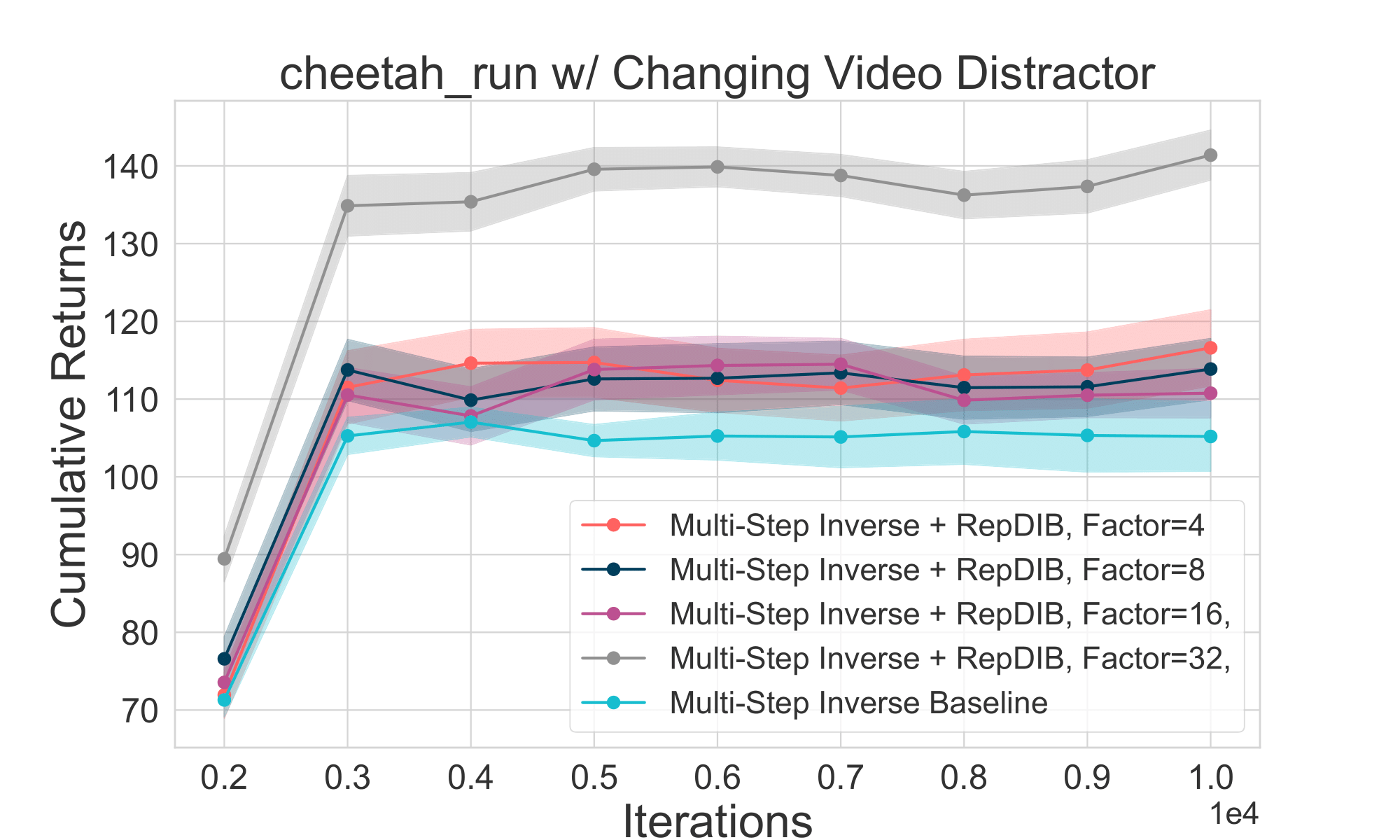}
}
\subfigure{
\includegraphics[
trim=1cm 0cm 1cm 0cm, clip=true,
width=0.3\textwidth]{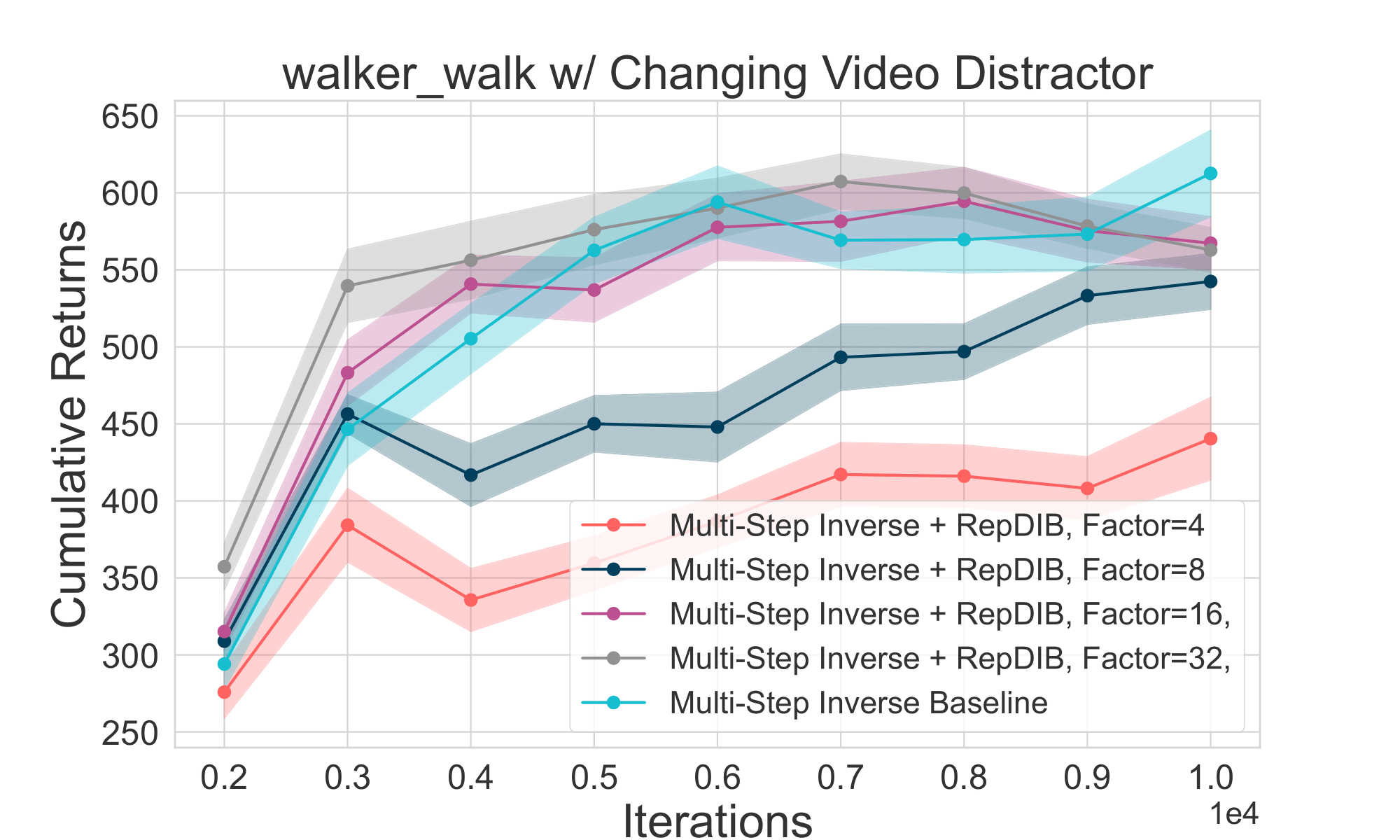}
}
\subfigure{
\includegraphics[
trim=1cm 0cm 1cm 0cm, clip=true,
width=0.3\textwidth]{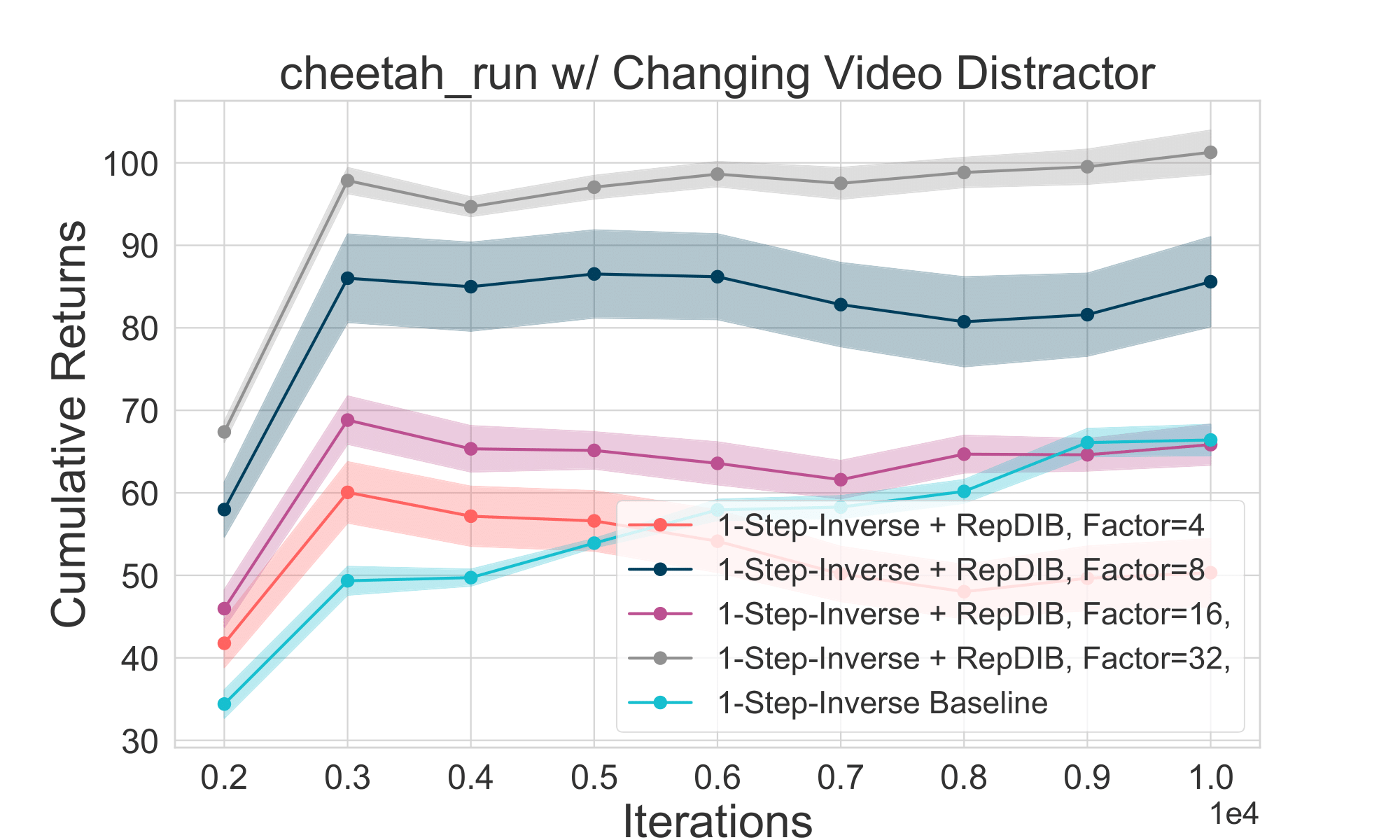}
}
\\
\centering
\subfigure{
\includegraphics[
trim=1cm 0cm 1cm 0cm, clip=true,
width=0.3\textwidth]{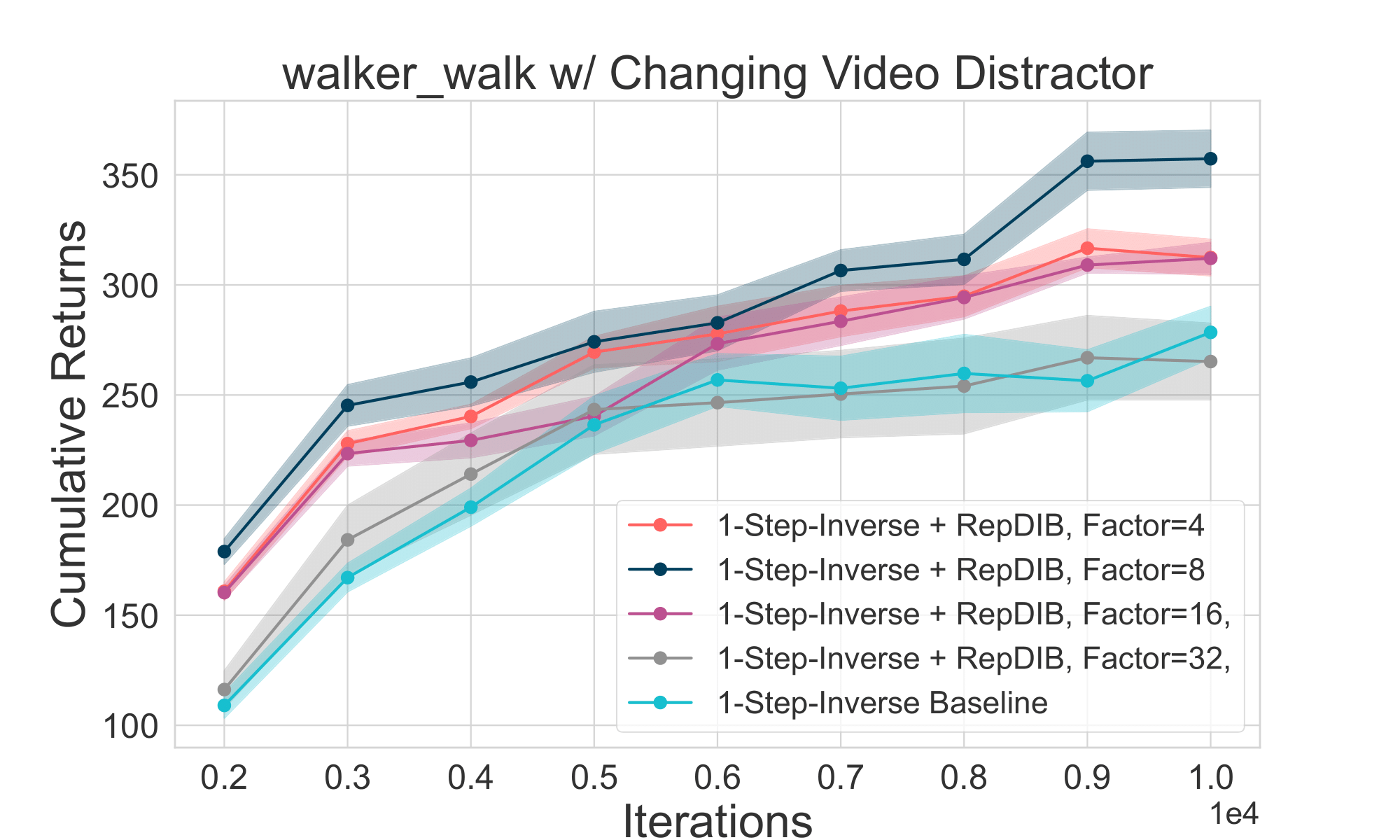}
}
\subfigure{
\includegraphics[
trim=1cm 0cm 1cm 0cm, clip=true,
width=0.3\textwidth]{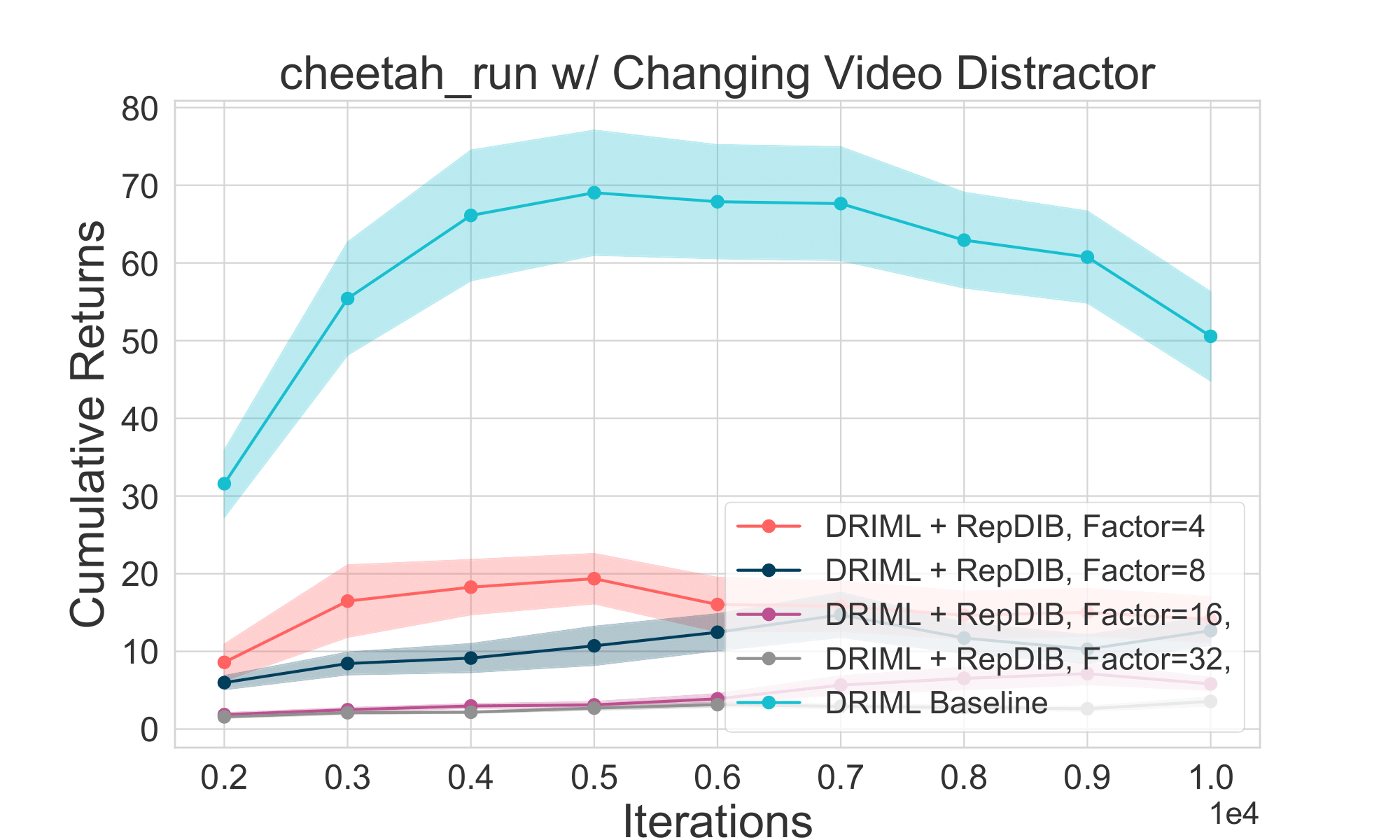}
}
\subfigure{
\includegraphics[
trim=1cm 0cm 1cm 0cm, clip=true,
width=0.3\textwidth]{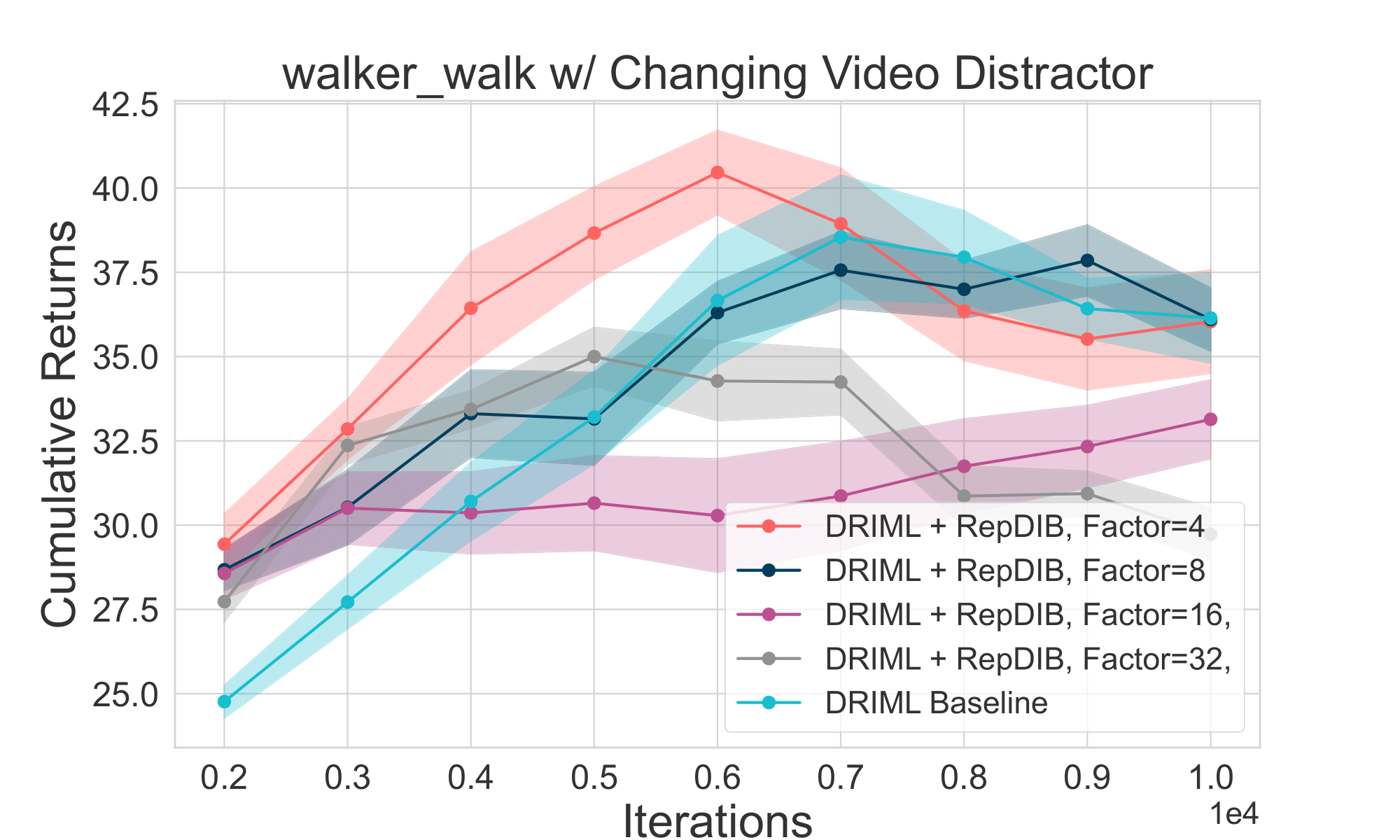}
}
\caption{\textbf{Changing video exogenous noise in offline datasets}. We then consider a setting where there is background video distractors that changes per episode during data collection. Using $\modelname$ on top of the learnt representations from the same 3 different representation objectives, we find that in particular, the multi-step and one-step inverse models learns more robust representations compared to the DRIML objective. The changing background video distractors is considered to be a hard offline setting, since there is time correlated exogenous information continuous changing and playing in the background. We show that $\modelname$ improves the sample efficiency and overall performance of these methods, when used on visual offline data where learning robust representations plays a key role.}
\label{fig:offline_change_video_appendix}
\end{figure}

\begin{figure}[!htbp]
\centering
\subfigure{
\includegraphics[
trim=1cm 0cm 1cm 0cm, clip=true,
width=0.3\textwidth]{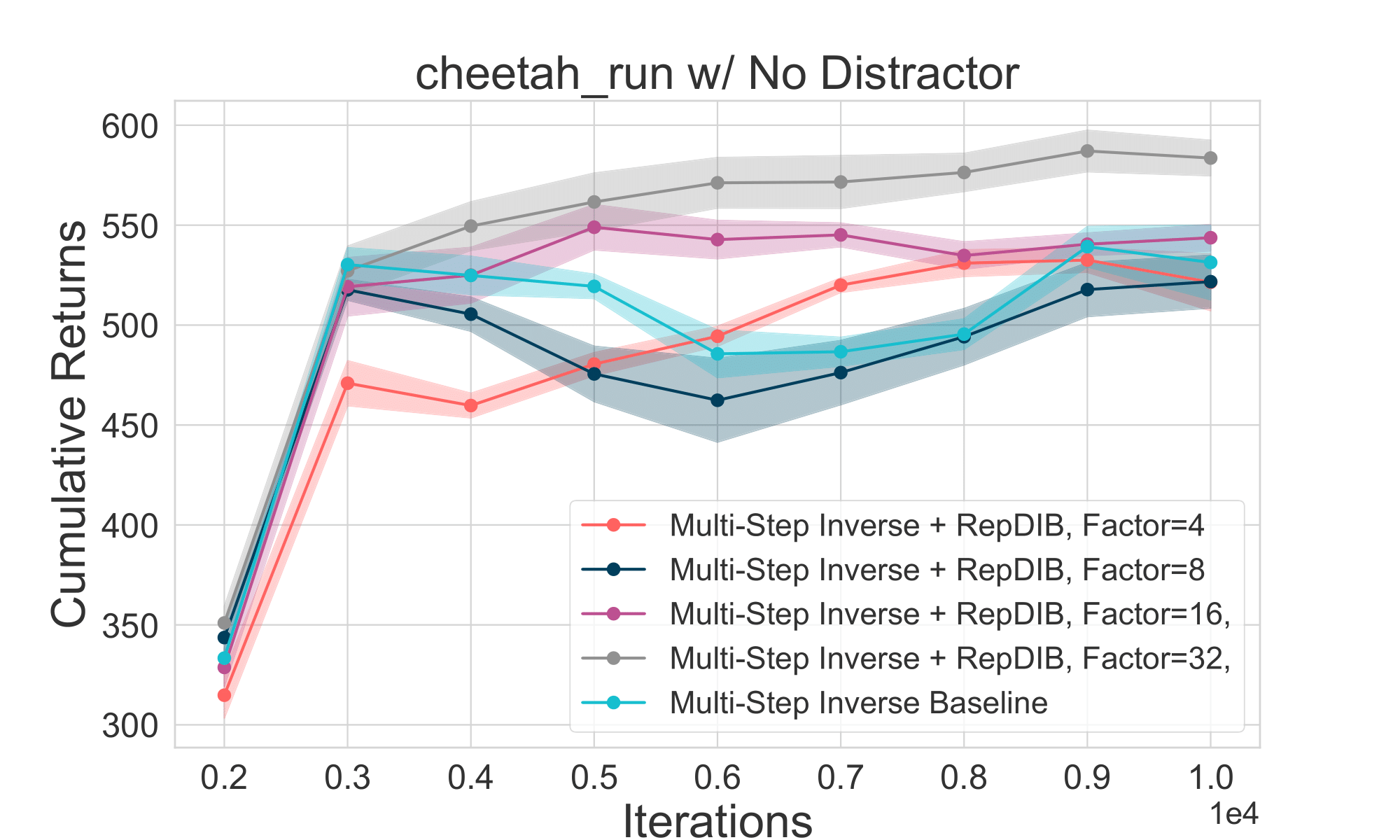}
}
\subfigure{
\includegraphics[
trim=1cm 0cm 1cm 0cm, clip=true,
width=0.3\textwidth]{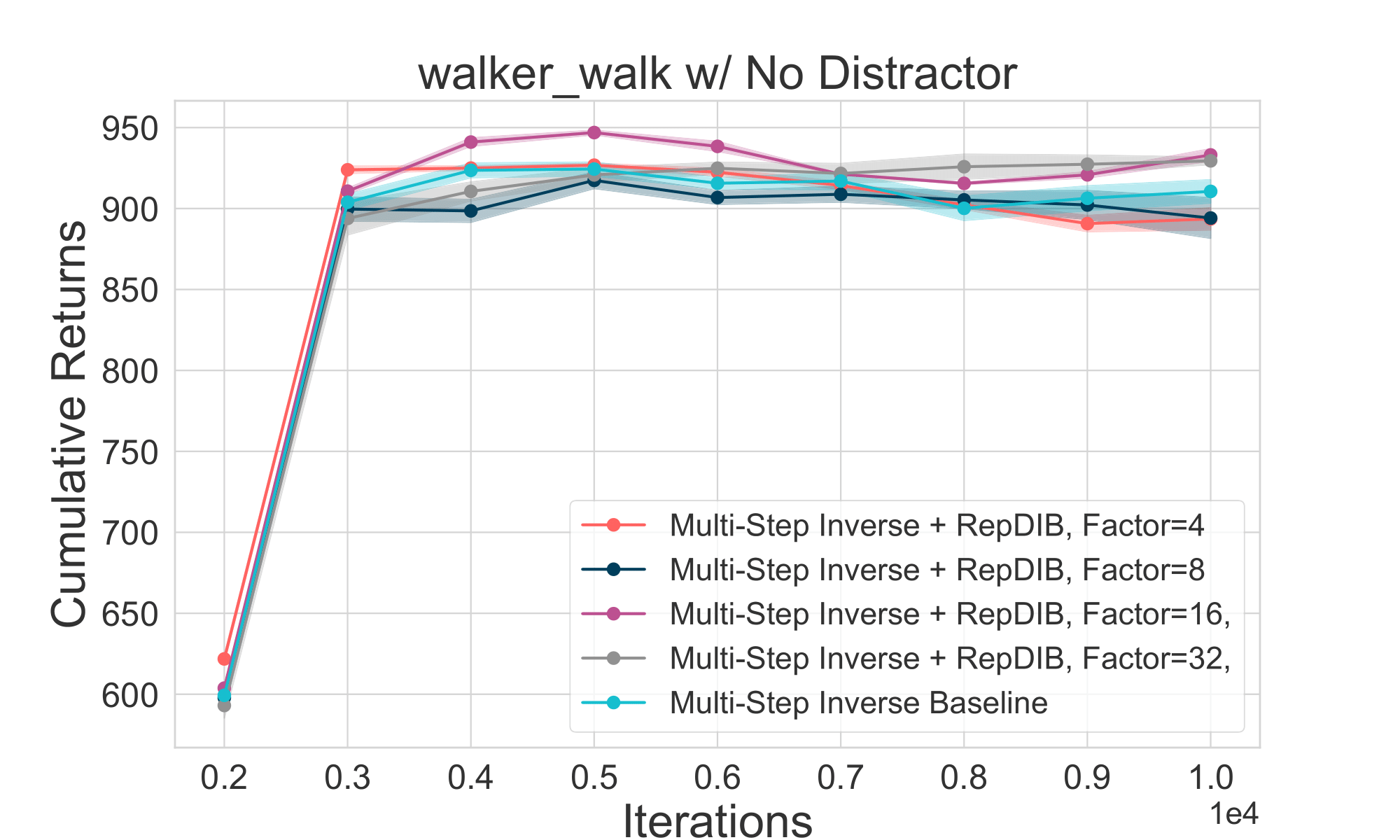}
}
\subfigure{
\includegraphics[
trim=1cm 0cm 1cm 0cm, clip=true,
width=0.3\textwidth]{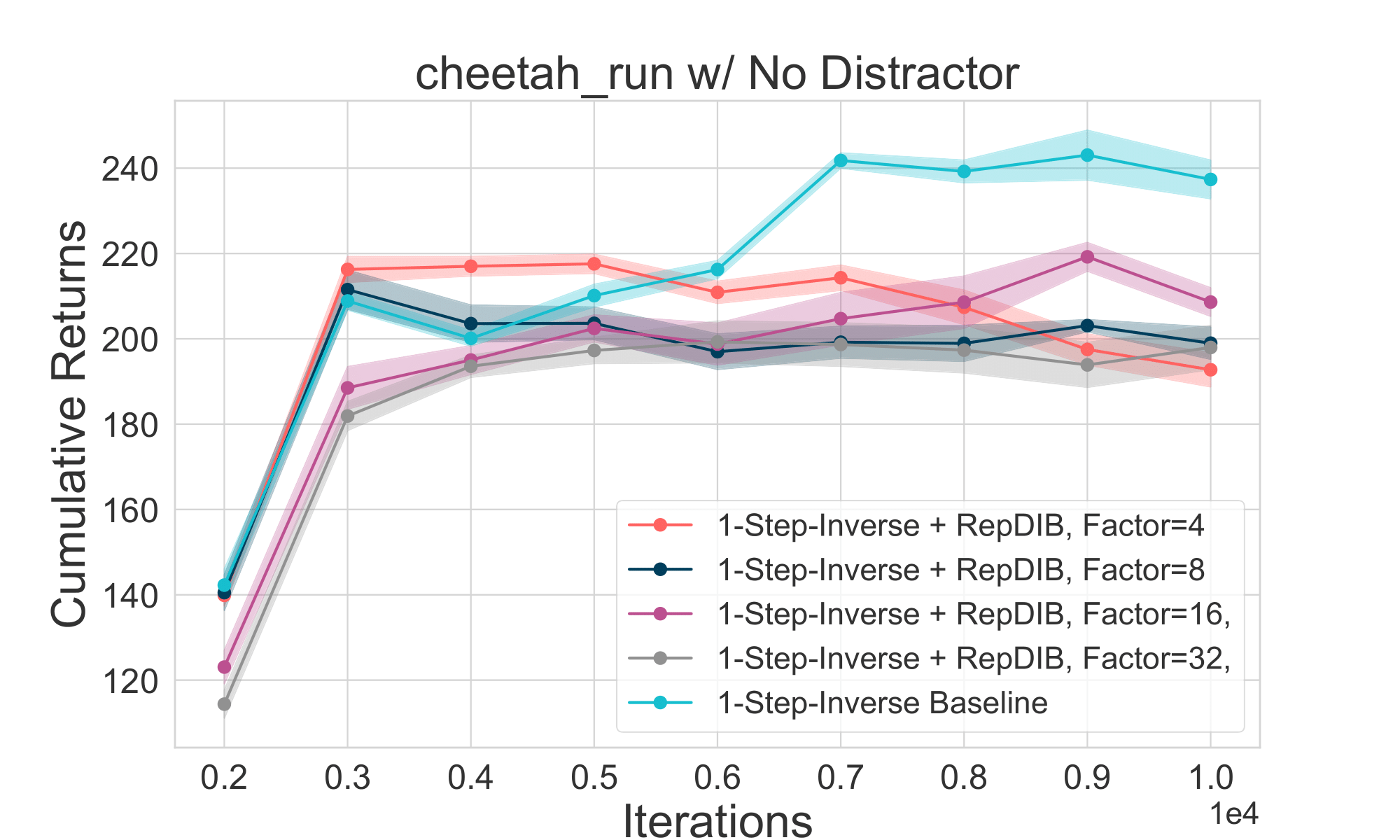}
}
\\
\centering
\subfigure{
\includegraphics[
trim=1cm 0cm 1cm 0cm, clip=true,
width=0.3\textwidth]{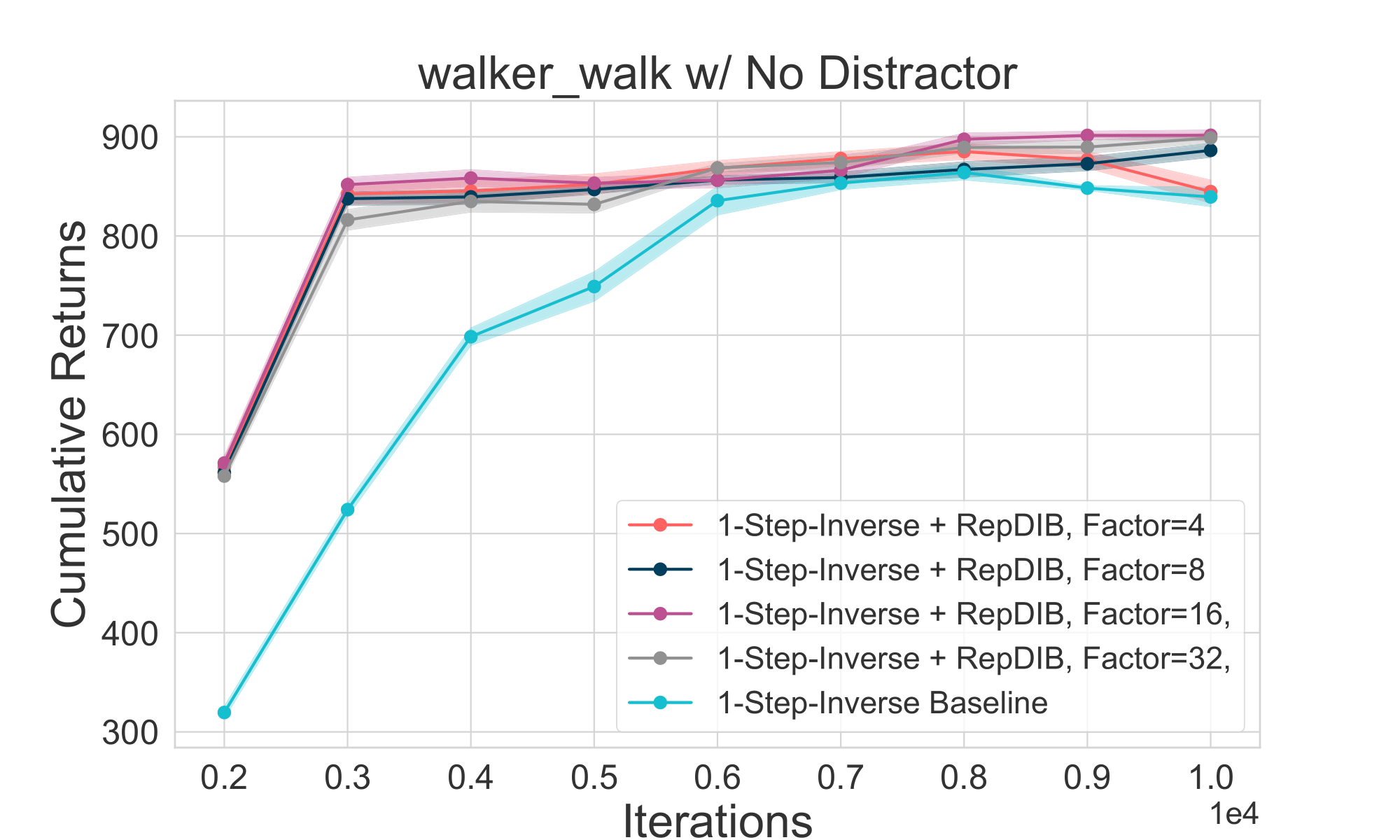}
}
\subfigure{
\includegraphics[
trim=1cm 0cm 1cm 0cm, clip=true,
width=0.3\textwidth]{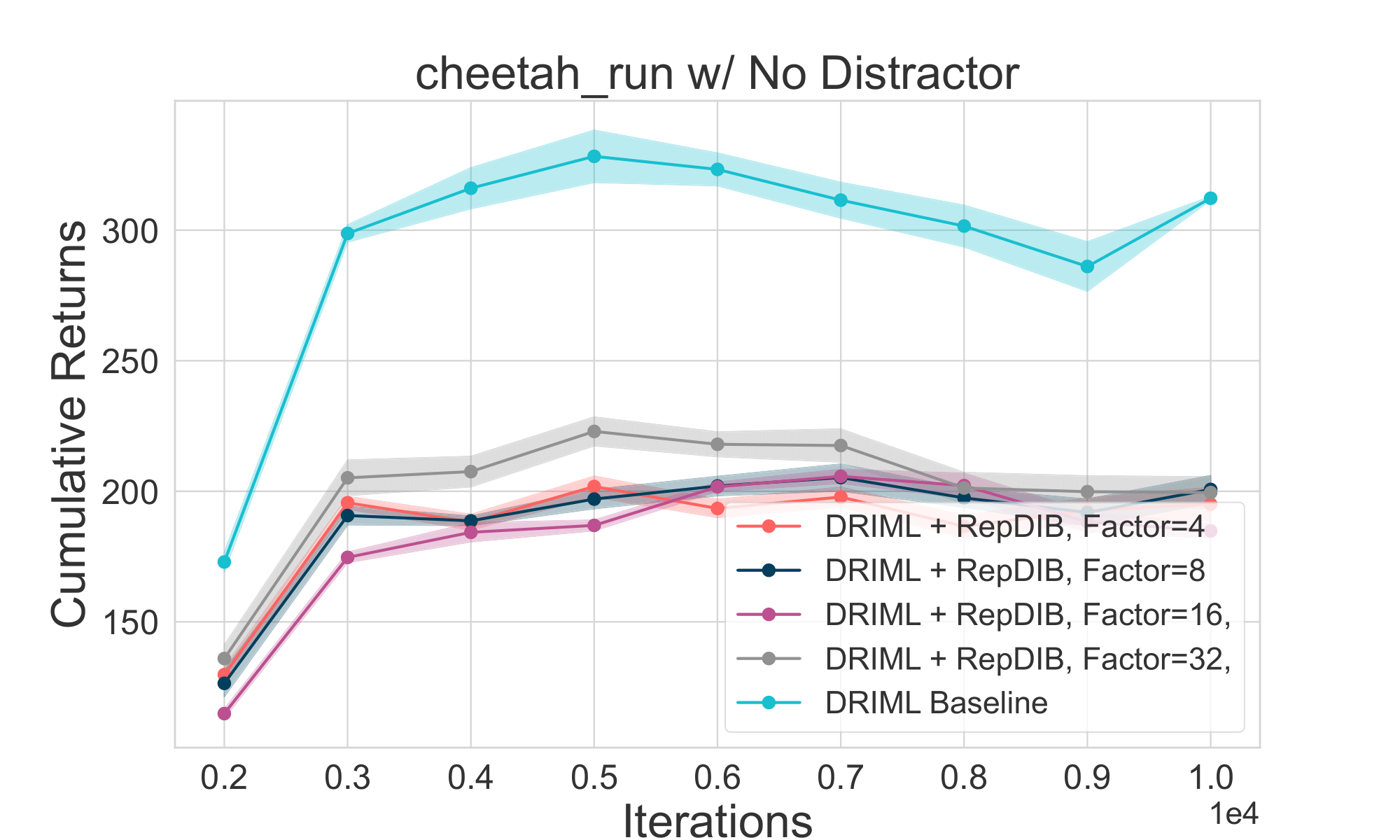}
}
\subfigure{
\includegraphics[
trim=1cm 0cm 1cm 0cm, clip=true,
width=0.3\textwidth]{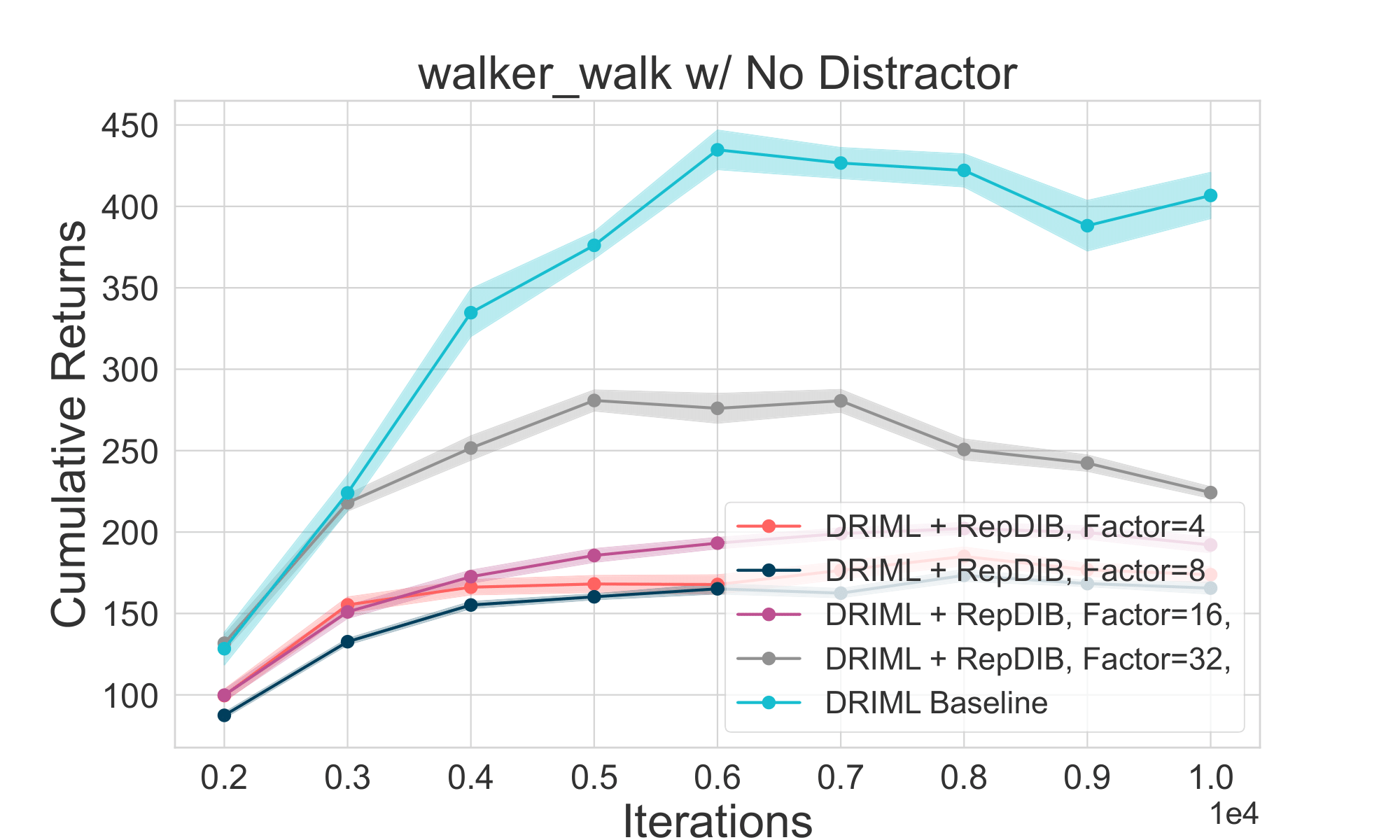}
}
\caption{\textbf{Visual offline datasets from the v-d4rl benchmark \cite{vd4rl} without any additional exogenous distractors}. We showed that $\modelname$ can learn effectively robust robust representations in presence of correlated exogenous noise as in figures \ref{fig:offline_correlated_appendix} and \ref{fig:offline_change_video_appendix}. Here we show that without any distractors being present, $\modelname$ does not necessarily always outperform baselines, as shown in the results with the DRIML objective. This validates our claim that $\modelname$ based bottlenecks can be particularly effectively when observations consist of exogenous information, such that $\modelname$ can be used to learn more robustly. Without any distractors, it is not always necessary that $\modelname$  based representations would always outperform baselines without bottlenecks. }
\label{fig:offline_no_distractor_appendix}
\end{figure}

\section{Significance of VIB and DIB for \modelname}
\label{rebuttal:vib_dib_comparison}
\begin{figure}[!htbp]
\centering
\subfigure{
\includegraphics[
trim=1cm 0cm 1cm 0cm, clip=true,
width=0.3\textwidth]{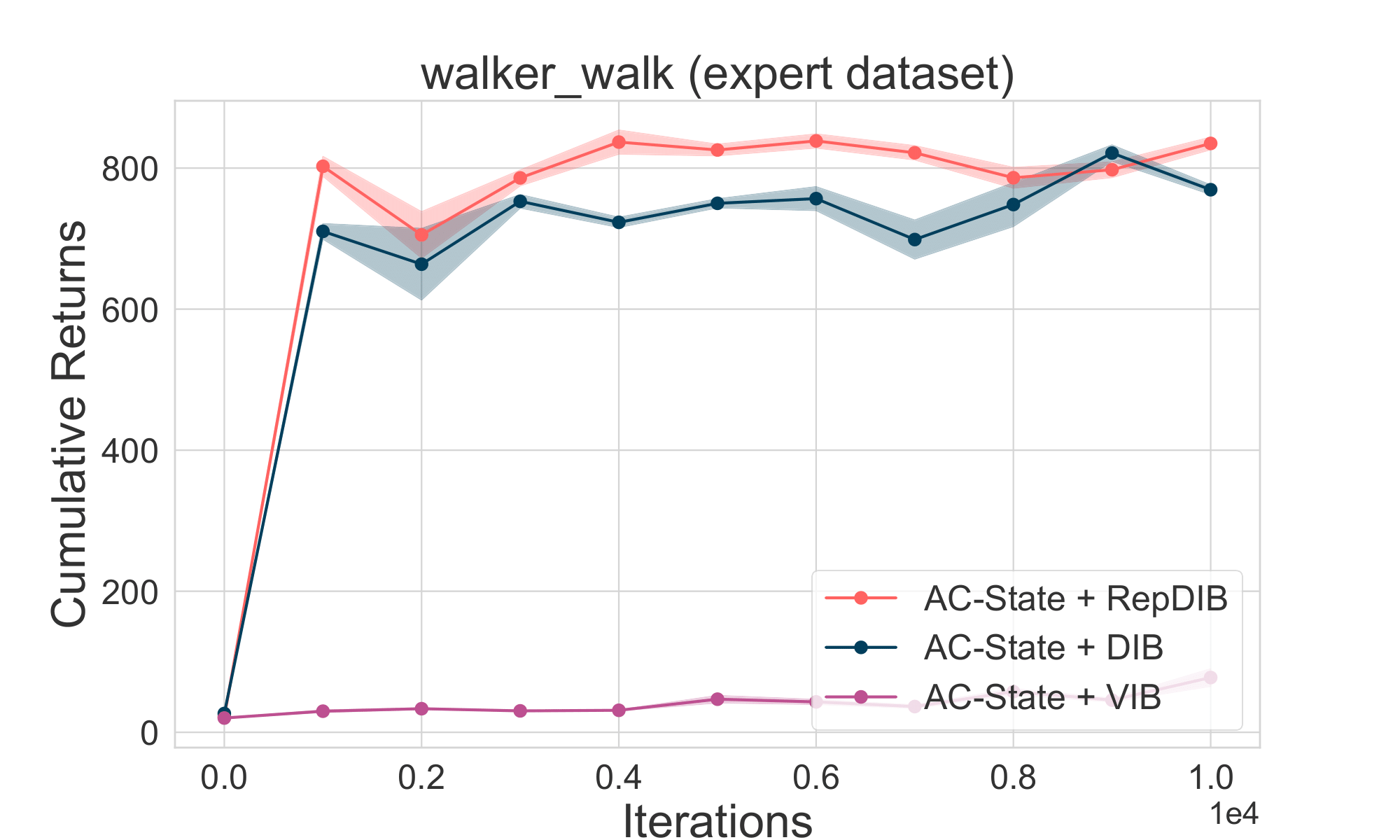}
}
\subfigure{
\includegraphics[
trim=1cm 0cm 1cm 0cm, clip=true,
width=0.3\textwidth]{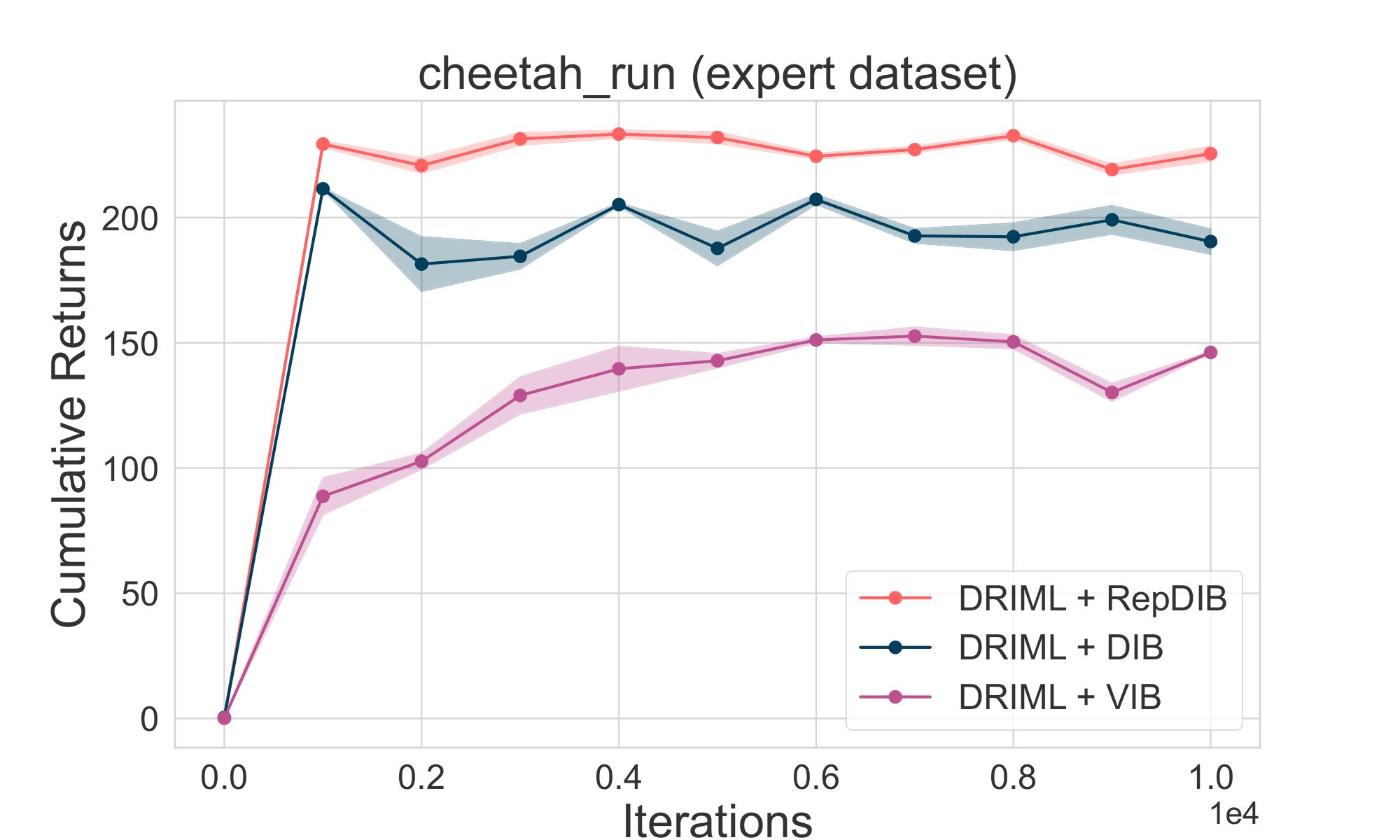}
}
\subfigure{
\includegraphics[
trim=1cm 0cm 1cm 0cm, clip=true,
width=0.3\textwidth]{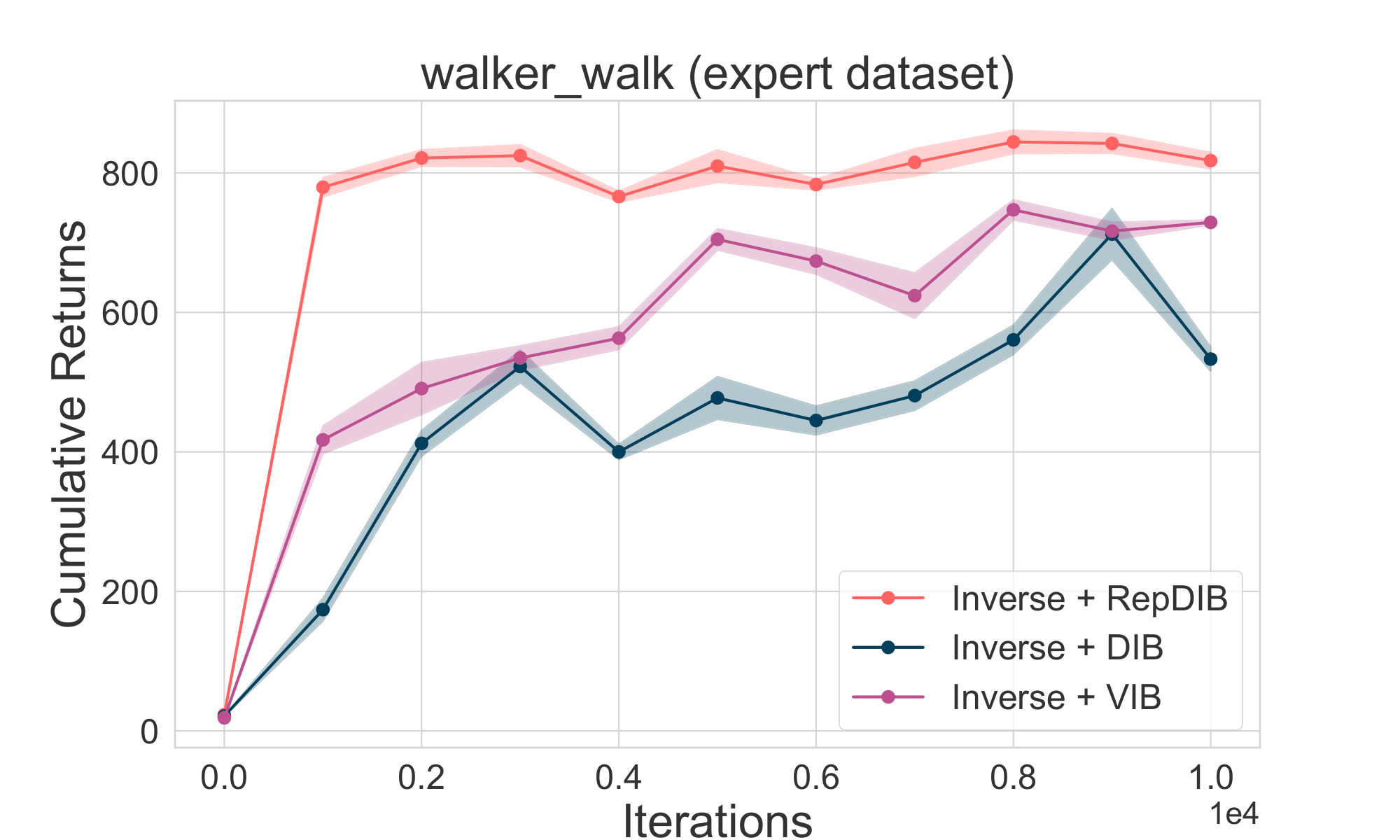}
}
\caption{Ablation studies on the \modelname bottleneck on time correlated exogenous distractors in the observations of offline datasets, as per the setup described in section \ref{sec:offline_exo}}
\label{fig:rebuttal_ablation_image}
\end{figure}

\begin{figure}[!htbp]
\centering
\subfigure{
\includegraphics[
trim=1cm 0cm 1cm 0cm, clip=true,
width=0.3\textwidth]{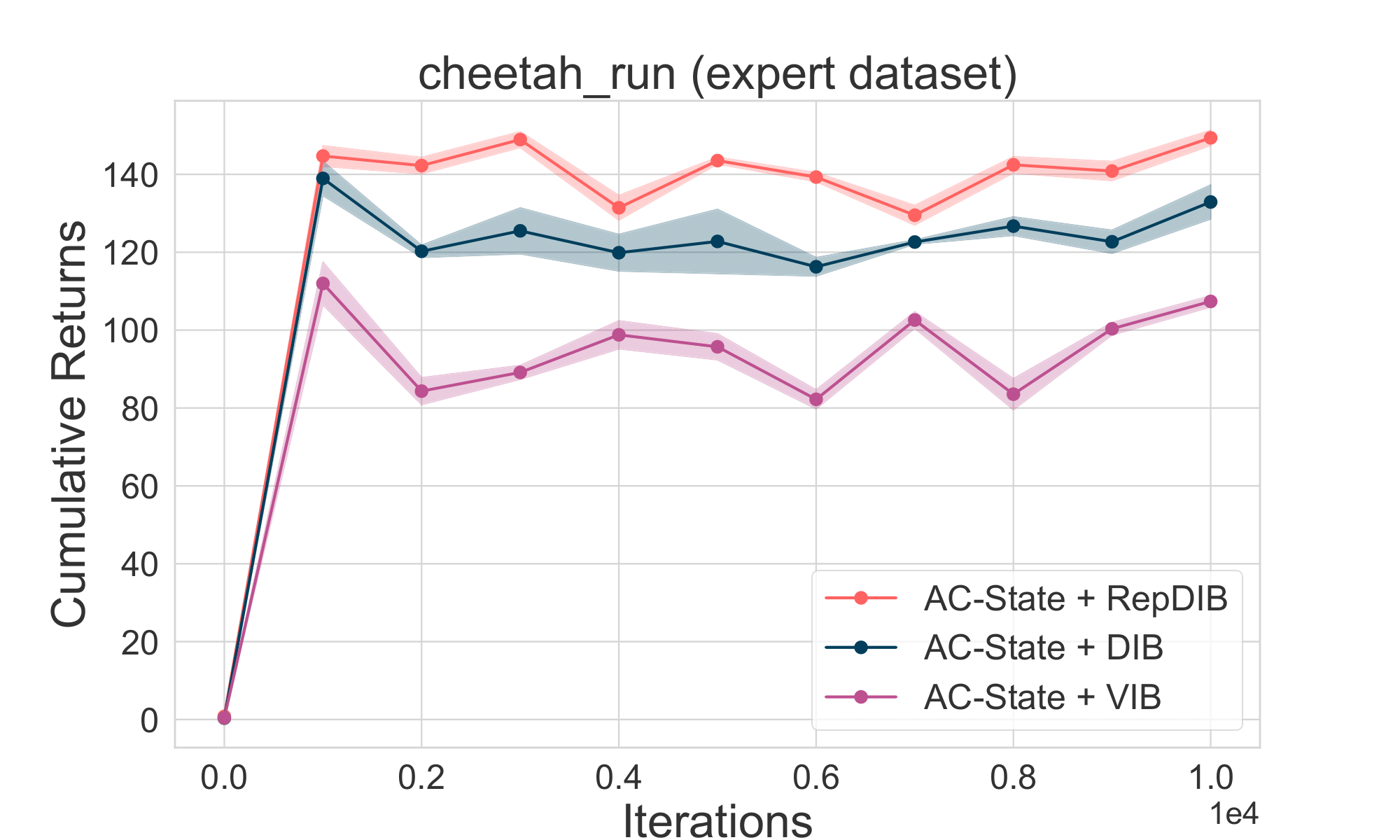}
}
\subfigure{
\includegraphics[
trim=1cm 0cm 1cm 0cm, clip=true,
width=0.3\textwidth]{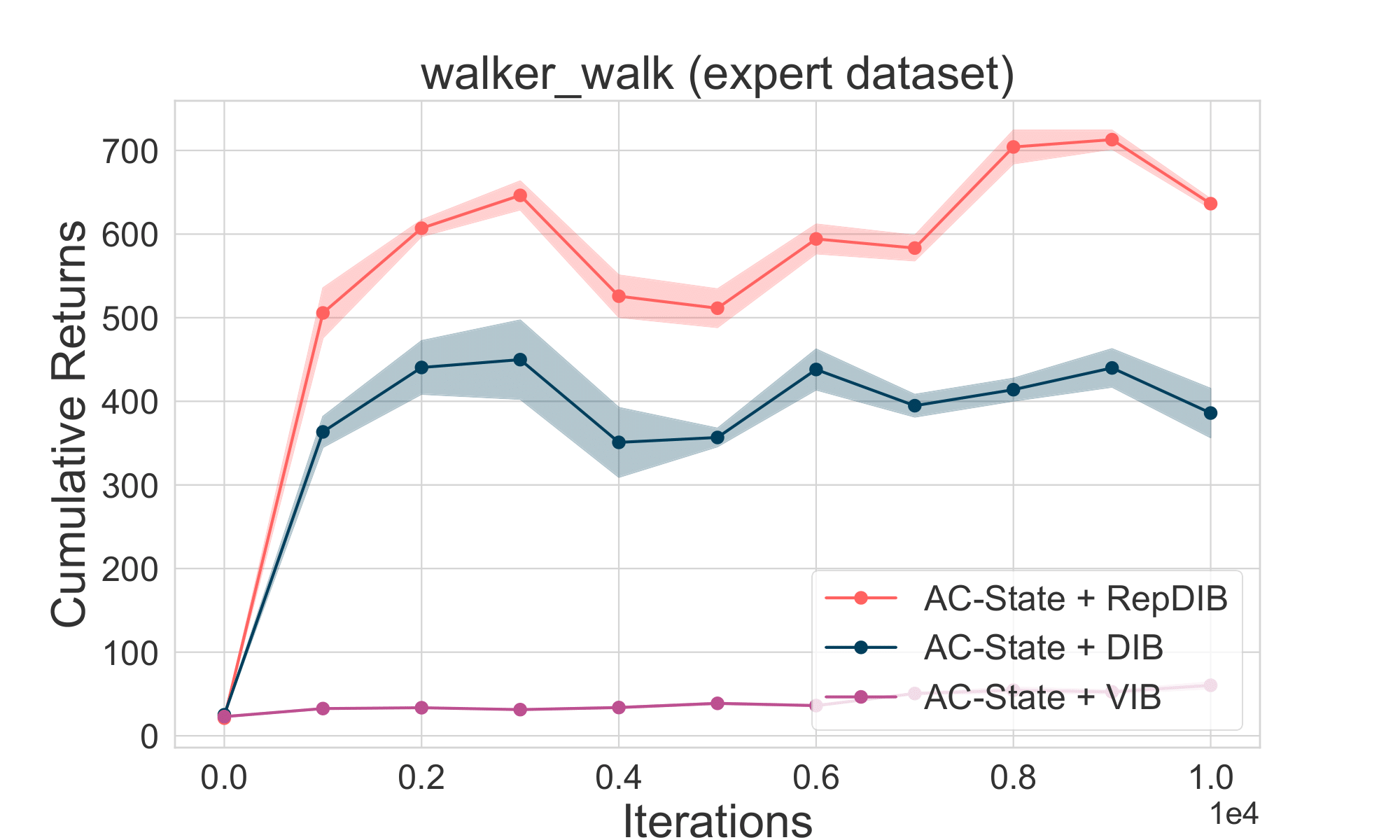}
}
\subfigure{
\includegraphics[
trim=1cm 0cm 1cm 0cm, clip=true,
width=0.3\textwidth]{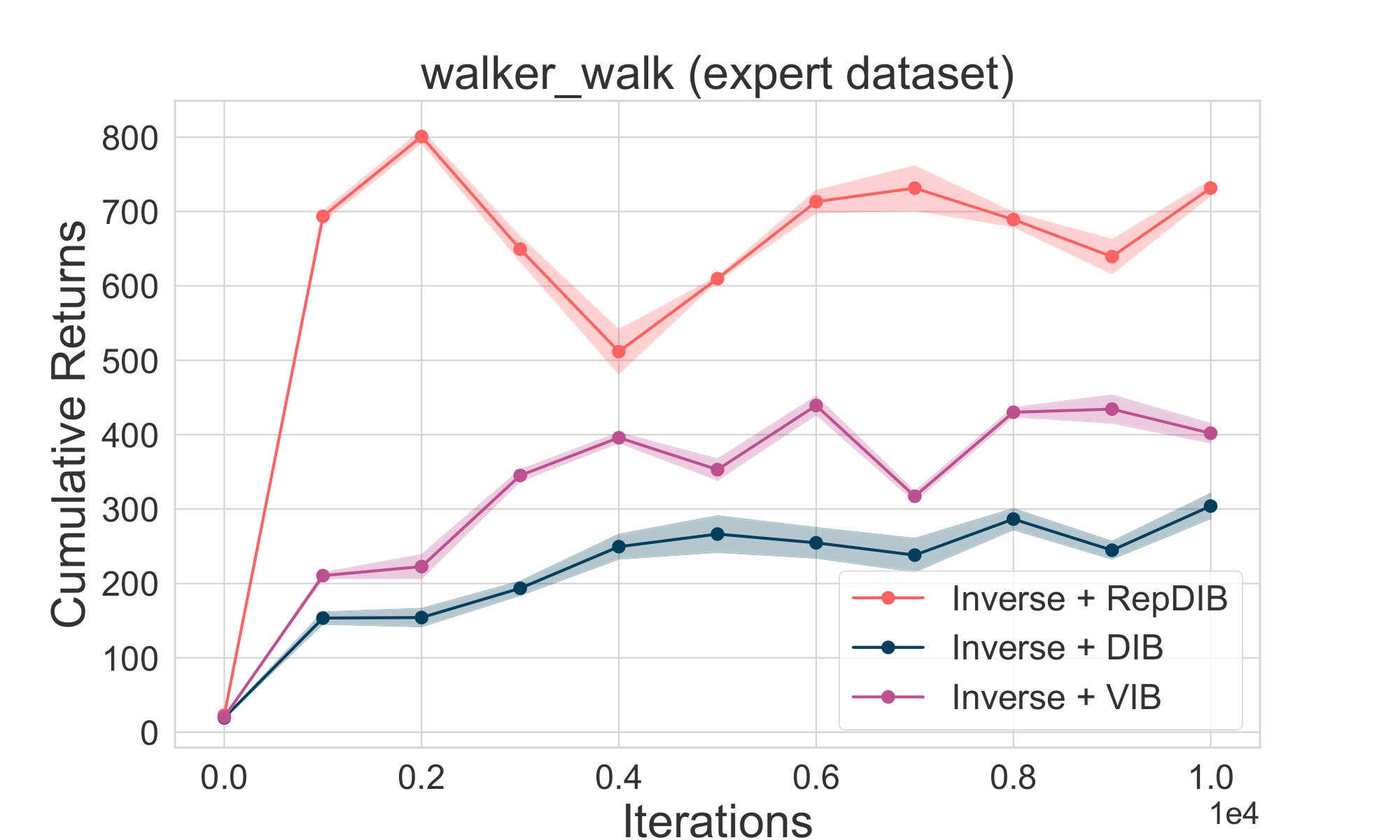}
}
\caption{Ablation studies on the \modelname bottleneck on changing background video based exogenous distractors in the observations of offline datasets, as per the setup described in section \ref{sec:offline_exo}}
\label{fig:rebuttal_ablation_video}
\end{figure}

In this section, we include additional results based on ablation studies of the \modelname objective. In figures \ref{fig:rebuttal_ablation_image} and \ref{fig:rebuttal_ablation_video} we include ablation studies where we compare \modelname with \textit{only} using the discrete information bottleneck (DIB) compared to \textit{only} using the variational information bottleneck (VIB). We do this on top of several existing representation objectives as described in section \ref{sec:visual_offline_exps}. Experimental results show that the significance in performance improvement of \modelname can primarily be achieved when we use the  VIB bottleneck prior to the DIB bottleneck, as we have explained previously in the main draft. Without the combination of the two, simply using one of the bottlenecks does not lead to the expected performance improvements. 

\subsection{Generalization on Continuous Control Tasks using URLB Benchmark}

\begin{figure}[!htbp] 
\centering
\includegraphics[
width=0.9\textwidth]{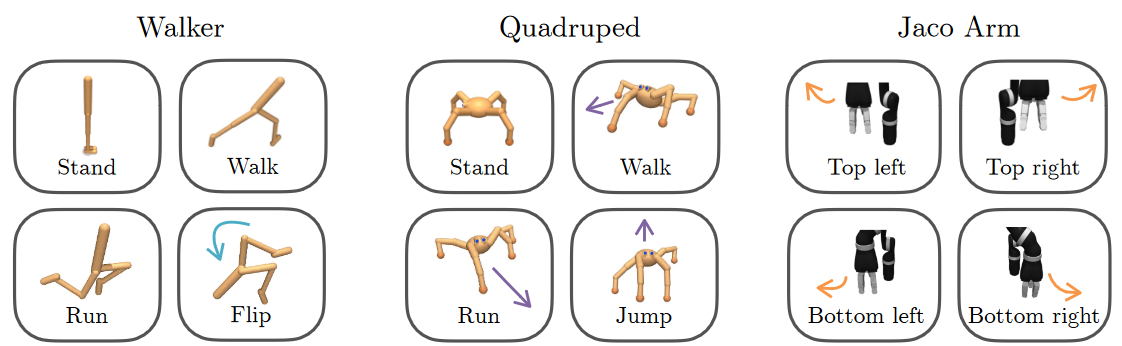}

\caption{The three domains (walker, quadruped, jaco arm) and twelve downstream tasks.}
\label{fig:dmc_env}
\end{figure}

\textbf{Experiment Setup and Details : } We follow the same set of domains and downstream tasks in \cite{URLB} (See Figure \ref{fig:dmc_env}). Specifically, from easiest to hardest, the domains and tasks are: \textbf{Walker} (\textit{Stand}, \textit{Walk}, \textit{Flip}, \textit{Run}):  An improved planar walker based on the one introduced in \cite{LillicrapHPHETS15}.  In \textit{Stand} task, the reward is a combination of terms encouraging an upright torso and some minimal torso height, and in \textit{Walk} and \textit{Run} tasks, the reward is proportional to forward velocity, while in \textit{Flip} task, it is relative to angular velocity.  \textbf{Quadruped} (\textit{Stand}, \textit{Walk}, \textit{Jump}, \textit{Run}): A quadruped within a a 3D space. The reward function defined in quadruped is similar to that in walker, but quadruped is harder due to a high-dimensional state and action spaces and 3D environment. \textbf{Jaco Arm} (\textit{Reach top left}, \textit{Reach top right}, \textit{Reach bottom left}, \textit{Reach bottom right}): Jaco Arm is a 6-DOF robotic arm with a three-finger gripper that tests the ability to control the robot arm to perform simple manipulation tasks. More detailed explanation refer to \cite{URLB}. In Table \ref{tab:hyper_cont} we present a set of hyper-parameters used in continuous control tasks.

\begin{figure*}[!htbp]
\centering
\subfigure{
\includegraphics[
width=0.33\textwidth]{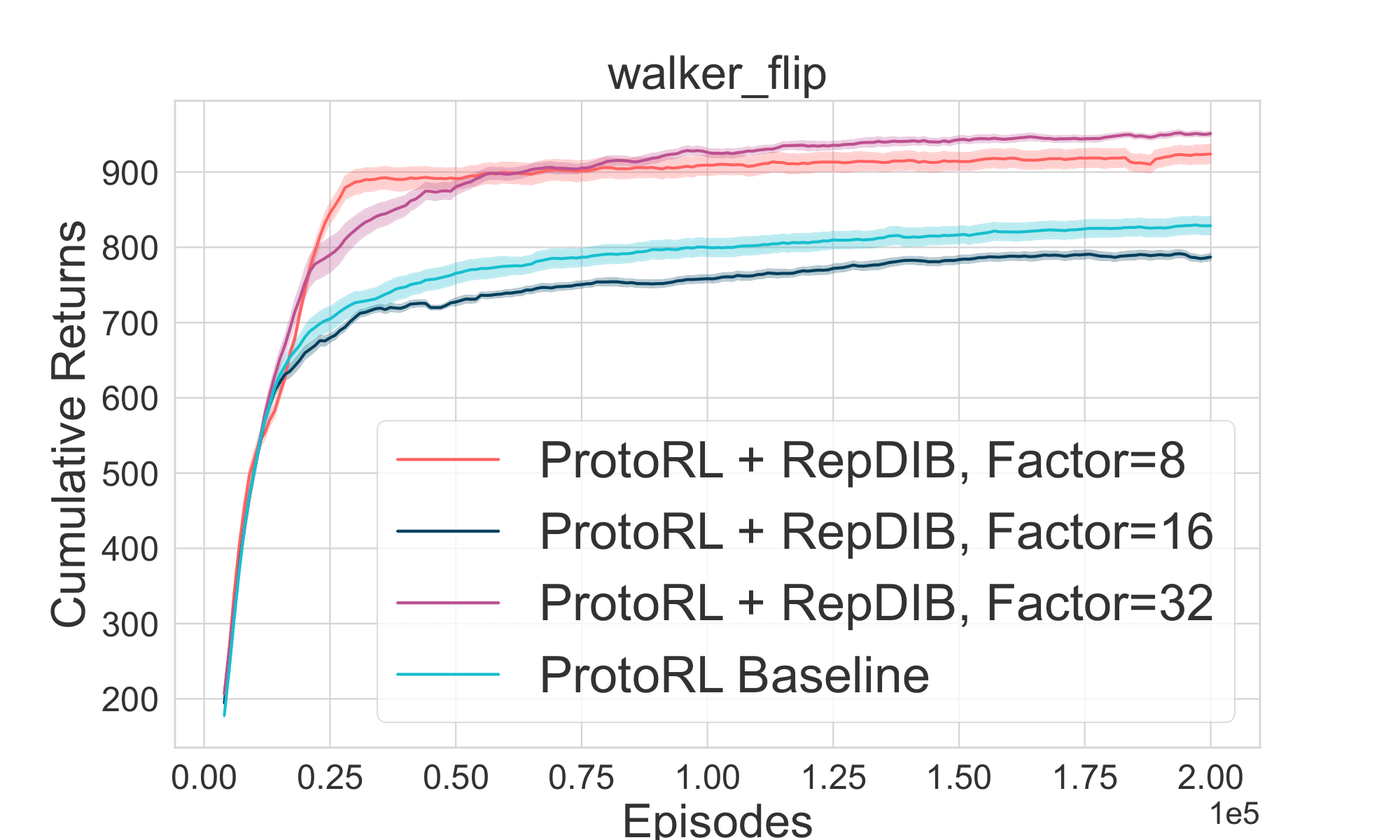}
}
\hspace{-0.8cm}
\subfigure{
\includegraphics[
width=0.33\textwidth]{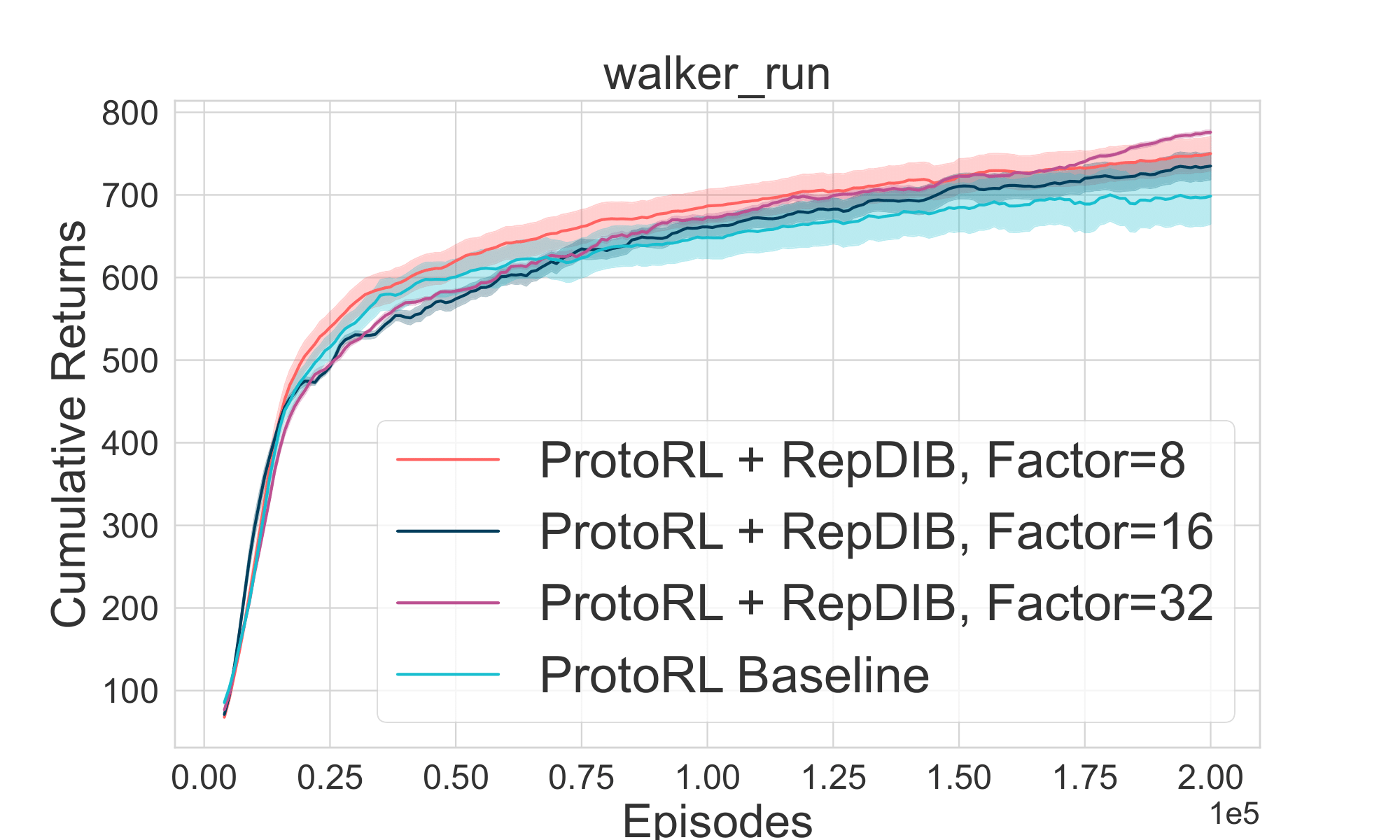}
}
\hspace{-0.8cm}
\subfigure{
\includegraphics[
width=0.33\textwidth]{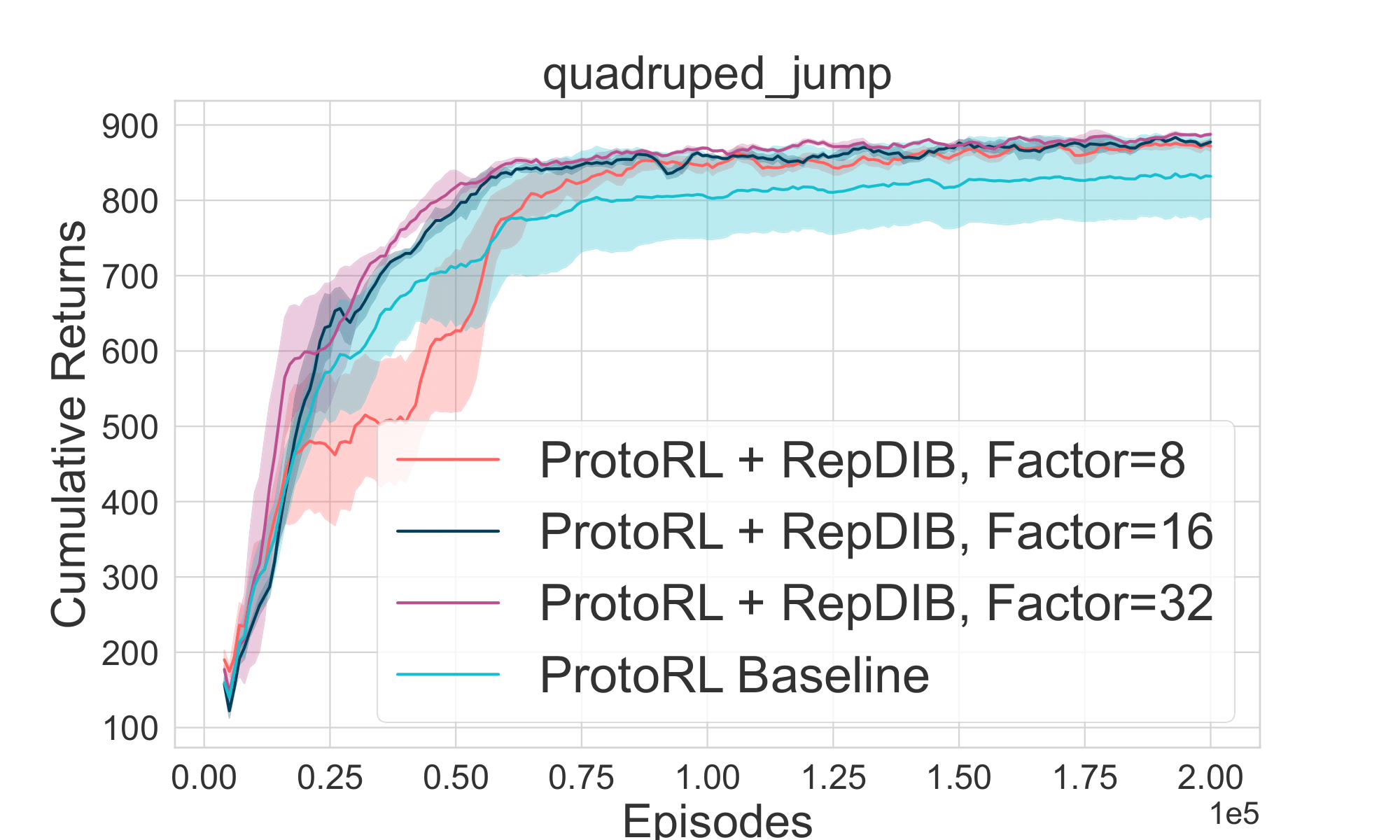}
}\\
\hspace{-0.8cm}
\subfigure{
\includegraphics[
width=0.33\textwidth]{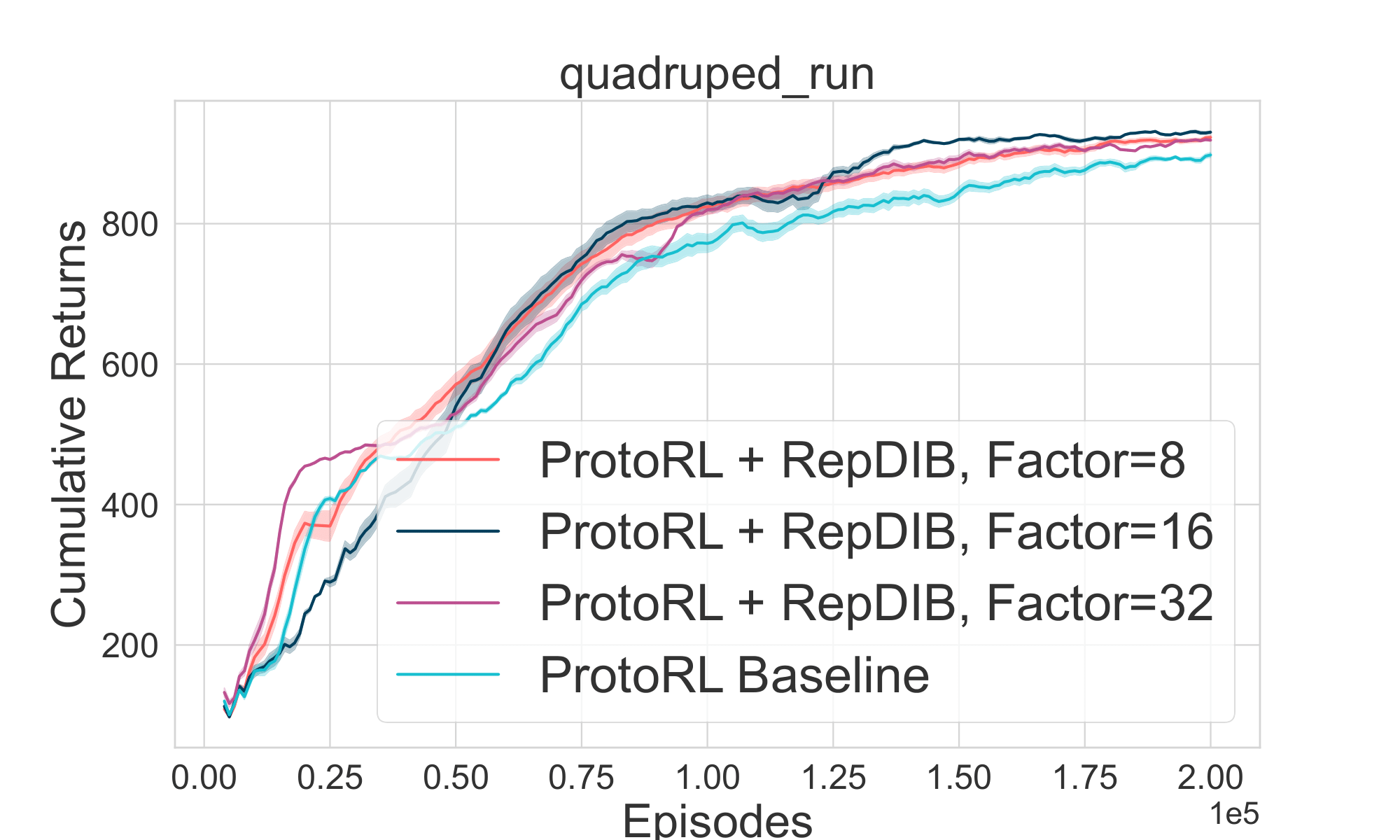}
}
\hspace{-0.8cm}
\subfigure{
\includegraphics[
width=0.33\textwidth]{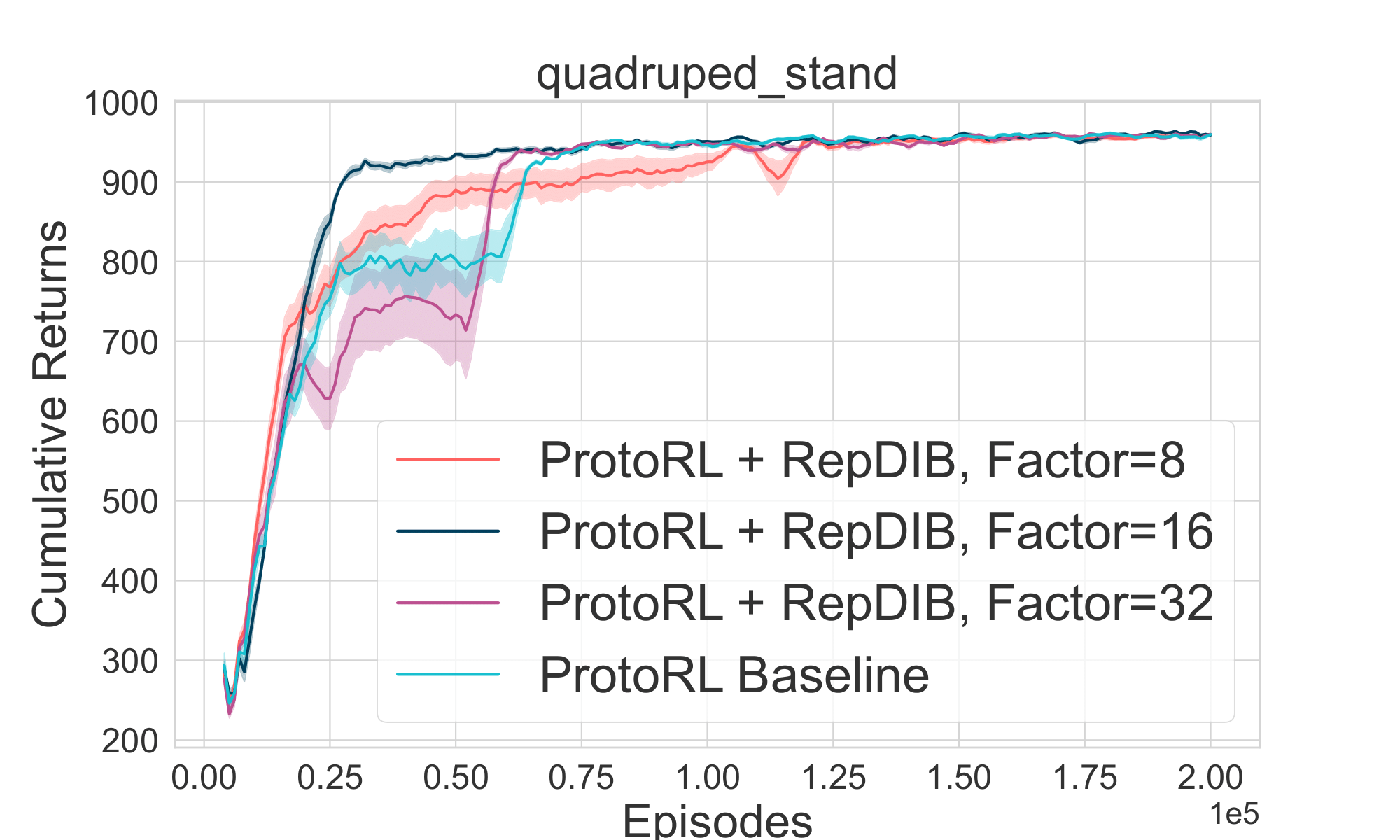}
}
\hspace{-0.8cm}
\subfigure{
\includegraphics[
width=0.33\textwidth]{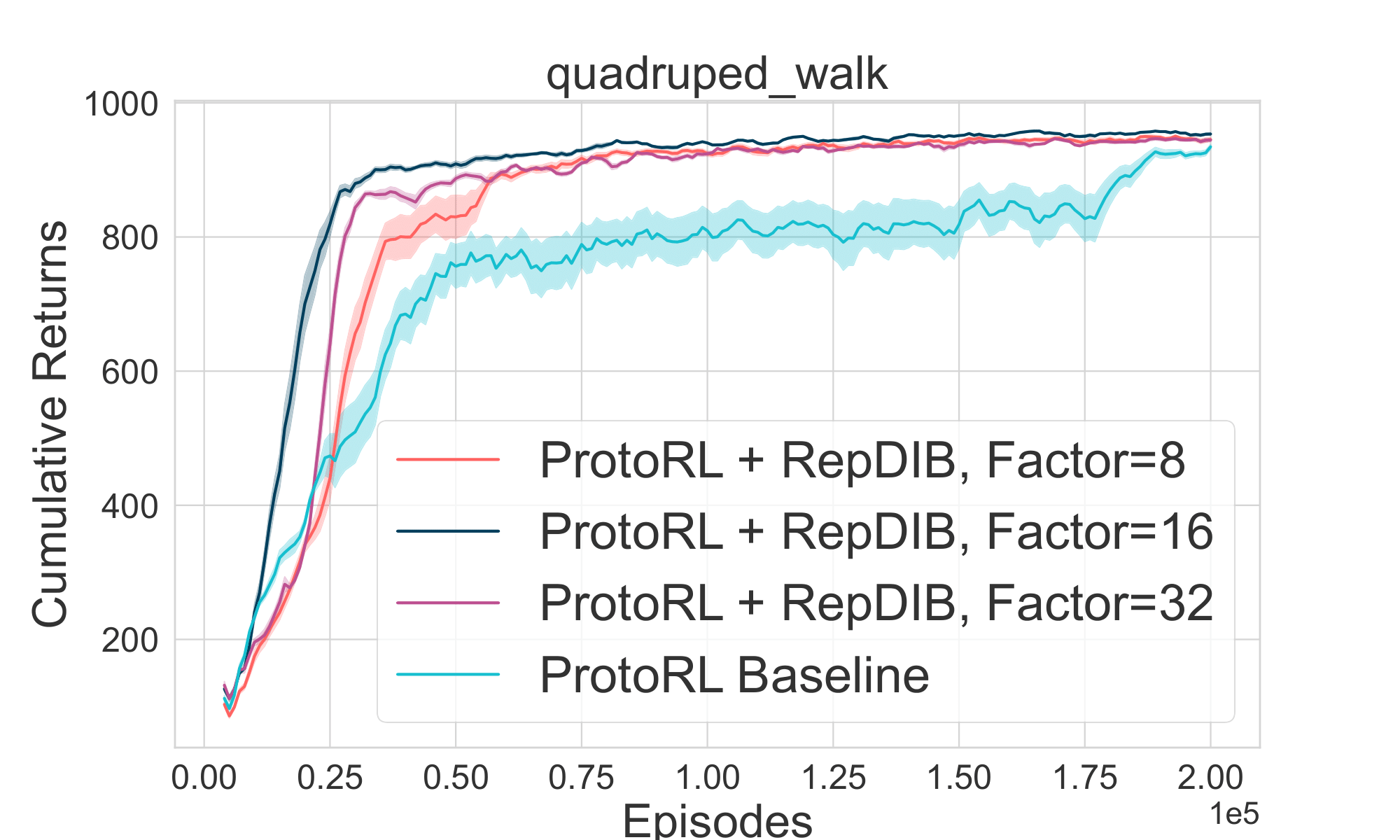}
}\\
\hspace{-0.8cm}
\subfigure{
\includegraphics[
width=0.33\textwidth]{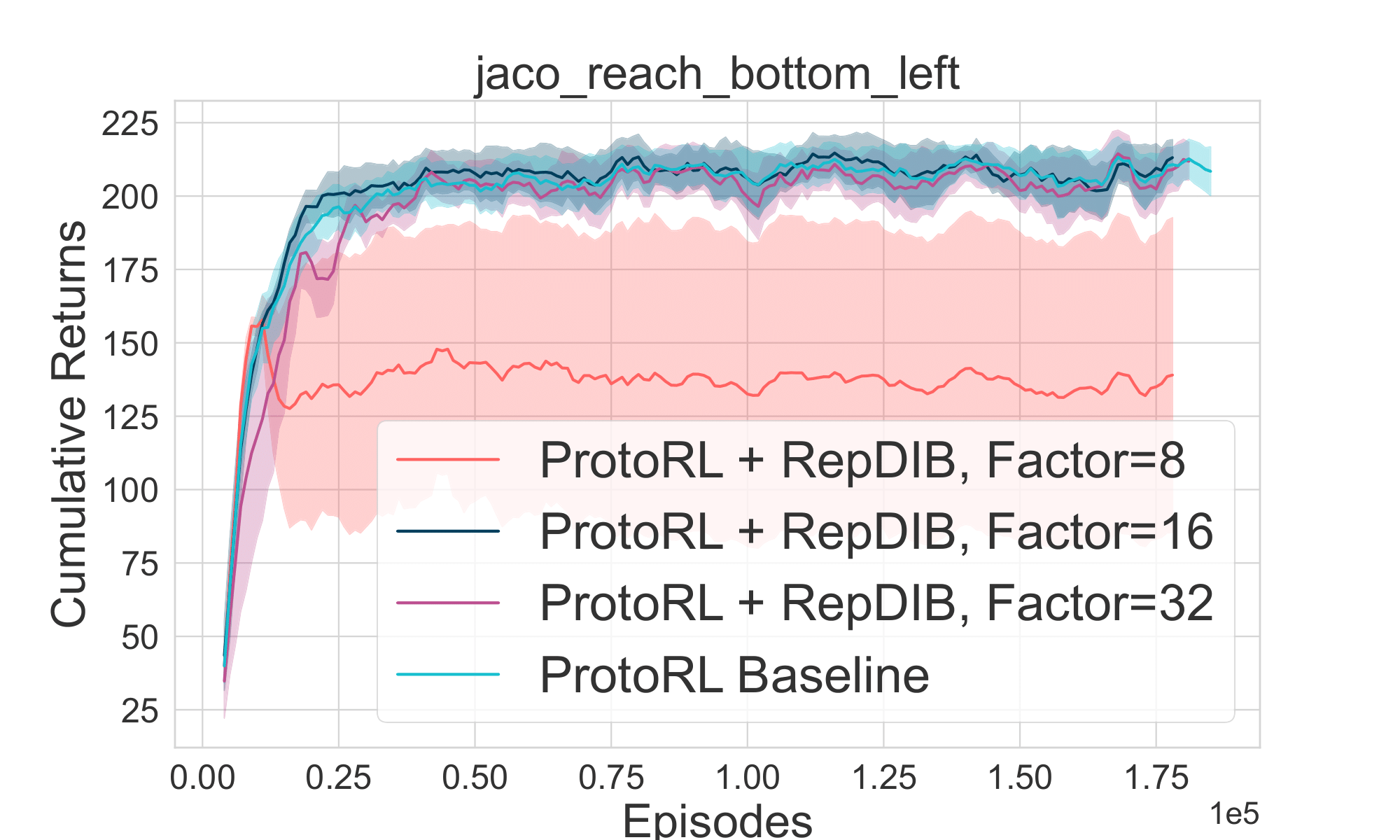}
}
\hspace{-0.8cm}
\subfigure{
\includegraphics[
width=0.33\textwidth]{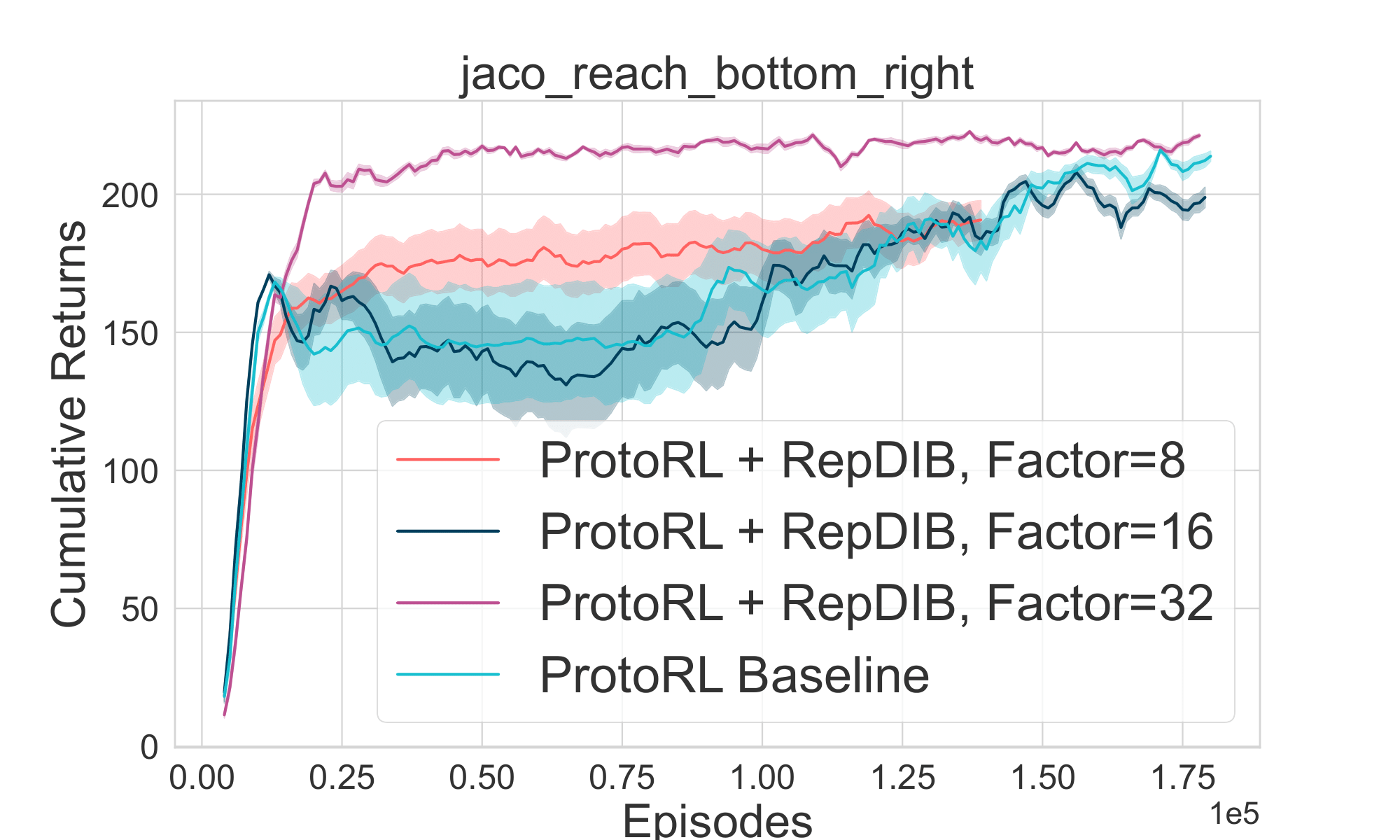}
}
\hspace{-0.8cm}
\subfigure{
\includegraphics[
width=0.33\textwidth]{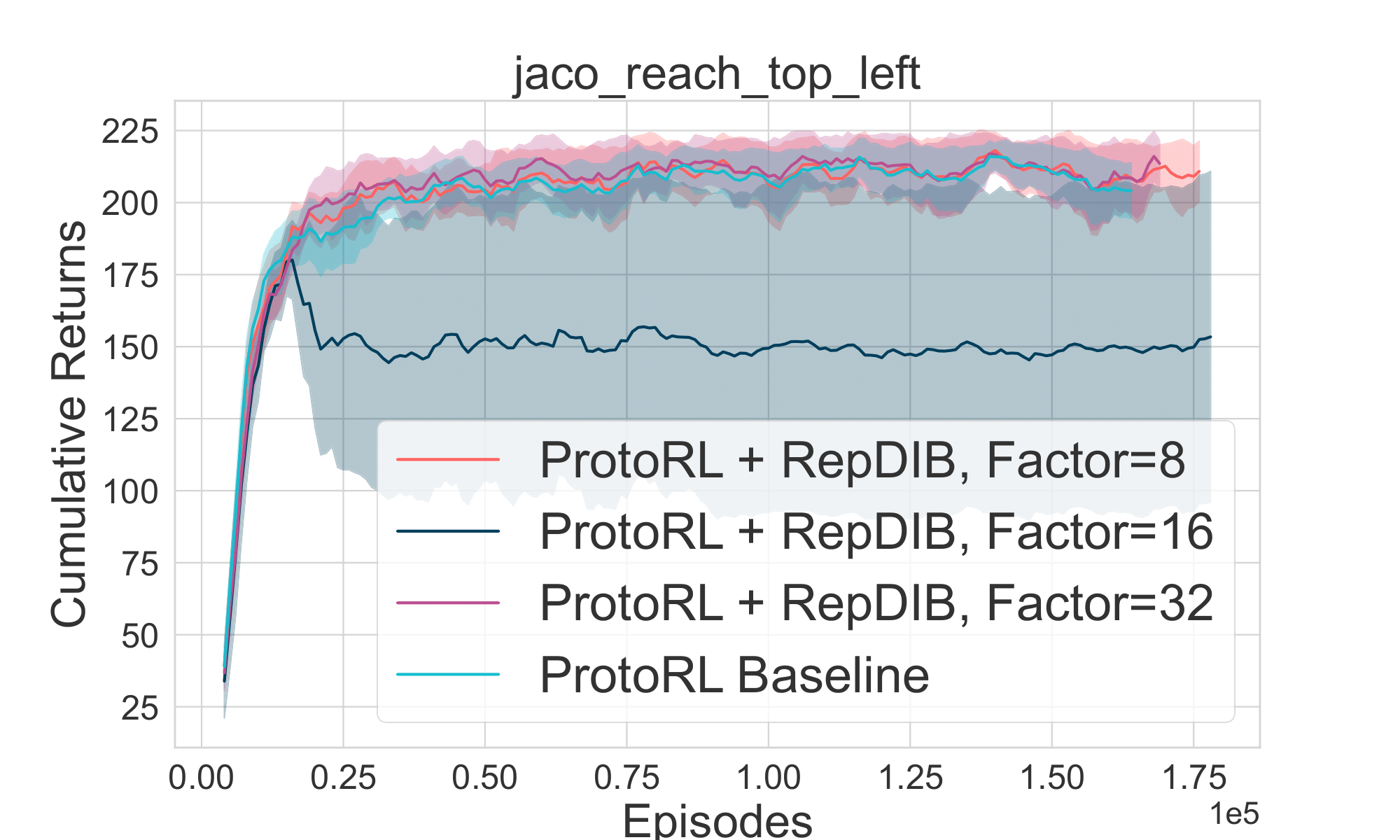}
}
\caption{\textbf{URLB Benchmark for Continuous Control} Ablation analysis on the URLB benchmark, integrating $\modelname$ on top of the ProtoRL baseline with different factorizations of the discrete bottleneck. Our experiment results show that the factorization in representation, depending on the number of factors, can play a vital role in improving the performance on the generalization task.}
\label{fig:urlb_ablations}
\end{figure*}

\textbf{Experiment Results : } For the online continuous control tasks, we test for generalization using the URLB benchmark \cite{URLB}, on 12 different environments as shown in figure \ref{fig:dmc_env}. In these experiments, representations are pre-trained on one environment for $100k$ pre-training steps, followed by fine-tuning both the RL algorithm and the encoder in a different environment. Existing baselines such as the ProtoRL \cite{yarats2021protorl} has already shown impressive performance compared to other baselines on the URLB benchmark. For more details on experiment setup and comparisons of ProtoRL with other baselines, see \cite{URLB}. In this task, we take the open source code of the ProtoRL baseline and simply integrate $\modelname$ on top of the encoders, where we have different factors for learning representations. The goal of the experiments is to show that when using compressed representations that are structured and factorial in nature, the compression on the pre-training task helps in fine-tuning on other tasks when the same bottleneck is again applied. Figure \ref{fig:urlb_ablations} summarizes the ablation studies of $\modelname$ built on top of the ProtoRL baseline. Our results in figure \ref{fig:urlb_ablations} show that fine-tuning performance is mostly improved compared to the baseline, typically for higher factors of representation.

\subsection{Robot Arm Experiment}

\begin{figure}[!htbp] 
\centering
\subfigure{
\includegraphics[
width=0.33\textwidth]{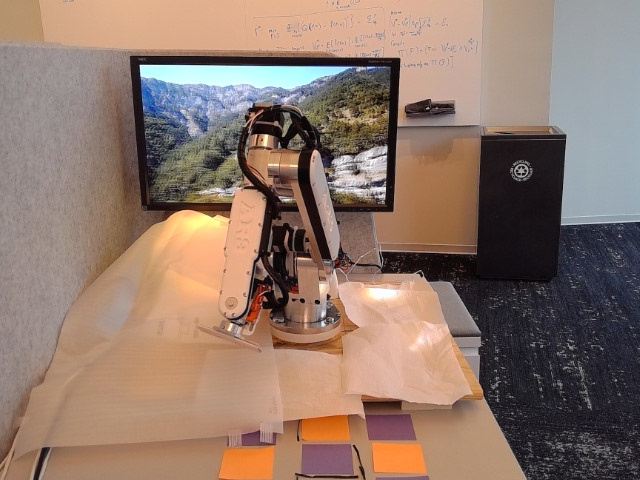}
}
\hspace{-0.8cm}
\subfigure{
\includegraphics[
width=0.33\textwidth]{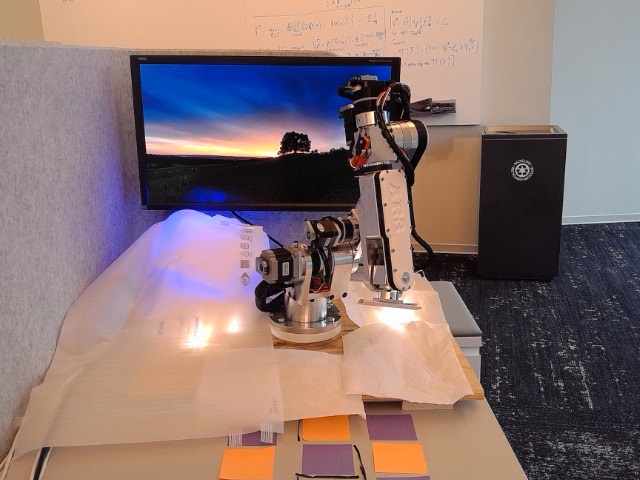}
}
\hspace{-0.8cm}
\subfigure{
\includegraphics[
width=0.33\textwidth]{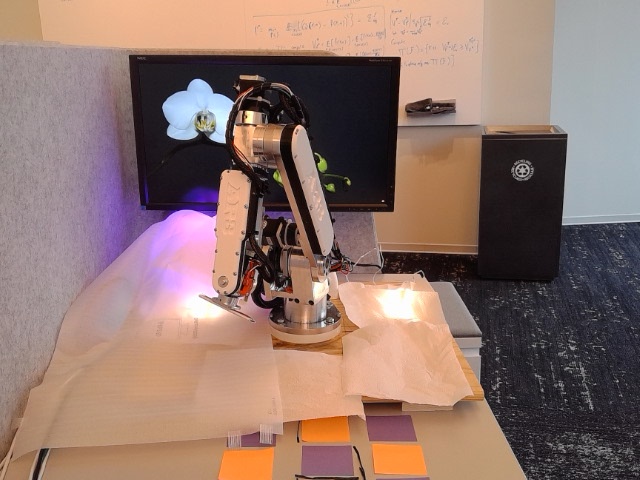}
}
\caption{Experiment setup for robot data collection in presence of exogenous or irrelevant background information. Figure shows three different images, with varying background information, for the robot arm, which are part of the collected dataset.}
\label{fig:robot_arm}
\end{figure}

\paragraph{Robot Arm Experiment with Background Video Exogenous Distractors:}  The robot arm in our experiments moves in a grid with $9$ different positions. We use two cameras to take images, for the dataset, one from the front side of the robot and the other with a top down view from above. We collect an image after each action is taken. The robot has 5 actions to take : move forward, backwards, right, left or stay in the current state. We use an episodic length of $500$, ie, the robot arm moves for $500$ steps after which we re-calibrate. The robot arm dataset is collected with a random uniform policy, for a total of $6$ hours collecting $14000$ samples. 

For learning the representation $\phi$ given the images, we use a small convolutional neural network to get an estimate $\phi(x)$ of the images $x$. In addition to the CNN network, we further learned the latent state representation with a multi-step inverse dynamics model $p(a \mid \phi(x), \phi(x_k))$, which predicts actions, given current representation $\phi(x)$ and a future representation $\phi(x_k)$. The model is trained with a cross entropy loss, with the ground truth actions available in the dataset. We use a metric of classification accuracy for evaluating the performance of $\modelname$.

\begin{figure}
    \centering
   \subfigure[ Seaquest]{\includegraphics[width = 6cm]{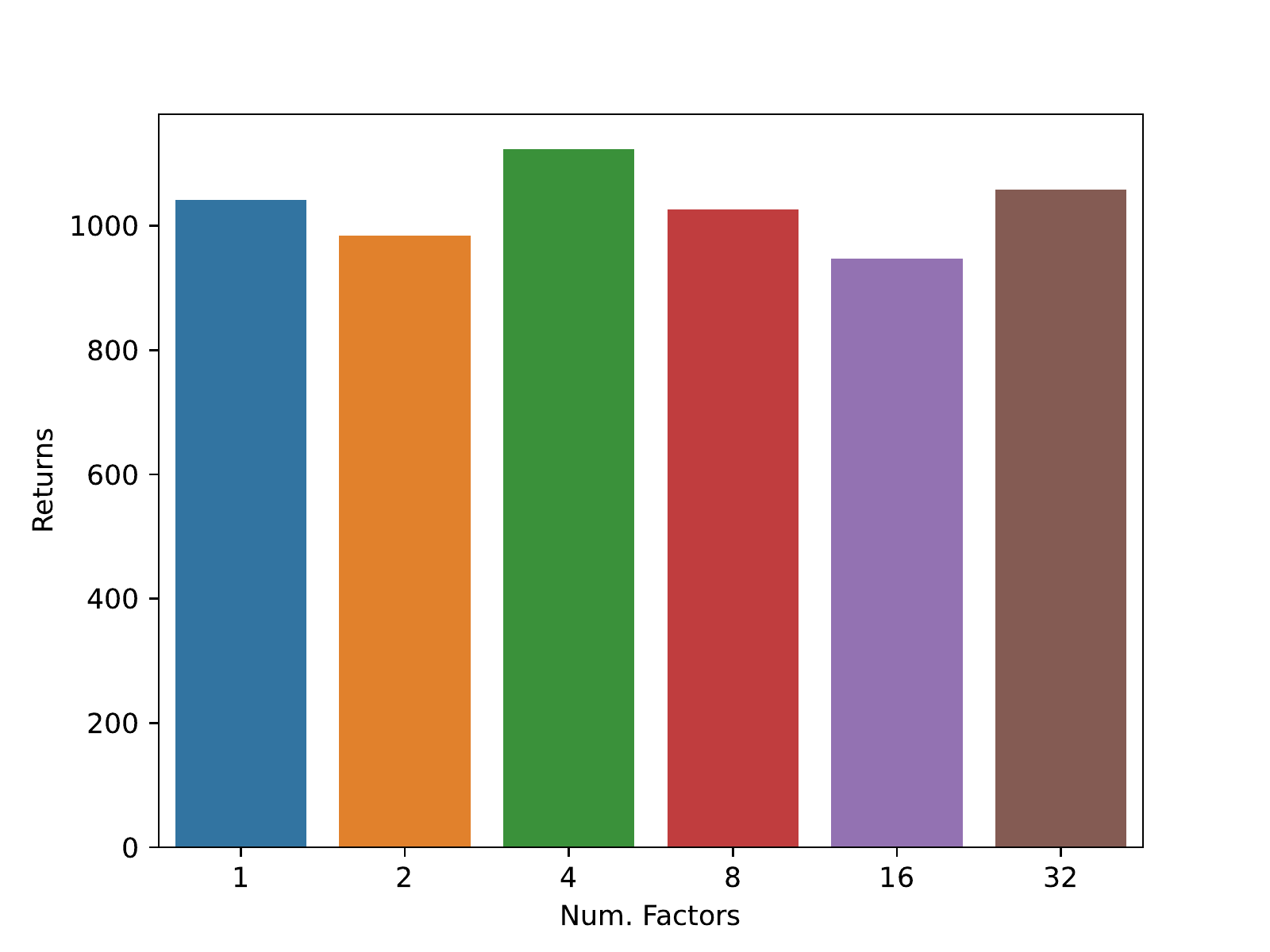}}
   \subfigure[ Breakout]{\includegraphics[width = 6cm]{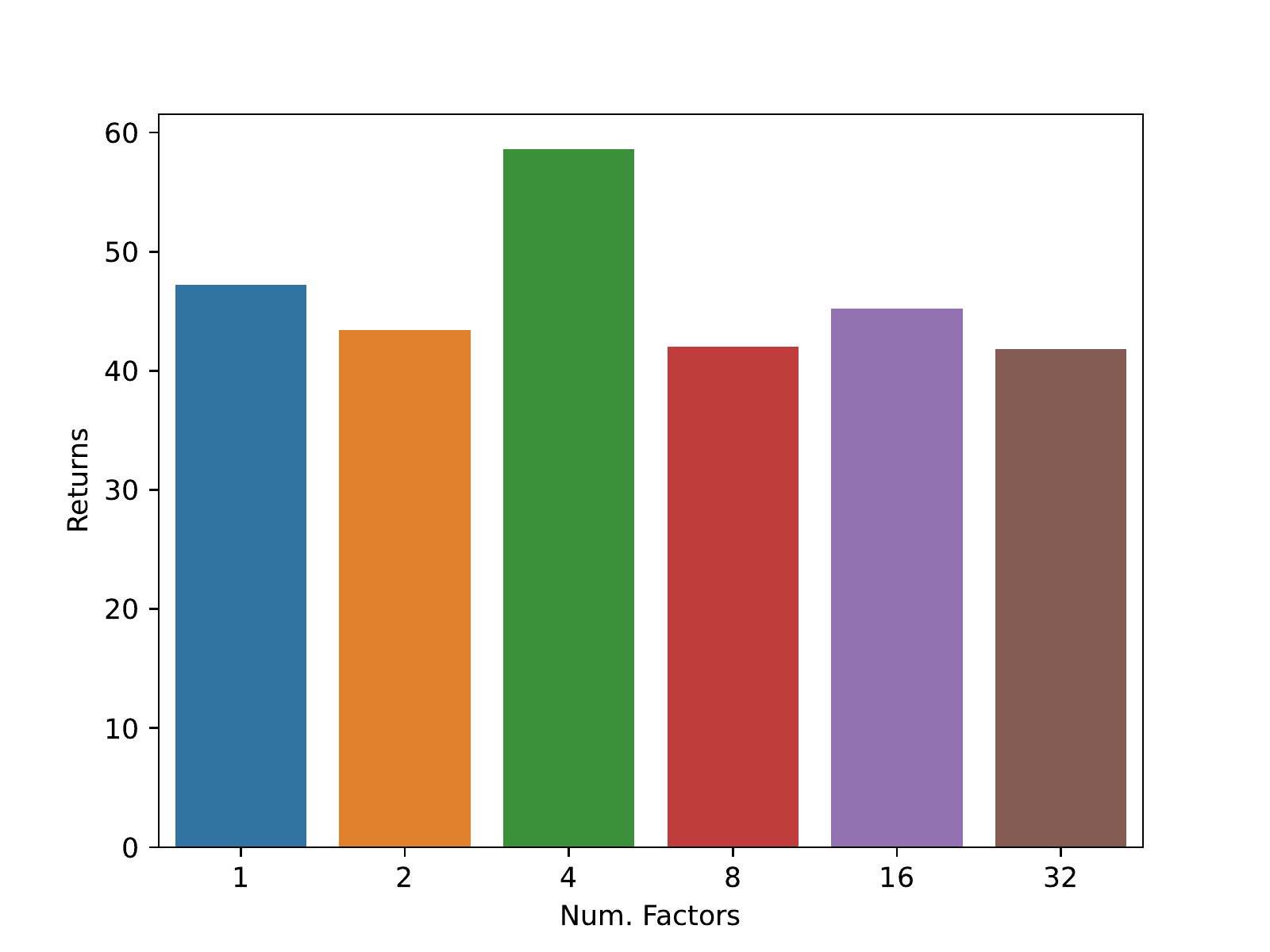}}
    \caption{Here we show the effect of the number of discretization factors on the model performance for 2 different Atari games. On the ALE benchmark, we find that factor $4$ usually outperforms other factors when learning representations with factorial structure.}
    \label{fig:factors}
\end{figure}

\subsection{Atari Benchmark with Exogenous Observations}

\begin{figure}
    \centering
   \subfigure[ Pong]{\includegraphics{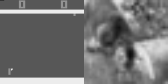}}
   \subfigure[ Qbert]{\includegraphics{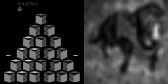}}
   \subfigure[ Seaquest]{\includegraphics{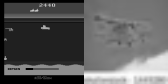}}
   \subfigure[ Breakout]{\includegraphics{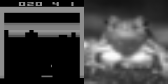}}
    \caption{Example observations from 4 different Atari games, with exogenous images placed on the side of environment observations. We add exogenous noise to show the importance of learning robust representations using an information bottleneck following $\modelname$. }
    \label{fig:atari_example_observations}
\end{figure}

\textbf{Experiment Setup : } We follow the experiment setup of decision transformers on the Atari domain following \cite{NEURIPS2021DT}. However, in addition to the environment observations from Atari games, we additionally augment the observations with an exogenous noise on the side. For this, we use CIFAR images placed on side of environment observations as exogenous noise. In Figure \ref{fig:atari_example_observations}, we show example observations from atari games with exogenous noise added. The goal is to see the effect of $\modelname$ when integrated on top of a multi-step inverse dynamics objective for learning robust representations \cite{lamb2022guaranteed}. We keep most of the hyperparameter details same as used in \cite{NEURIPS2021DT}. They use episodes of fixed length during training - also referred to as the \textit{context length}. We use a context length of 30 for Seaquest and Breakout. Similar to \cite{NEURIPS2021DT}, we consider one observation to be a stack of 4 atari frames. To implement the multi-step inverse objective, we sample 8 different values for $k$ and calculate the objective for each value of $k$, obtaining the final loss by taking the sum across all the sampled values of $k$. We do not feed the embedding for $k$ in the MLP that predicts the action while computing the multi-step inverse objective. 

Figure \ref{fig:factors} shows the effect of different factors used in the discrete information bottleneck of $\modelname$. We check the effect of the number of discretization factors on the model performance. The effect of the number of factors on overall performance of the Decision Transformer can vary depending on the game and domain. 

\begin{figure}[!t]
    \centering
    \includegraphics[width=\textwidth]{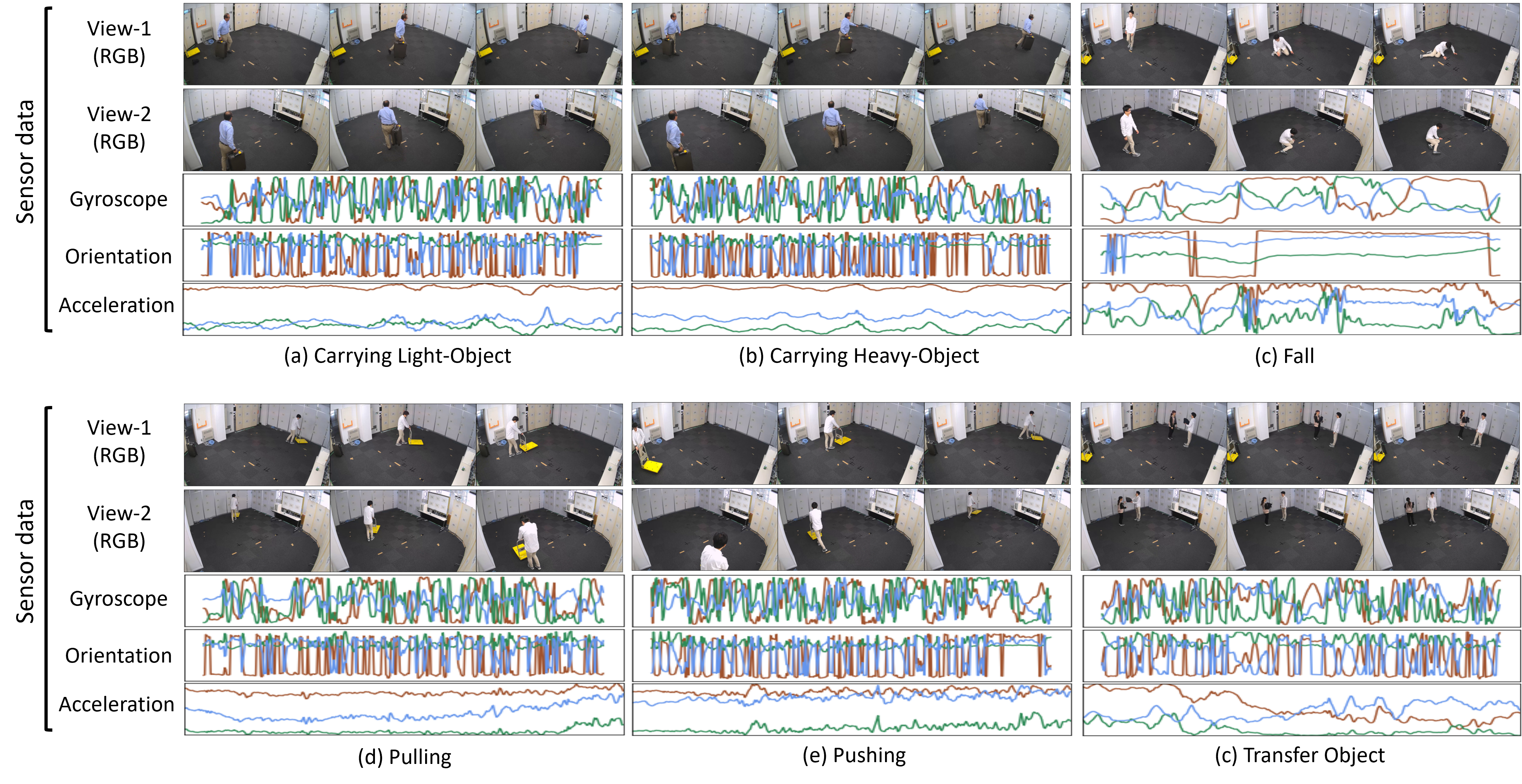}
    \caption{Sample data of human activities from MMAct dataset with five modalities: Visual views 1 \& 2, Gyroscope, Orientation, and Acceleration.)}
    \label{fig:mmact_sample}
\end{figure}

\subsection{Multi-Modal Representation Learning on Human Activity Recognition Task}
\begin{table}[!t]
\centering
\small
\begin{tabular}{ccc}
\toprule
    Method & F1-Score (\%) \\ \hline
    SVM+HOG \cite{ofli2013berkeley_mhad}  & 46.52 \\
    TSN (RGB) \cite{wang2016temporal_tsn}  & 69.20 \\
    TSN (Optical-Flow) \cite{wang2016temporal_tsn}  & 72.57 \\
    MMAD \cite{kong2019mmact} & 74.58 \\
    TSN (Fusion) \cite{wang2016temporal_tsn}  & 77.09 \\
    MMAD (Fusion) \cite{kong2019mmact} & 78.82\\
    Keyless \cite{keyless} & 81.11 \\
    HAMLET \cite{islam2020hamlet} & 83.89 \\
    RepDIB+MM(Keyless) & 71.35 \\
    \textbf{RepDIB+MM(RepDIB+Uni)} & \textbf{84.96} \\
\bottomrule
\end{tabular}
\caption{\textbf{Cross-session performance} comparison (F1-Score) of multimodal learning methods on MMAct dataset}
\label{tab:mmact_session}
\end{table}

\paragraph{Dataset: } MMAct dataset contains 37 activities (e.g., carrying objects, fall, kicking, talking on the phone, jumping, using PCs, sitting). Twenty people performed each activity five times, resulting in $37k$ data samples. All the activities are captured using data from seven modalities: four RGB views, acceleration, gyroscope, and orientation. We used data from two opposing RGB visual views, acceleration, gyroscope, and orientation modalities to train and test. MMAct dataset contains visually occluded data samples, which allows evaluating the effectiveness of HAR approaches for real-world settings. Human activity sample data are depicted in Figure~\ref{fig:mmact_sample}.

\paragraph{Experimental Setup for Multimodal Model Evaluation in Cross-Session Setting: } In this supervised learning task, the model uses multimodal sensor data to recognize human activities. We extend state-of-the-art multimodal representation learning models to extract salient representation using RepDIB information bottleneck. We extended the baseline multi-modal models in two ways to incorporate VQ bottleneck: \textbf{\modelname+MM: } We extract multi-modal representations using existing models (e.g. Keyless \cite{keyless} and HAMLET \cite{islam2020hamlet}) and then apply VQ bottleneck on the fused multi-modal representations. \textbf{\modelname+MM(\modelname+Uni):} We applied VQ bottleneck in two steps. First, we extract unimodal representations and apply VQ bottleneck to produce discretized unimodal representations. These discretized representations are fused and passed thorough a VQ bottleneck to produce task representations for the activity recognition.
In the baselines, we used five modalities: two viewpoints of RGB videos and three wearable sensors (acceleration, gyroscope, and orientation). We evaluated all the baselines on the MMAct dataset in a cross-session evaluation setting and reported F1-Score of activity recognition task \cite{kong2019mmact}. In the cross-session evaluation setting, the training and testing datasets can contain data from the same human subjects. 

We train these models using cross-entropy loss. We use Adam optimizer with weight decay regularization and cosine annealing warm restarts learning scheduler, where the initial learning rate is set to $3e^{-4}$.  To train the learning model on the MMAct dataset, we set the cycle length ($T_0$) and cycle multiplier ($T_{mult}$) to $30$ and $2$, respectively. We trained the models for $210$ epochs in the distributed GPUs cluster environment, where each node contains $8$ A100 GPUs. We used Pytorch and Pytorch-Lightning frameworks to implement all the models. To ensure reproducibility we a fixed seed.

\paragraph{Experimental Results: } The results in Table~\ref{tab:mmact_session} suggest that incorporating VQ bottleneck on the existing multi-modal learning model (Keyless) degrades the F1-score of activity recognition from $81.11\%$ to $71.35$. However, applying the VQ bottleneck both on the unimodal and multimodal representations improves the performance of the models compared to the models that do not use $\modelname$ or use $\modelname$ only on the multimodal representations. For example, $\modelname+MM(\modelname+Uni)$ model uses the same HAMLET model and applies $\modelname$ on the unimodal and multimodal representations. $\modelname+MM(\modelname+Uni)$ improves the F1-score of activity recognition to $84.96$ and outperforms all the evaluated multimodal models. Thus, hierarchical VQ bottlenecks can help to extract salient multimodal representation for accurately recognizing activities.

\subsection{Maze Navigation Tasks}

\begin{figure}[!htbp] 
\centering
\subfigure[GridWorld]{
\includegraphics[
width=0.3\textwidth]{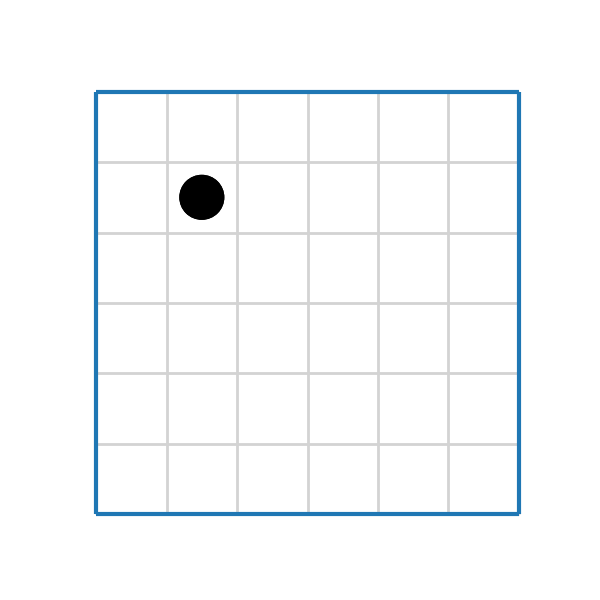}
}
\hspace{-0.8cm}
\subfigure[LoopWorld]{
\includegraphics[
width=0.3\textwidth]{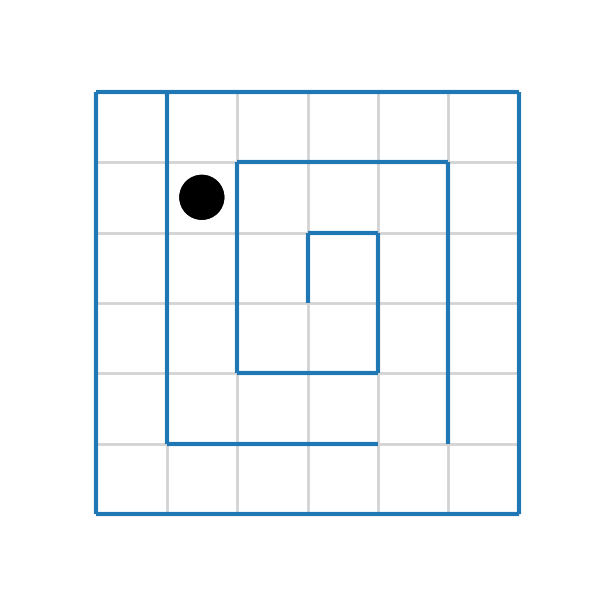}
}
\hspace{-0.8cm}
\subfigure[SpiralWorld]{
\includegraphics[
width=0.3\textwidth]{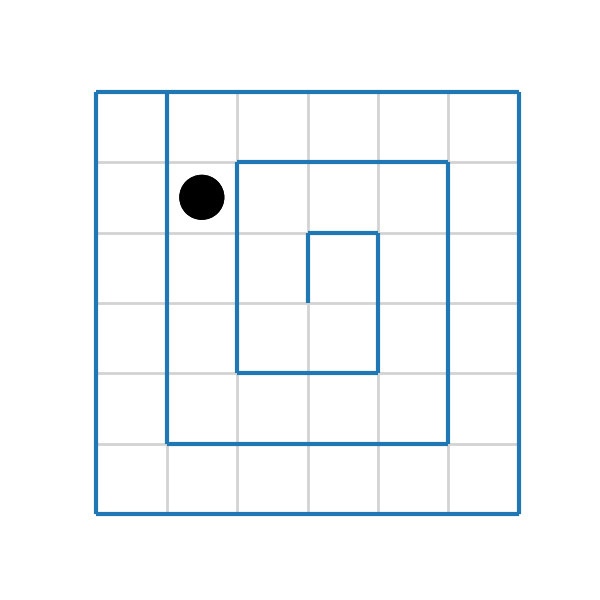}
}
\caption{Three environments in maze navigation tasks. The blue lines represent the walls.}
\label{fig:maze_env}
\end{figure}

\paragraph{Maze Navigation Tasks} We develop three kinds of environments for maze navigation tasks: \textit{GridWorld}, \textit{SpiralWorld}, \textit{LoopWorld} (see Figure \ref{fig:maze_env}). All of these environments share the same action space and state space, but their dynamics are slightly different. \textit{Gridworld} is the easiest task that without any walls so that the agent can go wherever it wants. \textit{SpiralWorld} is the hardest one that has spiral-shaped walls blocking the path of the agent, where the agent can only navigate along the spiral grid.  \textit{LoopWorld} is a variant of \textit{SpiralWorld} in which the agent can pass through a vacancy in the bottom right corner of the spiral-shaped wall. The task is to choose from one of four directions to travel in at each timestep. The reward function given is -1 at all steps until it reaches the goal where it receives a reward of 0 and the episode is terminated. During pre-training stage, we learn the state representations on \textit{GridWorld} with the data collected by a random policy. During the fine-tuning stage, the agent is trained to reach a goal from a small finite set of training goals, and the agent is tasked with reaching a fixed goal at the center of the maze during evaluation. In Table \ref{tab:hyper_maze} we present a set of hyper-parameters used in maze navigation tasks.

\section{Demonstrating Factorial Representation}
\label{rebuttal:factorial_representation}

\begin{figure*}
    \centering
   \subfigure[Brightness and Details]{\includegraphics[width=0.9\textwidth]{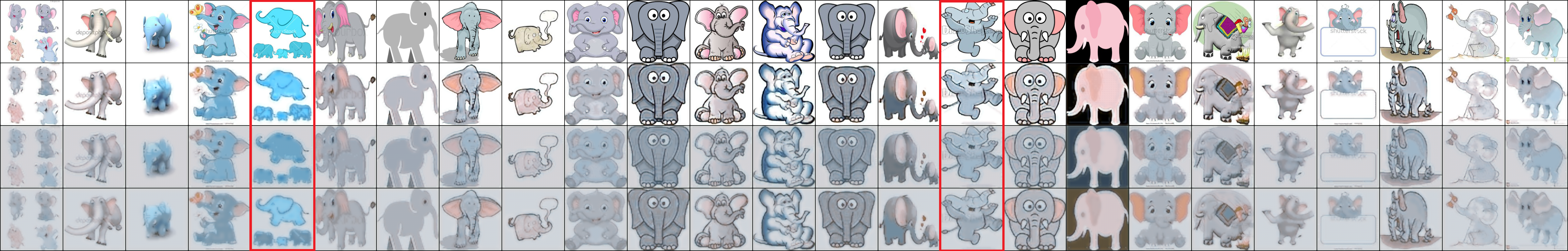}}
   \subfigure[Different colors]{\includegraphics[width=0.9\textwidth]{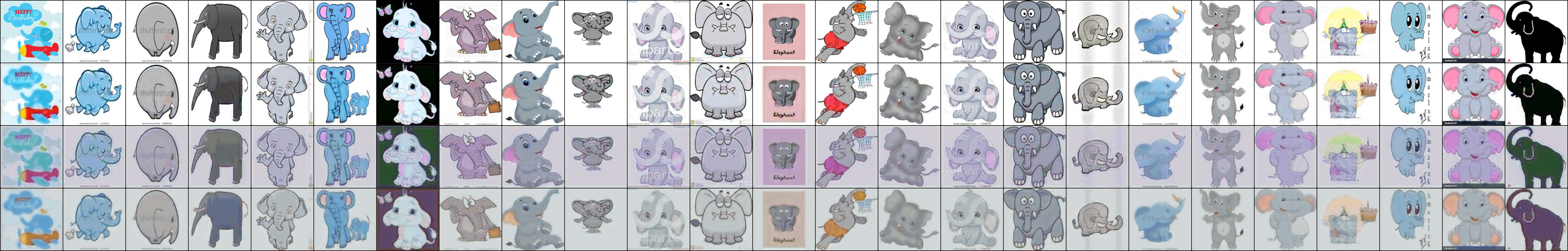}}
    \caption{PACS-cartoon-elephant dataset example to demonstrate factorized representations. Top row: Original Image; Second row: Reconstructed Image without substitution; Third row: Reconstructed Image with one groups of discrete codes substituted by zero vectors; Last row: Reconstructed Image with the other groups of discrete codes substituted by zero vectors.}
    \label{fig:pacs}
\end{figure*}

We demonstrate that with the discrete factorial information bottleneck, the agent is capable of learning factorial representations on real world data. We provide more details as follow.
\paragraph{Experiment details} To investigate whether the agent has the ability to learn semantic factorial representation with \modelname, we use the cartoon domain images from a benchmark dataset called PACS~\cite{PACS}, where only the elephant category is utilized for training and evaluation for the purpose of the intuitively illustration. The pixel-based input, with the size of 224x224, is first passed through an encoder (consists of CNN layers with the resnet block) to obtain its latent representation with the dimension of 32, then latent representation is quantized into two groups of discrete codes, where the codebook size is 512. After that, two groups of discrete codes are concatenated to obtain the representation, and finally passed through a decoder network (consists of CNN layers with the resnet block). Here we used reconstruction loss (MSE loss) combining with the loss for vector quantization to train the network.  For visualizing the semantic meaning of different groups, we randomly sample 25 pictures from the dataset, and pass the images into the network to obtain reconstruction of the images. Ideally, we would like to know whether different groups capture different semantic meaning of one image. For this purpose, we used zero vector to substitute one group of the discrete codes and acquire the reconstructed image by concatenating it with the other group of the discrete codes. As a consequence, we have three reconstructed images in total, as shown in Figure~\ref{fig:pacs}.
\paragraph{Experiment Results} Figure~\ref{fig:pacs} shows the reconstructed images from a trained decoder operating on a discretized 2-factor representation. We find that different factors capture different semantic information. As an example, it is obvious to see that there are 4 elephants in the fifth column in Figure~\ref{fig:pacs}(a), where the elephant at the top and the elephant at the bottom-middle are brighter than the other two elephants. For this image input, factor 1 tends to only capture the shape of the elephant without the brightness, while factor 2 capture specific details of each elephant. The similar observation can be found in the 16th column, where factor 1 captures the ``shadow'' in the picture, and factor 2 captures the brightness of elephant's skin. Another example is in Figure~\ref{fig:pacs}(b), it is shown that two factors learn ``green'' and ``purple'' separately for reconstructing ``black'', and two factors learn ``pink'' and ``orange'' separately for reconstructing ``red''.

\section{Explanation and Significance of \modelname}
\label{rebuttal:significance_work}

We would like to provide further clarification about the significance of our work. In this work, we do not propose any new representation learning objective; rather we simply propose that discrete information based bottlenecks can be significant when it comes to learning representations. Moreover, an approach based on \modelname is demonstrated to be even more impactful especially when the learnt representation needs to discard exogenous or irrelevant information from the observations. We demonstrate this across a range of experiments, not only based on RL, but also based on other tasks such as human activity recognition. Our experiments however are primarily based on RL benchmarks, where we demonstrate that \modelname can be easily applied on top of any learnt representations. To do this, we take existing baseline approaches proposing representation learning objectives and demonstrate the ease with which \modelname can be integrated on top of learnt representations. 

We emphasize that although information bottlenecks has been studied extensively in past literature, the use of discrete information bottleneck is rather new; and moreover to apply bottlenecks on top of representation learning objectives, especially to discard exogenous information, has been little studied in the past. Our aim is to propose information bottleneck, which not only captures factorial or compositional representations, but also plays key role in extracting only the relevant latent representation; and most importantly, can be suitably applied on any deep RL algorithm relying on additional representation learning module. 
\section{Visualization}

In our experiments, we constantly find that the use of a variational information bottleneck (VIB) prior to the discretization bottlenck significantly helps performance of $\modelname$

\begin{figure}[ht]
\centering
\subfigure{
\includegraphics[
width=0.33\textwidth]{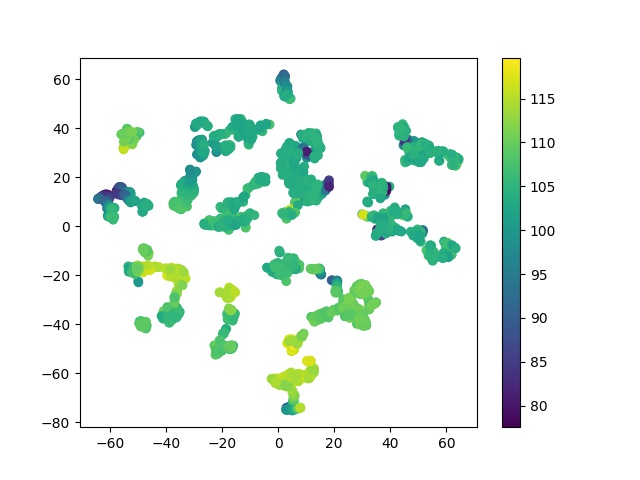}
}
\hspace{-0.8cm}
\subfigure{
\includegraphics[
width=0.33\textwidth]{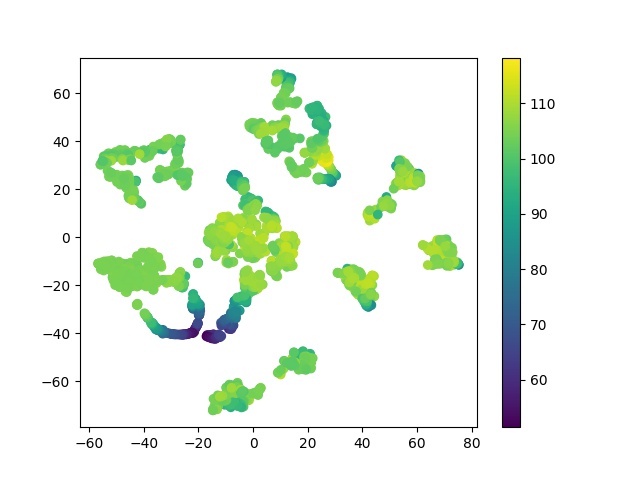}
}
\hspace{-0.8cm}
\subfigure{
\includegraphics[
width=0.33\textwidth]{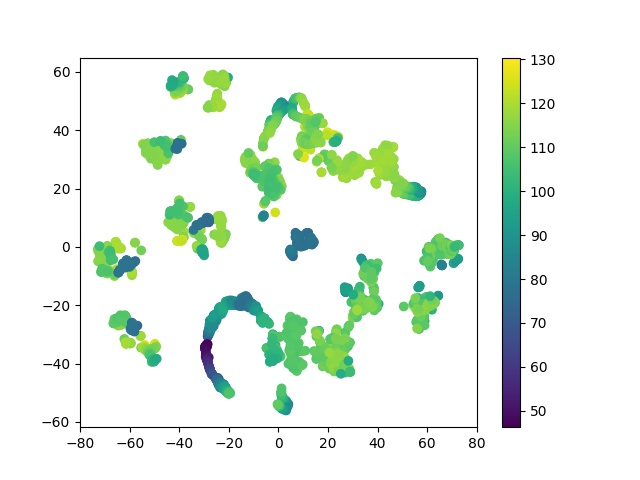}
}
\caption{\textbf{Comparing Visualizations with and without bottleneck representations} t-SNE of latent spaces in the JacoReachTopRight task learned with Proto-RL (left t-SNE), $\modelname$ (middle t-SNE), and $\modelname$ with VIB (right t-SNE) after training has completed, color-coded with predicted state values (higher value yellow, lower value purple).}
\label{fig:vib_kl_weightings}
\end{figure}

\section{Hyperparameter Details}

\begin{table}[h]
\caption{\label{tab:hyper_maze} A set of hyper-parameters used in maze navigation tasks.}
\centering
\begin{tabular}{lc}
\hline
Hyper-parameter       & Value \\
\hline
\: Size of Maze & $6 \times 6$ \\
\: Mini-batch size & $128$\\
\: Discount ($\gamma$) & $0.99$ \\
\: Optimizer & Adam \\
\: Learning rate & $3\times 10^{-3}$ \\
\: Critic target EMA rate ($\tau_Q$) & $0.01$ \\
\: Features dim. & $128$\\
\: Hidden dim. & $128$ \\
\: Number pre-training frames & $1\times 10^4$ \\
\: Number of discrete codes & $50$\\
\: Number of groups & $8, 16, 32$ \\
\: VIB coefficient & $0.01$ \\
\hline
\end{tabular}

\end{table}

\begin{table}[h]
\caption{\label{tab:hyper_cont} A set of hyper-parameters used in continuous control tasks.}
\centering
\begin{tabular}{lc}
\hline
Common hyper-parameter       & Value \\
\hline
\: Replay buffer capacity & $10^6$ \\
\: Seed frames & $4000$ \\
\: $n$-step returns & $3$ \\
\: Mini-batch size & $1024$\\
\: Seed frames & $4000$ \\
\: Discount ($\gamma$) & $0.99$ \\
\: Optimizer & Adam \\
\: Learning rate & $10^{-4}$ \\
\: Agent update frequency & $2$ \\
\: Critic target EMA rate ($\tau_Q$) & $0.01$ \\
\: Features dim. & $1024$\\
\: Hidden dim. & $1024$ \\
\: Exploration stddev clip & $0.3$ \\
\: Exploration stddev value & $0.2$ \\
\: Number pre-training frames & up to $2\times 10^6$ \\
\: Number fine-turning frames & up to $2\times 10^6$  \\
\: Number of discrete codes & $50$\\
\: Number of groups & $8, 16, 32$ \\
\: VIB coefficient & $0.01$ \\
\hline
\end{tabular}

\end{table}

\begin{table}[h]
\caption{\label{tab:hyper_offline} A set of hyper-parameters used in offline tasks.}
\centering
\begin{tabular}{lc}
\hline
Common hyper-parameter       & Value \\
\hline
\: $n$-step returns & $3$ \\
\: Mini-batch size & $256$\\
\: Seed frames & $4000$ \\
\: Discount ($\gamma$) & $0.99$ \\
\: Optimizer & Adam \\
\: Learning rate & $3\times 10^{-4}$ \\
\: Critic target EMA rate ($\tau_Q$) & $0.01$ \\
\: Features dim. & $256$\\
\: Hidden dim. & $1024$ \\
\: Number pre-training frames & $1\times 10^5$ \\
\: Number fine-turning frames & $1\times 10^5$  \\
\: Number of discrete codes & $512$\\
\: Number of groups & $4, 8, 16, 32$ \\
\: VIB coefficient & $0.01$ \\
\hline
\end{tabular}

\end{table}

\clearpage
\newpage

\end{document}